\documentclass[10pt,twocolumn,letterpaper]{article}
\usepackage{authblk}

\usepackage[pagenumbers]{wacv}
\definecolor{dark}{rgb}{0, 0, 0}
\usepackage[pagebackref,breaklinks,colorlinks,allcolors=dark]{hyperref}
\usepackage{titletoc}
\usepackage[utf8]{inputenc}
\usepackage{times}
\usepackage{microtype}
\usepackage{url}
\usepackage{xspace}
\usepackage{amsmath}
\usepackage{amssymb}
\usepackage{amsfonts}
\usepackage{amsthm}
\usepackage{bm}
\usepackage{nicefrac}
\usepackage{cases}
\usepackage{siunitx}
\usepackage{graphicx}
\usepackage[dvipsnames]{xcolor}
\usepackage{pifont}
\usepackage{grffile}
\usepackage[export]{adjustbox}
\usepackage{booktabs}
\usepackage{multirow}
\usepackage{enumitem}

\usepackage{listings}
\usepackage{mdframed}
\usepackage{numprint}
\usepackage{epsfig}

\usepackage{algorithm2e}
\RestyleAlgo{ruled}
\usepackage{thmtools,thm-restate}

\usepackage[toc,page]{appendix}
\usepackage{etoc}

\usepackage{float} 
\usepackage{listings}
\DeclareCaptionStyle{ruled}{labelfont=normalfont,labelsep=colon,strut=off}
\floatstyle{ruled}
\newfloat{listing}{tb}{lst}{}
\floatname{listing}{Listing}

\title{SAGA: Learning Signal-Aligned Distributions for Improved Text-to-Image Generation}

\author[1]{Paul Grimal}
\author[2]{Michaël Soumm}
\author[1]{Hervé Le Borgne}
\author[1]{Olivier Ferret}
\author[3]{Akihiro Sugimoto}

\affil[1]{Université Paris-Saclay, CEA, List, F-91120, Palaiseau, France}
\affil[2]{Télécom Paris}
\affil[3]{National Institute of Informatics, Japan}

\makeatletter
\renewcommand\AB@affilsepx{, \protect\Affilfont}
\makeatother

\affil[ ]{\tt\small \{paul.grimal, herve.le-borgne, olivier.ferret\}@cea.fr, soumm@telecom-paris.fr, sugimoto@nii.ac.jp}

\newcommand{\sdone}{SD~1.4\xspace}
\newcommand{\sdthree}{SD~3\xspace}
\newcommand{\gsn}{GSN\xspace}

\newtheorem{lemma}{Lemma}

\newcommand{\Var}{\mathrm{Var}}

\newcommand{\unique}{\text{\textbf{uni}}}

\newcommand{\saga}{\textit{SAGA}\xspace}
\newcommand{\sagaone}{\textit{SAGA}\textsubscript{\unique}\xspace}
\newcommand{\sagavar}{\textit{SAGA}\textsubscript{$\boldsymbol{\Sigma}$}\xspace}
\newcommand{\sagaonevar}{\textit{SAGA}\textsubscript{\unique,$\boldsymbol{\Sigma}$}\xspace}
\newcommand{\sagaplus}{\textit{SAGA+}\xspace}
\newcommand{\sagaplusone}{\textit{SAGA}\textsubscript{\unique}\textsuperscript{+}\xspace}
\newcommand{\sagaplusvar}{\textit{SAGA}\textsubscript{$\boldsymbol{\Sigma}$}\textsuperscript{+}\xspace}
\newcommand{\sagaplusonevar}{\textit{SAGA}\textsubscript{\unique, $\boldsymbol{\Sigma}$}\textsuperscript{+}\xspace}

\newcommand{\sagaonebbox}{\sagaone}
\newcommand{\sagaonebboxgsn}{\sagaplusone}
\newcommand{\sagabbox}{\saga}
\newcommand{\sagabboxgsn}{\sagaplus}

\newcommand{\variantA}{\xspace}
\newcommand{\variantB}{$\blacktriangle$\xspace}
\newcommand{\sagaStepsevenMomentumzerodotfour}{\saga \variantB}
\newcommand{\sagaStepfiveMomentumzerodotseven}{\saga \variantA}
\newcommand{\sagaoneStepsevenMomentumzerodotfour}{\sagaone \variantB}
\newcommand{\sagaoneStepfiveMomentumzerodotseven}{\sagaone \variantA}
\newcommand{\sagavarStepsevenMomentumzerodotfour}{\sagavar \variantB}
\newcommand{\sagavarStepfiveMomentumzerodotseven}{\sagavar \variantA}
\newcommand{\sagaonevarStepsevenMomentumzerodotfour}{\sagaonevar \variantB}
\newcommand{\sagaonevarStepfiveMomentumzerodotseven}{\sagaonevar \variantA}

\SetKwInput{KwInput}{Input}             
\SetKwInput{KwOutput}{Output}

\newcommand{\citefigure}[1]{\Cref{#1}}

\newcommand{\citetable}[1]{\Cref{#1}}

\newcommand{\citealgorithm}[1]{\Cref{#1}}

\newcommand{\citesection}[1]{\Cref{#1}}

\newcommand{\citeappendix}[1]{\Cref{#1}}

\newcommand{\citeprop}[1]{Proposition~\ref{#1}}

\begin{document}

\maketitle

\begin{abstract}
    State-of-the-art text-to-image models produce visually impressive results but often struggle with precise alignment to text prompts, leading to missing critical elements or unintended blending of distinct concepts. We propose a novel approach that learns a high-success-rate distribution conditioned on a target prompt, ensuring that generated images faithfully reflect the corresponding prompts. Our method explicitly models the signal component during the denoising process, offering fine-grained control that mitigates over-optimization and out-of-distribution artifacts. Moreover, our framework is training-free and seamlessly integrates with both existing diffusion and flow matching architectures. It also supports additional conditioning modalities -- such as bounding boxes -- for enhanced spatial alignment. Extensive experiments demonstrate that our approach outperforms current state-of-the-art methods. Code available at \href{Code}{https://github.com/grimalPaul/gsn-factory}
\end{abstract}

\section{Introduction}

Despite significant progress of text-to-image (T2I) generative models~\cite{rombach2021highresolution,podell2023sdxl,ramesh2022hierarchical,saharia2022photorealistic, esser2024scalingrectifiedflowtransformers}, ensuring that generated images accurately reflect the user's prompt remains a challenge. Textual alignment failures have been widely reported~\cite{ramesh2022hierarchical,saharia2022photorealistic,chefer2023attendandexcite,feng2023trainingfreestructureddiffusionguidance}, where generated images fail to  capture the semantic meaning of the target prompt accurately. Two types of inconsistencies are particularly challenging (\citefigure{fig:illu_problem}):\textit{ catastrophic neglect}, where central elements are omitted, and \textit{subject mixing}, where distinct entities are incorrectly blended. These failures significantly impact user experience and limit the actual utility of T2I models.

\begin{figure}
    \centering
    \includegraphics[height=0.64\linewidth]{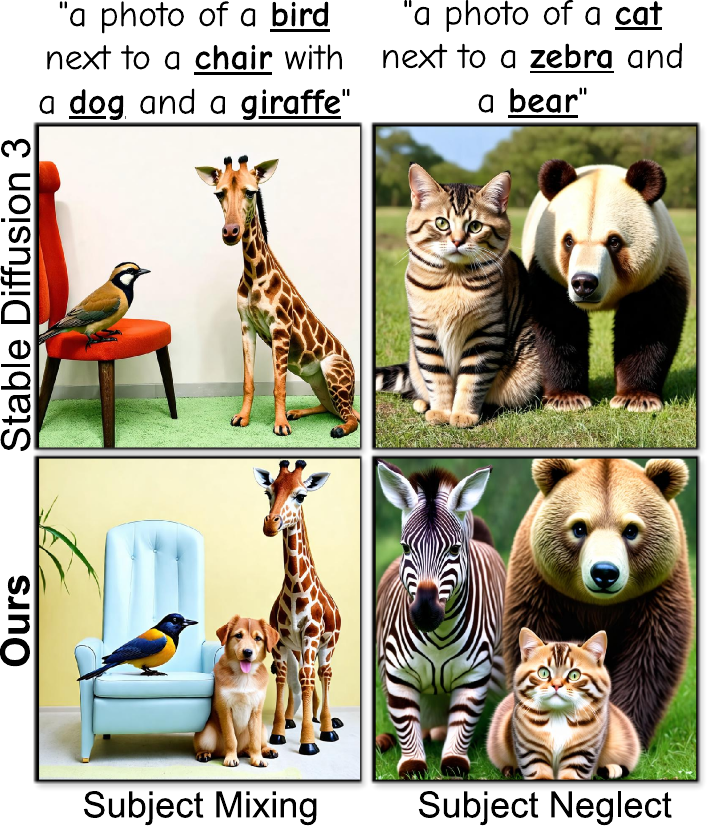}
    \includegraphics[height=0.66\linewidth]{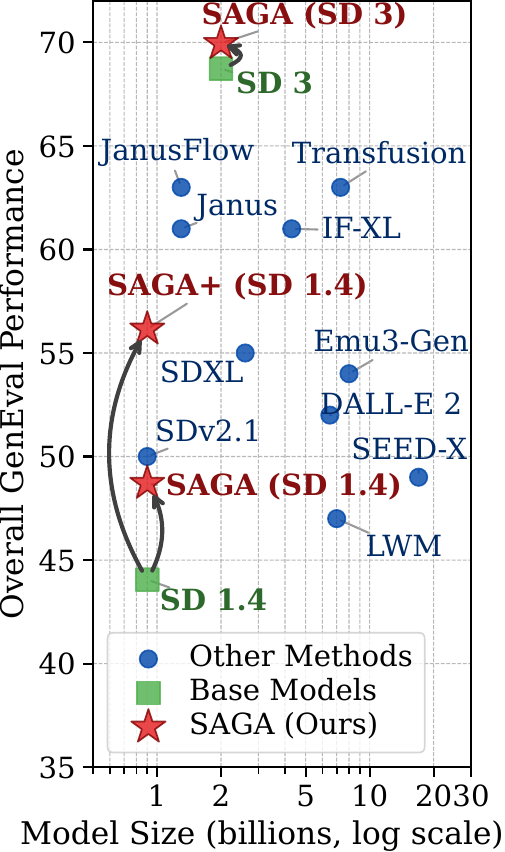}
    \caption{(Left) Text-image alignment issues on \sdthree vs. Ours. Subject mixing: the giraffe's features are incorrectly applied to the dog for \sdthree. Catastrophic neglect: the zebra is missing for \sdthree. (Right) GenEval performance of \saga against models.}
    \label{fig:illu_problem}
\end{figure}

\begin{figure*}[t]
    \centering
    \subfloat[Alignment with GSN \cite{chefer2023attendandexcite}\label{fig:interp_latent_GSN}]{
        \includegraphics[width=0.32\textwidth]{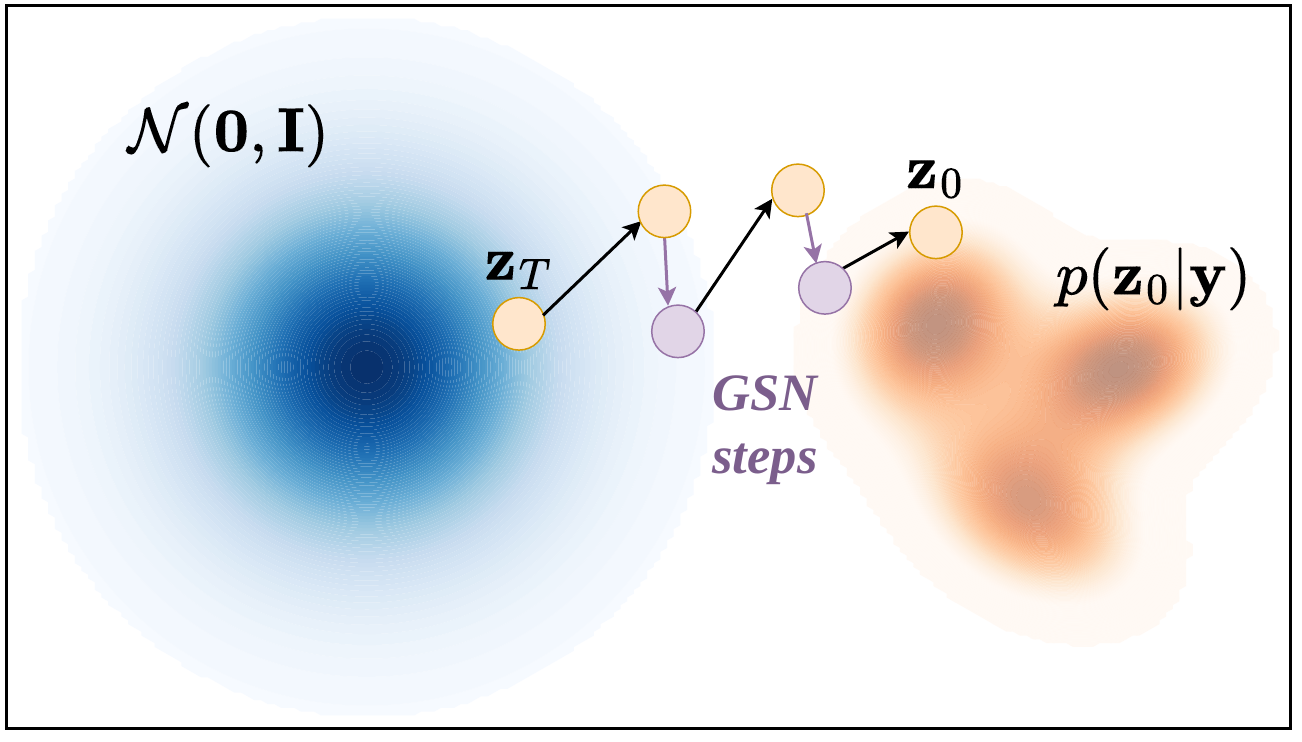}
    }\hfill
    \subfloat[Alignment with InitNO \cite{guo2024initno}\label{fig:interp_latent_InitNO}]{
        \includegraphics[width=0.32\textwidth]{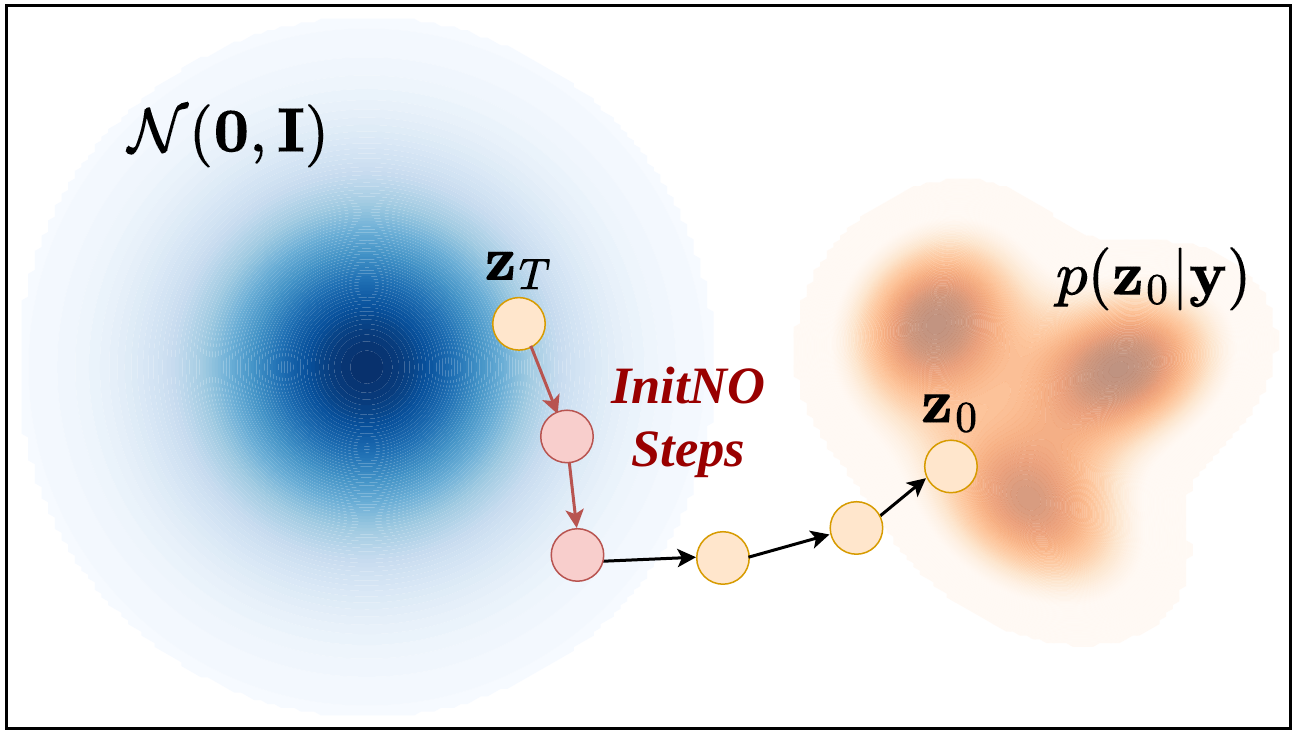}
    }\hfill
    \subfloat[Alignment with \textbf{SAGA}\label{fig:interp_latent_ours}]{
        \includegraphics[width=0.32\textwidth]{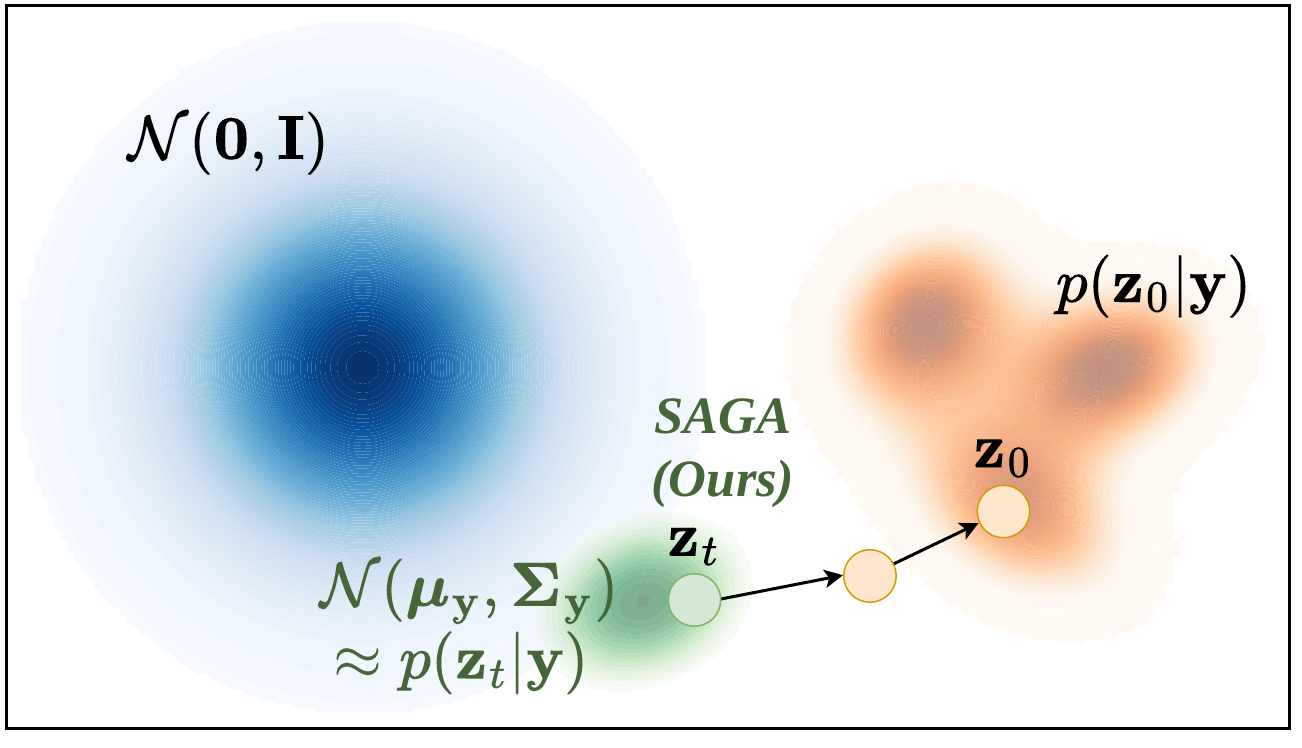}
    }
    \caption{Comparison of different alignment approaches, aiming to generate samples from a distribution $p(\mathbf{z}_0|\mathbf{y})$ for a given prompt $\mathbf{y}$. While GSN (\ref{fig:interp_latent_GSN}) corrects the latents at specific timesteps and InitNO (\ref{fig:interp_latent_InitNO}) optimizes the initialization in the prior Gaussian distribution at $t=T$, our SAGA approach (\ref{fig:interp_latent_ours}) directly samples from a conditional Gaussian prior that approximates $p(\mathbf{z}_t|\mathbf{y})$ for a timestep $t<T$.}

    \label{fig:interp_latent_all}
\end{figure*}

To mitigate these issues, some approaches refine training data by improving captions~\cite{chen2024pixartsigma,segalis2023picture} or explore novel architectures~\cite{Peebles2022DiT}. They nevertheless require retraining the entire model, which is computationally demanding. Alternatively, training-free methods based on inference-time optimization have been proposed to guide generation without requiring model retraining, relying on external knowledge from other models to guide generation~\cite{yu2023freedom, bansal2024universal}. They extend the classifier guidance framework~\cite{dhariwal2021diffusion}, which uses a time-dependent classifier, by generalizing it to arbitrary networks in a time-independent manner. This is achieved by estimating the final image and using it to steer the denoising process.

However, adjusting the latent representation with an external model may lead to inconsistencies in the denoising process. An alternative is
Generative Semantic Nursing (\gsn) \cite{rassin2023linguistic, Agarwal_2023_ICCV, chefer2023attendandexcite, li2023divide, guo2024initno}, which modifies the latent image during the denoising process to satisfy a given criterion. However, this optimization process can produce out-of-distribution samples~\cite{guo2024initno}, as the latent updates are driven solely by the criterion without theoretical guarantees on distribution preservation.

We propose \textit{SAGA, Signal-Aligned Gaussian Approximation}, a new approach that generalizes the \gsn framework to distribution learning within diffusion~\cite{ho2020denoising} and flow matching~\cite{lipman2023flow} models, enhancing T2I alignment. While GSN approaches shift an existing latent $\mathbf{z}_t$ towards a point better aligned with a prompt $\mathbf{y}$, our method learns a distribution $p(\mathbf{z}_t|\mathbf{y}) \approx \mathcal{N}\left(\boldsymbol{\mu}_\mathbf{y}, \boldsymbol{\Sigma}_\mathbf{y}\right)$ from which one can sample latents $\mathbf{z}_t$ that are well-aligned with the prompt $\mathbf{y}$.
The mean vector $\boldsymbol{\mu}_\mathbf{y}$ explicitly captures the central \textit{signal} component, providing superior control over the generative process and limiting saturated, out-of-distribution samples. In addition, through extensive experimentation, we show that our method leads to improved sample generation with better adherence to the intended semantics, establishing its superiority over state-of-the-art approaches in generating semantically accurate images.

Another key advantage is that once this distribution is learned, we can efficiently sample multiple examples without requiring a separate optimization process for each $\mathbf{z}_t$, as is necessary in GSN methods. This is particularly advantageous for real-world applications that often generate several synthetic images internally before selecting the best candidate via downstream selection mechanisms. While many state-of-the-art approaches rely on an external model to guide the sampling, our approach relies solely on the model's internal knowledge, ensuring consistency in the denoising process while remaining training-free and seamlessly integrable with existing methods. Furthermore, it can incorporate other conditioning modalities, such as bounding boxes, to refine spatial alignment and improve generation quality further.

The contributions of the paper are as follows:
\begin{itemize}
    \item We propose a novel framework for learning distributions designed to capture the underlying manifold of high-fidelity, prompt-aligned latents for text-to-image generation, ensuring compatibility with both diffusion and flow matching frameworks.
    \item Extensive quantitative and qualitative evaluations demonstrate that our method achieves state-of-the-art text-to-image alignment. This superiority manifests at both the distribution level and on a per-sample basis, highlighting the advantages of our learning framework.
\end{itemize}

\section{Related Work}
\paragraph{Guided Generation}

Classifier-Free Guidance (CFG)~\cite{ho2022classifierfree} improves text alignment by interpolating between unconditional and text-conditioned predictions. However, achieving precise adherence to prompts remains challenging. Some works introduce trainable modules that impose additional constraints on frozen models to enhance controllability. For instance, methods such as GLIGEN~\cite{li2023gligen}, T2I-Adapter~\cite{mou2023t2i}, and ControlNet~\cite{zhang2023adding} integrate external conditions such as bounding boxes or sketches to refine the generation process.
Another direction leverages external models to guide generation. Classifier guidance~\cite{dhariwal2021diffusion} involves training a classifier on noisy images and using its predictions to adjust the latent image during denoising. To circumvent the need for training an external model on noisy data, some approaches~\cite{yu2023freedom,bansal2024universal} instead leverage pretrained models on clean images and apply similar optimization techniques to a blurred estimate of the final image, inferred from a noisy input. However, these methods rely on external knowledge without explicitly accounting for how the denoising model perceives the guidance signal internally. Since there is no guarantee that the model interprets the signal in the same way as the external guidance modifies it, these approaches may fail to enforce precise alignment. Prior studies~\cite{hertz2022prompttopromptimageeditingcross,tang2023daam} have demonstrated that diffusion models encode meaningful semantic relationships between textual and visual features through the attention mechanism, enabling both interpretability and control over the generation process.
\gsn, introduced by \citeauthor{chefer2023attendandexcite}, exploits this property to refine the latent image during the denoising process, enhancing adherence to the target prompt without requiring model retraining. As illustrated in \citefigure{fig:interp_latent_GSN}, this approach extracts attention maps and optimizes an objective formulated as a loss function, which subtly adjusts the latent representation to improve semantic alignment with the prompt. Various loss formulations have been explored~\cite{chefer2023attendandexcite,rassin2023linguistic,li2023divide,Agarwal_2023_ICCV} to enhance text-image consistency, \eg, ensuring the presence of all referenced entities or mitigating their overlap.

\paragraph{Coarse-to-fine Generation and Signal Control}
Previous works~\cite{park2023understanding, rissanen2023generative} demonstrated the coarse-to-fine nature of diffusion models, where low-frequency structures (the main scene) are reconstructed first, followed by finer details. Others~\cite{balaji2023ediffi, park2023understanding} further highlighted that textual guidance has the strongest influence at the early stages of generation, where the core elements of the scene are formed, making semantic information crucial.
\citeauthor{Lin_2024_WACV} emphasized the phenomenon of signal leakage, which may explain why certain types of noise perform better than others~\cite{Grimal_2024_WACV}. This leakage can be further exploited~\cite{Everaert_2024_WACV} to control generated images' style, brightness, and colors. More recently, the concept of \textit{winning tickets} was introduced~\cite{mao2024theLottery}, which corresponds to specific pixel blocks in the initial latent space that are particularly effective for generating certain concepts. By analyzing how the model perceives these pixel blocks and organizing them into bounding boxes per entity, they construct a latent representation that better aligns with the user’s intended scene, enabling finer control over object placement in the final image. InitNO~\cite{guo2024initno} focuses on the issues of under-optimization and over-optimization in the \gsn~approach due to adjustments across multiple denoising steps. As shown in \citefigure{fig:interp_latent_InitNO}, they propose optimizing the initial noise at the first sampling step (at $t=T$) to ensure a more effective latent initialization, facilitating better image generation while preserving the original noise distribution. This approach samples from specific regions of the prior Gaussian distribution that minimize the risk of generating out-of-distribution samples. Our method enables learning a valid distribution that outperforms InitNO while effectively conditioning the latent representations based on a given prompt. Additionally, it supports conditioning with bounding boxes, similar to the lottery ticket hypothesis.

\section{Preliminaries}
\paragraph{Unified View of Generative Modeling}\label{sec:unified_view}
\newcommand{\psource}{p_{\text{source}}}
\newcommand{\ptarget}{p_{\text{target}}}

Diffusion models~\cite{ho2020denoising} and flow matching~\cite{lipman2023flow} are generative frameworks that learn to map a simple source distribution $\psource$ to a complex target distribution $\ptarget$ (distribution of images in our case). Both rely on a forward process that progressively corrupts data over time $t$, and a reverse process that learns to reconstruct the original data. We focus on the variance-preserving diffusion in \sdone~\cite{rombach2021highresolution} and the linear flow model in \sdthree~\cite{esser2024scalingrectifiedflowtransformers}. Both operate in a latent space, where an encoder $\mathcal{E}$ maps an image $\mathbf{x}_0$ to a latent code $\mathbf{z}_{0}=\mathcal{E}(\mathbf{x}_0)$, and a decoder $\mathcal{D}$ reconstructs it.

These approaches construct a probabilistic path from $\psource$ to $\ptarget$. The forward process is defined by a conditional Gaussian distribution $p_t(\mathbf{z}_t\mid{}\mathbf{z}_0) = \mathcal{N}(\mathbf{z}_t; a_t \mathbf{z}_0, b_t^2 \mathbf{I})$ governed by the noise schedule of continuous, positive functions $a_t, b_t$ on $t \in [0, T]$. This defines a trajectory of latent variables $\mathbf{z}_t$ for $t \in [0, T]$, where $t=0$ corresponds to the clean data latent $\mathbf{z}_0$ and $t=T$ to pure noise. Using the reparameterization trick, this path is written as $ \mathbf{z}_t = a_t \mathbf{z}_0 + b_t \boldsymbol{\epsilon}$, with $\boldsymbol{\epsilon} \sim \mathcal{N}( \mathbf{0}, \mathbf{I})$.  This process can be interpreted as interpolating the original signal $\mathbf{z}_0$ and noise $\boldsymbol{\epsilon}$, where $b_t$ gradually amplifies the noise part and obscures the signal, while $a_t$ attenuates the signal component.

For conditional generation, a text prompt $\mathbf{y}$ is mapped to a conditioning vector $\mathbf{c}$ via an encoder function $f$, such that $\mathbf{c} = f(\mathbf{y})$. The specific encoders $f$ are CLIP~\cite{radford2021learning} for \sdone, and a combination of CLIP and T5~\cite{raffel2020exploring} for \sdthree. The training objectives then diverge. Diffusion models are trained to predict the noise component via a network $\epsilon_\theta(\mathbf{z}_t, \mathbf{c}, t)$. In contrast, linear flow models learn the velocity field $v_\theta(\mathbf{z}_t, \mathbf{c}, t)$ that governs the dynamics. For the linear path from $\mathbf{z}_0$ to $\boldsymbol{\epsilon}$, this target velocity is simply the direction vector $u_t(\mathbf{z}_t \mid \mathbf{z}_0) = \boldsymbol{\epsilon} - \mathbf{z}_0$. Once trained, we can sample new images by using the model's outputs.

\paragraph{Sampling and Data Estimation}

During inference, a latent $\mathbf{z}_T$ is sampled from $\psource$ $\mathcal{N}(\mathbf{0}, \mathbf{I})$ and is iteratively denoised by stepping backward in time using a numerical solver $\mathcal{T}(\cdot)$. In practice, $\mathcal{T}(\cdot)$ leverages model outputs, with the time discretized into fewer sampling steps to accelerate denoising and then reducing computation. We denote iterative denoising as $\mathbf{z}_{t-\Delta} = \mathcal{T}(\mathbf{z}_t)$, where $\Delta$ is the sampling step interval. Note that we can construct an estimator for the expected value of $\mathbf{z}_0$ at any time step $t$ according to $\mathbf{z}_t$, thus the signal in $\mathbf{z}_t$.
The estimator is defined as:
\begin{subnumcases}{\hat{\mathbf{z}}_0(\mathbf{z}_t, \mathbf{c}, t) = \hat{\mathbb{E}}[\mathbf{z}_0\mid \mathbf{z}_t, \mathbf{c}] =}
    \frac{(\mathbf{z}_t - b_t\epsilon_\theta( \mathbf{z}_t, \mathbf{c}, t))}{a_t} \label{eq:estimator_diffusion} \\
    \frac{\mathbf{z}_t - b_t v_\theta(\mathbf{z}_t, \mathbf{c}, t)}{a_t + b_t} \label{eq:estimator_flow}
\end{subnumcases}
where \eqref{eq:estimator_diffusion} is used for diffusion and \eqref{eq:estimator_flow} for linear flow. Then, a mean image can be recovered as $\hat{\mathbf{x}}_0 = \mathcal{D}(\hat{\mathbf{z}}_0)$.

\paragraph{Attention-based Text Conditioning.}

Recent advances in image generation~\cite{rombach2021highresolution,podell2023sdxl,saharia2022photorealistic,chen2024pixartsigma,balaji2023ediffi,esser2024scalingrectifiedflowtransformers} mostly condition the model on the prompt $\mathbf{y}$ by leveraging attention mechanisms. In \gsn methods, the attention maps are used to adjust the generation. Given a prompt containing a set of subject tokens $\mathcal{S} = \{s_1, \dots, s_k\}$, the corresponding attention maps $\mathbf{M}^s$ are extracted for each subject token $s$. A \gsn criterion $\mathcal{L}$ is applied to guide the generation process by adjusting the latent image representation $\mathbf{z}_t$ through a gradient-based update $\mathbf{z}_t \leftarrow \mathbf{z}_t - \alpha_t \cdot \nabla_{\mathbf{z}_t} \mathcal{L}(\mathbf{M})$, where $\alpha_t$ controls the step size of the modification. This adjustment helps align the generated image with the semantic information captured in the attention maps. This process of repeatedly adjusting $\mathbf{z}_t$ across multiple denoising steps is what we term \emph{\gsn guidance}.

\section{Learning Signal-Aligned Distributions}

\subsection{Methodology and Formulation}

\newcommand{\xlearned}{\boldsymbol{\tilde{\mu}}_\mathbf{y}}
\newcommand{\pseudoimage}{\mathbf{\hat{x}}_{0\mid t}}

We aim to generate a final image $\mathbf{x}_0$ that is aligned with a given prompt $\mathbf{y}$. As noted by previous studies~\cite{hertz2022prompttopromptimageeditingcross,chefer2023attendandexcite}, the alignment of both can be assessed by taking a latent $\mathbf{z}_t$ for high values of $t$ and using the cross-attention maps of the model between $\mathbf{z}_t$ and $\mathbf{y}$.
While GSN (\citefigure{fig:interp_latent_GSN}) shifts latents at specific timesteps during denoising and InitNO (\citefigure{fig:interp_latent_InitNO}) optimizes the initialization in the prior Gaussian distribution, our SAGA approach (\citefigure{fig:interp_latent_ours}) takes a different perspective by directly sampling from a conditional Gaussian prior that approximates $p(\mathbf{z}_t|\mathbf{y})$ for an intermediate timestep $t<T$, when the signal is partially formed but still retains stochasticity. This formulation offers several advantages: (1) it ensures samples remain in-distribution by explicitly modeling the conditional probability, (2) it allows efficient generation of multiple aligned samples without repeated optimizations, and (3) it provides direct control over the signal component, enabling more principled optimization. Our approach consists of three key components. First, we formulate a learnable conditional distribution $q(\mathbf{z}_t \vert\mathbf{y})$ that approximates the true data distribution $p(\mathbf{z}_t|\mathbf{y})$. Second, we develop an optimization procedure using gradient descent to refine this distribution. Third, we leverage the  estimator of the expected value $\mathbf{z}_0$
to provide a natural initialization for this optimization, significantly enhancing the efficiency of our approximation of $p(\mathbf{z}_t|\mathbf{y})$.

\textbf{Approximating $p(\mathbf{z}_t|\mathbf{y})$}. Let us consider a visual generative model in which the forward process can be completely defined by $a_t$ and $b_t$, as explained in~\citesection{sec:unified_view}. First, we compute an expansion of $p(\mathbf{z}_t|\mathbf{y})$ with respect to $a_t$ and $b_t$, given by Proposition \ref{theorem:expansion} (proof in \citeappendix{sec:proof_gaussian_approach}).

\begin{restatable}{prop}{expansion}
    \label{theorem:expansion}
    Consider a generative model that produces latent representations $\mathbf{z}_0$ conditioned on a prompt $\mathbf{y}$. In the forward process, the latents are noised according to $\mathbf{z}_t = a_t\mathbf{z}_0 + b_t \boldsymbol{\epsilon}, \ \boldsymbol{\epsilon} \sim \mathcal{N}(\mathbf{0},\mathbf{I})$ where $a_t$ and $b_t$ are continuous, positive-valued functions defined on $t \in [0, T]$. Let
    $\boldsymbol{\mu}_\mathbf{y} = \mathbb{E}[\mathbf{z}_0|\mathbf{y}]$ and $\boldsymbol{\Sigma}_\mathbf{y} = \Var(\mathbf{z}_0|\mathbf{y})$. If $\underset{t\rightarrow T}{\lim}(a_t) = 0$, $\underset{t\rightarrow T}{\lim}(b_t) \neq 0$, and the cumulants of $\mathbf{z}_0|\mathbf{y}$ are finite, then:
    \begin{equation}
        p(\mathbf{z}_t|\mathbf{y}) = \mathcal{N}\left(\mathbf{z}_t; a_t \boldsymbol{\mu}_\mathbf{y}, a^2_t \boldsymbol{\Sigma}_\mathbf{y} + b^2_t \mathbf{I}\right)+ \underset{t\rightarrow T}{O}(a^3_t)
    \end{equation}
\end{restatable}

Based on this asymptotic behavior, we choose to parametrize $q(\mathbf{z}_t|\mathbf{y})$ as a Gaussian density:
\begin{equation}
    q(\mathbf{z}_t|\mathbf{y}, \boldsymbol{\tilde{\mu}}_\mathbf{y}, \boldsymbol{\tilde{\Sigma}}_\mathbf{y}) = \mathcal{N}(\mathbf{z}_t; a_t\boldsymbol{\tilde{\mu}}_\mathbf{y}, a_t^2\boldsymbol{\tilde{\Sigma}}_\mathbf{y}+b_t^2 \mathbf{I})
\end{equation}
where we aim to optimize $\boldsymbol{\tilde{\mu}}_\mathbf{y}$ and $\boldsymbol{\tilde{\Sigma}}_\mathbf{y}$.

\textbf{Optimizing $q(\mathbf{z}_t|\mathbf{y}, \boldsymbol{\tilde{\mu}}_\mathbf{y}, \boldsymbol{\tilde{\Sigma}}_\mathbf{y})$}. Since $\boldsymbol{\mu}_\mathbf{y} = \mathbb{E}[\mathbf{z}_0|\mathbf{y}]$ is intractable, we cannot directly optimize $\boldsymbol{\tilde{\mu}}_\mathbf{y}$. However, we hypothesize we have a criterion $\mathcal{L}(\mathbf{z}_t, \mathbf{y})$ that measures misalignment between samples drawn from $p(\mathbf{z}_t|\mathbf{y})$ and the prompt $\mathbf{y}$. By drawing samples $\mathbf{z}_t$ from $q(\mathbf{z}_t|\mathbf{y}, \boldsymbol{\tilde{\mu}}_\mathbf{y}, \boldsymbol{\tilde{\Sigma}}_\mathbf{y})$ and minimizing $\mathcal{L}(\mathbf{z}_t, \mathbf{y})$, we indirectly optimize $q(\mathbf{z}_t|\mathbf{y}, \boldsymbol{\tilde{\mu}}_\mathbf{y}, \boldsymbol{\tilde{\Sigma}}_\mathbf{y})$ to be close to $p(\mathbf{z}_t|\mathbf{y})$. More precisely, we solve:

\begin{equation}
    \boldsymbol{\tilde{\mu}}_\mathbf{y}^*, \boldsymbol{\tilde{\Sigma}}_\mathbf{y}^* = \underset{\boldsymbol{\tilde{\mu}}_\mathbf{y}, \boldsymbol{\tilde{\Sigma}}_\mathbf{y}}{\arg\min} \ \mathbb{E}_{q(\mathbf{z}_t|\mathbf{y}, \boldsymbol{\tilde{\mu}}_\mathbf{y}, \boldsymbol{\tilde{\Sigma}}_\mathbf{y})}[\mathcal{L}(\mathbf{z}_t, \mathbf{y}, t)]
\end{equation}

\begin{figure}
    \centering
    \includegraphics[width=\linewidth]{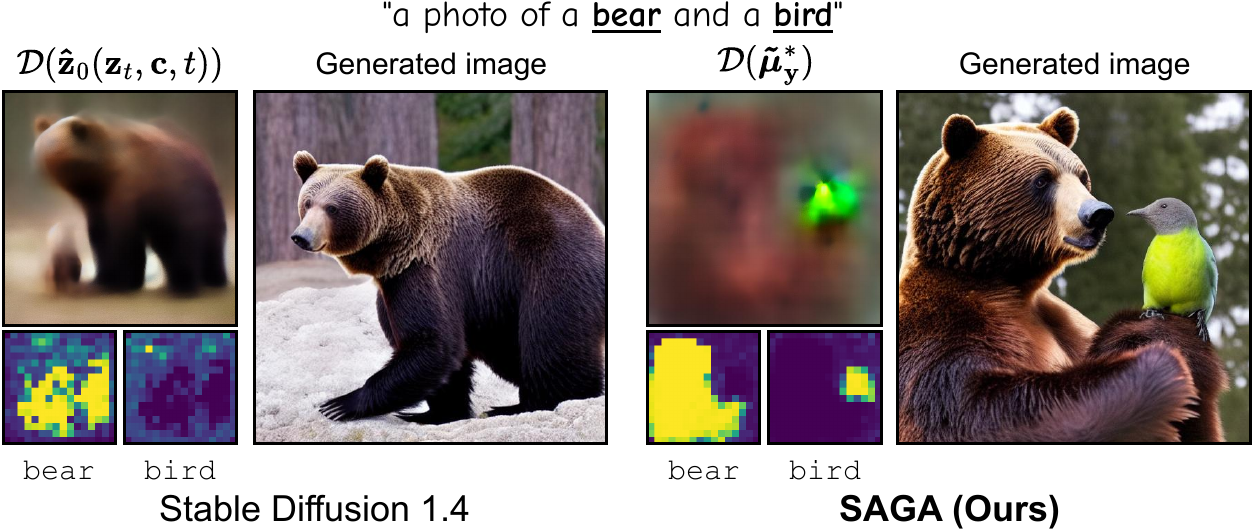}
    \caption{Only one of the cross-attention maps between the entities \texttt{bear} and \texttt{bird} (bottom-left) with the considered diffused latent $\mathbf{z}_t$ (illustrated at top-left by the corresponding final image estimation $\hat{\mathbf{z}}_0(\mathbf{z}_t, \mathbf{c}, t)$) is active before the process (\textit{left}, with the standard \sdone) while both maps are active after our optimization procedure (\textit{right}, with \saga).}
    \label{fig:illu}
\end{figure}

\begin{algorithm}[t]
    \caption{Find optimal $\boldsymbol{\tilde{\mu}}_\mathbf{y} \approx \mathbb{E}[\mathbf{z}_0|\mathbf{y}]$} \label{alg:method}
    \DontPrintSemicolon
    \KwInput{$t$, a prompt $\mathbf{y}$, $N$ steps of optimization, an optimization step size $\alpha$, noise scheduler parameters $a_t, b_t$}
    \KwOutput{ $\boldsymbol{\tilde{\mu}}_\mathbf{y}$ s.t. $q(\mathbf{z}_t|\mathbf{y}, \boldsymbol{\tilde{\mu}}_\mathbf{y})$ approximates $p(\mathbf{z}_t|\mathbf{y})$.}
    Initialize $\boldsymbol{\tilde{\mu}}_\mathbf{y}$

    \For{$i = 1$ to $N$}
    {
        $\boldsymbol{\epsilon} \sim \mathcal{N}(\mathbf{0}, \mathbf{I})$

        $\mathbf{z}_t = a_t \boldsymbol{\tilde{\mu}}_\mathbf{y} + b_t\boldsymbol{\epsilon}$

        $\boldsymbol{\tilde{\mu}}_\mathbf{y} \leftarrow \boldsymbol{\tilde{\mu}}_\mathbf{y} - \alpha \nabla_{\boldsymbol{\tilde{\mu}}_\mathbf{y}}\mathcal{L}(\mathbf{z}_t, \mathbf{y},t)$
    }
    \textbf{return} $\boldsymbol{\tilde{\mu}}_\mathbf{y}$
\end{algorithm}

This proposition mathematically confirms that at early stages of generation, the distribution of latents $\mathbf{z}_t$ conditioned on a prompt $\mathbf{y}$ can be well-approximated by a simple Gaussian. This insight allows us to simplify a complex problem into learning the mean and covariance of a Gaussian.

In practice, for high values of $t$, we find that neglecting the $a_t^2\boldsymbol{\tilde{\Sigma}}_\mathbf{y}$ term in the variance is computationally interesting while still producing highly aligned samples with the tested prompts $\mathbf{y}$. Therefore, we introduce our primary method, \saga, which learns only the optimal mean $\boldsymbol{\tilde{\mu}}_\mathbf{y}$ while keeping the covariance fixed, as detailed in \citealgorithm{alg:method}. The variant that learns the covariance, named \sagavar, is detailed in
\citeappendix{sec:variance_details}. Finally, both approaches can be combined with GSN guidance after the distribution is learned; we denote these enhanced variants \sagaplus and \sagaplusvar, respectively.

From a signal-processing perspective, diffusion and flow models construct images in a coarse-to-fine manner, first establishing the image's foundational structures before progressively adding finer details. Our method leverages this by learning an optimal mean, $\boldsymbol{\tilde{\mu}}_\mathbf{y}$, that explicitly represents the signal, the image's low-frequency, coarse structural foundation. By isolating and optimizing this deterministic structural component, the generation process becomes more resilient to stochastic high-frequency variations, ensuring that core visual concepts are robustly aligned with the prompt (see \citeappendix{sec:appendix_interpretation} for a detailed analysis). \citefigure{fig:illu} illustrates that the learned $\boldsymbol{\tilde{\mu}}_\mathbf{y}^*$ corresponds to the coarse, foundational features that lead to a more coherent final result.

\paragraph{Initialization Strategy.}

For the gradient-based optimization to converge efficiently, a good initialization of $\boldsymbol{\tilde{\mu}}_\mathbf{y}$ is essential, particularly under a limited optimization budget. Ideally, $\boldsymbol{\tilde{\mu}}_\mathbf{y}$ should approximate $\mathbb{E}[\mathbf{z}_0|\mathbf{y}]$, which is the mean image conditioned on a prompt. For any given $t$, we can estimate $\mathbf{\hat{z}}_0(\mathbf{z}_t, \mathbf{y}, t)$. Initialization is performed using the per-channel spatial mean of $\mathbf{\hat{z}}_0(\mathbf{z}_t, \mathbf{y}, t)_{HW}$, equivalent to the zeroth-frequency coefficient (or DC component) in the Fourier domain, thus representing the average color of each channel. This provides a robust foundation for adding more precise coarse features. The initialization strategy is discussed further in \citeappendix{appendix:initialization_strategy}.

\paragraph{Sampling}

Our method provides two sampling strategies to generate $N$ images for a prompt $\mathbf{y}$. The first involves denoising a single latent $\mathbf{z}_T$ to an intermediate step $t$. This single $\mathbf{z}_t$ initializes a distribution $q(\mathbf{z}_t|\mathbf{y})$ from which we sample and fully denoise $N$ latents to obtain the final images $\mathbf{z}_0$. The second strategy, which we adopt to maximize sample diversity, starts with $N$ independent latents $\mathbf{z}_T$. Each is denoised to step $t$, and each of the $N$ resulting latents $\mathbf{z}_t$ initializes its own distinct distribution. We then draw one latent from each of these $N$ distributions and complete their denoising. While we use the latter approach, both strategies yield similar quantitative scores (see~\citeappendix{sec:evaluation_appendix}).

\subsection{Cross-modal Attention Criterion \label{sec:criterion}}

By extracting and pre-processing the attention maps of \sdone and \sdthree, we obtain $\mathbf{M}^s\in\mathbb{R}^{h\times w}$ for each subject token in the prompt $\mathcal{S} = \{s_1, \dots, s_k\}$. To enforce meaningful signal learning, we apply two \gsn criteria. The first, from \cite{chefer2023attendandexcite}, promotes the emergence of attention for each subject token in the prompt:
\begin{equation}
    \mathcal{L}_1 = \underset{s \in S}{\max} (1 - \underset{i,j}{\max}(\mathbf{M}^{s}_{i,j}))
\end{equation}
where $\mathbf{M}^s_{i,j}$ represents the attention map value at position $(i,j)$ for the subject token $s$.
The second is based on Intersection over Union (IoU), as used in \cite{guo2024initno, Agarwal_2023_ICCV}. It encourages minimal overlap between all pairs of subject tokens:
\begin{equation}
    \mathcal{L}_{2} =  \frac{1}{\vert \mathcal{C}\vert}\underset{(m, n) \in \mathcal{C}}{\sum} \left( \frac{\underset{i,j}{\sum} \min(\mathbf{M}^m_{i,j}, \mathbf{M}^n_{i,j})}{\underset{i,j}{\sum} (\mathbf{M}^m_{i,j} +\mathbf{M}^n_{i,j} )} \right)
\end{equation}
where $\mathcal{C}$ represents all possible pairs of subject tokens $(m, n)$, and $\mathbf{M}^n_{i,j}$ denotes the attention map value at position $(i, j)$ for subject token $n$. These criteria ensure a well-defined signal with sufficient attention for each subject token while minimizing overlap, guiding the optimization towards constructing an effective distribution given a prompt $\mathbf{y}$.
The final criterion is computed as $\mathcal{L} = \frac{\mathcal{L}_{1} + \mathcal{L}_{2}}{2}$. While we adopt this formulation, other criteria could have been used as well.
In~\citeappendix{sec:implementation_details}, we detail the extraction of attention features and the application of the criterion, along with additional processing steps for \sdone and \sdthree. \citefigure{fig:illu} compares the attention maps from standard \sdone with those from our method. The standard model activates only one entity's map, failing to represent both concepts simultaneously. In contrast, our optimization procedure ensures that both attention maps are properly activated, demonstrating the successful capture of all entities specified in the prompt.

\subsection{Rescaling the Signal}\label{sec:rescale_signal}

\begin{figure}[t]
    \centering
    \centering
    \includegraphics[width=0.7\linewidth]{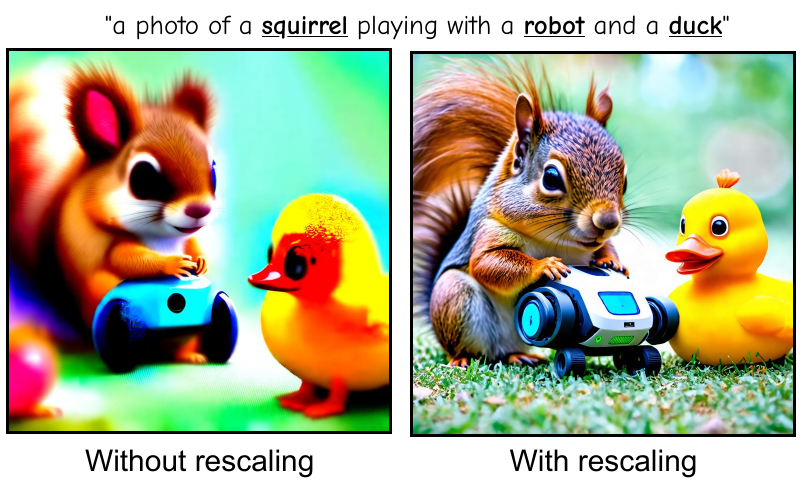}
    \caption{Effect of our rescaling mechanism on the generated images on \sdthree with \saga, hyperparameters are identical.
    }
    \label{fig:rescaling}
\end{figure}

While our Gaussian formulation for $q(\mathbf{z}_t|\mathbf{y})$ is supported by a rigorous probabilistic framework, it does not guarantee that the final image $\mathbf{x}_0$ will be valid. In practice, the optimization can lead to over-saturated outputs, depending on the hyperparameter selection; the alignment score approximates the target condition but fails to enforce a natural dynamic range. As illustrated in~\citefigure{fig:rescaling}, an image generated without further control displays recognizable content yet suffers from excessive contrast and saturation. To control this, we adopt a rescaling strategy inspired by~\cite{Lin_2024_WACV}.
After each optimization step, we rescale the learned signal $\boldsymbol{\tilde{\mu}}_\mathbf{y}$ to ensure its standard deviation does not exceed that of the $\mathbf{\hat{z}}_0$ used for initialization.
Unlike the sample-wise guidance in \gsn, this gives us direct control over the global properties of the learned signal. Further details and alternative methods are discussed in \citeappendix{appendix:rescaling}.

\section{Experimental Analysis and Results}

\subsection{Implementation Details \label{sec:evaldetail}}

\paragraph{Methods.}
We use 50 sampling steps for the \sdone{} backbone and 28 sampling steps for \sdthree{}. The distribution parameters are optimized using SGD with a learning rate of $20$ for both models. We apply 50 optimization steps for a fair comparison with \gsn approaches.
For tasks conditioned on bounding boxes, we replace the standard $\mathcal{L}_2$ loss with a dedicated criterion (detailed in~\citeappendix{sec:lossbbox}). This same criterion is then used both to learn the distribution in \saga{} and for the subsequent GSN guidance in \sagaplus{}.
We selected all hyperparameters on a validation set. Details of this set and hyperparameters, as well as further ablations, are provided in \citeappendix{sec:hyperparameters}. We present a key study on the sampling step $t$ in \citesection{sec:ablation_step}.
On \sdone, we compare various training-free approaches.
With \sdone, we compare our method to \gsn approaches Attend\&Excite~\cite{chefer2023attendandexcite}, SynGen~\cite{rassin2023linguistic}, BoxDiff~\cite{Xie_2023_ICCV}, InitNO~\cite{guo2024initno}. InitNO+ refers to the variant of InitNO incorporating a \gsn guidance.
For layout-to-image generation, we use BoxDiff~\cite{Xie_2023_ICCV}, UGD~\cite{bansal2024universal}, and Lottery Tickets~\cite{mao2024theLottery} as baselines.
On \sdthree, we compare our method to the baseline model. Comprehensive implementation details are available in \citeappendix{sec:implementation_details}

\paragraph{Benchmarks and Metrics}
The evaluation is performed on the TIAM benchmark~\cite{Grimal_2024_WACV}, a Semantic Object Accuracy method based on object detection that verifies whether the objects specified in the prompt appear in the generated image. Compositional datasets with 2, 3, and 4 entities are used, each containing 300 prompts, with 16 images generated per prompt as recommended for robust scoring. We extend the datasets by adding bounding box annotations to enable evaluation with bounding box-conditioned generation (see \citeappendix{sec:evaluation_appendix}). In addition to the TIAM score, we compute the VQA score~\cite{linvqascore2025}, an automatic metric assessing prompt-image consistency and considered more human-aligned than CLIP-based metrics, which behave like bags-of-words~\cite{yuksekgonul2023when}. A user study is also conducted. Following prior GSN work, we compute CLIP-based metrics and use the Aesthetic score~\cite{schuhmann2022laion5bopenlargescaledataset} to ensure image quality is not degraded, with the full results available in~\citeappendix{sec:evaluation_appendix}. We also evaluated our method on the GenEval benchmark~\cite{ghosh2023geneval}, which is a general-purpose evaluation not specifically designed to measure improvements in image composition.

\subsection{Quantitative results}
\label{sec:quanti}

\begin{table}[tb]
    \centering
    {
        \setlength{\tabcolsep}{1mm}
        \small
        \begin{tabular}{llrrrrrr}
            \toprule
                                                                  & \multirow{2}{*}{Methods}             & \multicolumn{3}{c}{TIAM} & \multicolumn{3}{c}{VQA Score}                                                                             \\
                                                                  &                                      & 2                        & 3                             & 4                & 2                & 3                & 4                \\
            \midrule
            \multirow[c]{9}{*}{\rotatebox{90}{\textbf{\sdone}}}   & Stable Diffusion                     & 45.4                     & 8.4                           & 1.0              & 61.3             & 31.9             & 23.5             \\
                                                                  & InitNO                               & 62.1                     & 14.2                          & 1.2              & 73.5             & 37.9             & 23.6             \\
                                                                  & \saga                                & 74.7                     & 32.3                          & 6.8              & 83.7             & 56.6             & 34.5             \\
                                                                  & \sagavar                             & 75.9                     & 37.1                          & 9.0              & 82.6             & 59.3             & 37.5             \\
            \cmidrule(lr){2-8}
                                                                  & Attend\&Excite                       & 71.4                     & 32.0                          & 10.1             & 85.7             & 65.2             & 49.8             \\
                                                                  & InitNO+                              & 75.3                     & 33.0                          & 9.8              & 87.0             & 65.0             & 48.0             \\
                                                                  & Syngen                               & 78.5                     & 39.2                          & 13.1             & 85.4             & 63.4             & 47.3             \\
                                                                  & \sagaplus                            & \underline{85.5}         & \underline{50.7}              & \underline{17.9} & \textbf{88.3}    & \underline{70.5} & \underline{51.1} \\
                                                                  & \sagaplusvar                         & \textbf{86.1}            & \textbf{52.9}                 & \textbf{19.6}    & \underline{88.1} & \textbf{71.3}    & \textbf{51.5}    \\
            \midrule
            \multirow[c]{3}{*}{\rotatebox{90}{\textbf{\sdthree}}} & Stable Diffusion                     & 84.3                     & 62.3                          & 32.2             & 90.5             & 78.6             & 65.7             \\
                                                                  & \sagaStepfiveMomentumzerodotseven    & \textbf{87.0}            & \textbf{80.0}                 & \underline{63.2} & \textbf{93.5}    & \textbf{86.4}    & \textbf{81.2}    \\
                                                                  & \sagavarStepfiveMomentumzerodotseven & \underline{86.6}         & \underline{79.8}              & \textbf{64.0}    & \underline{93.4} & \underline{86.3} & \underline{80.9} \\
            \bottomrule
        \end{tabular}
    }

    \caption{Comparison of our method against vanilla sampling and state-of-the-art training-free methods on \sdone and \sdthree (scores by number of entities in the prompt: 2, 3, 4). + denotes GSN guidance. $\Sigma$ denotes the variant with covariance learning. Best and second-best results are shown in bold and underlined, respectively. }

    \label{tab:oursandgsn}
\end{table}

\begin{table}[tb]
    \centering
    {
        \setlength{\tabcolsep}{1mm}
        \small
        \begin{tabular}{lrrrr}
            \toprule
            \multirow{2}{*}{Methods} & \multicolumn{2}{c}{TIAM} & \multicolumn{2}{c}{VQA Score}                                       \\
                                     & 2                        & 3                             & 2                & 3                \\
            \midrule
            Lottery Tickets          & 42.5                     & 8.4                           & 58.6             & 31.3             \\
            UGD                      & 38.3                     & 6.2                           & 66.0             & 42.0             \\
            BoxDiff                  & 57.4                     & 18.0                          & 78.5             & 53.3             \\
            \sagabbox                & \underline{75.1}         & \underline{34.4}              & \underline{83.5} & \underline{59.6} \\
            \sagabboxgsn             & \textbf{78.0}            & \textbf{40.5}                 & \textbf{85.4}    & \textbf{64.8}    \\
            \bottomrule
        \end{tabular}
    }
    \caption{Results of the bounding box-conditioned methods on \sdone (scores by number of entities in the prompt: 2, 3)}
    \label{tab:bbox}
\end{table}

\begin{figure*}
    \centering
    \includegraphics[width=0.9\linewidth]{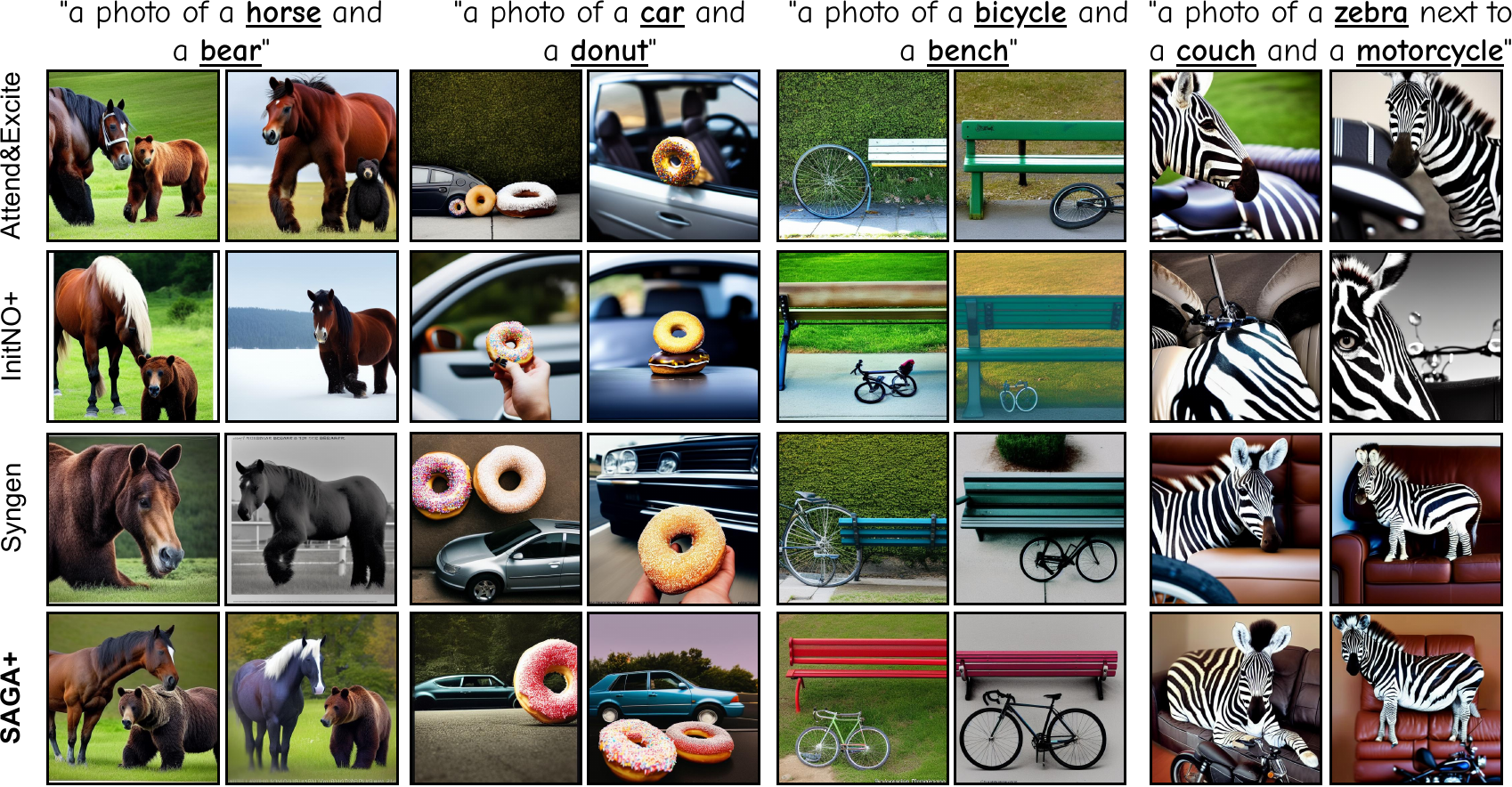}
    \caption{Generated images across different methods using \sdone. Images in the same column are generated with the same seed.}
    \label{fig:qualigsn}
\end{figure*}

\paragraph{Text Conditioning Only.}

\citetable{tab:oursandgsn} presents a comparison of different methods using TIAM and VQA scores (see~\citeappendix{sec:evaluation_appendix} for additional metrics). The results are divided into three parts. First, we compare our approach to InitNO, which applies a single intervention during generation, similar to our method.
Our approach achieves better scores. This supports our hypothesis that intervening after a foundational signal has begun to form is more effective than optimizing from pure noise.
Secondly, we compare \sagaplus with methods that employ \gsn guidance. Notably, approaches with a single intervention do not match the performance of techniques that adjust continuously during the diffusion process. Our method achieves nearly superior results across all datasets and metrics. It demonstrates excellent performance on \sdthree, yielding superior outcomes.
Learning the covariance ($\Sigma$-variants) yields a marginal improvement on \sdone but a slight performance decrease on \sdthree. We discuss this outcome in more detail in \citeappendix{sec:estim_variance}. The GenEval results, summarized in~\citefigure{fig:illu_problem} (right) and detailed in~~\citeappendix{sec:geneval}, show that on \sdone, our \sagaplus variant outperforms competing methods with comparable parameter counts. On \sdthree{}, \saga achieves the strongest results.

\paragraph{Bounding Box Conditioning.}
\citetable{tab:bbox} compares models conditioned on both text prompts and bounding boxes, where our method, \saga, outperforms all competing approaches.
The lower performance of UGD, for instance, can be attributed to its reliance on an external classifier for guidance. This external guidance may not ensure that modifications align with the model's internal knowledge. Finally, \sagaplus further boosts these results, especially for three objects, by incorporating GSN guidance alongside denoising.

\paragraph{User Study}

\begin{table}[tb]
    \centering
    {

        \setlength{\tabcolsep}{1mm}
        \small
        \npdecimalsign{.}
        \nprounddigits{0}
        \begin{tabular}{lccccccc}
            \toprule
                       & \multicolumn{4}{c}{\sdone} & \multicolumn{3}{c}{\sdthree}                                                                                                    \\
            \cmidrule(r){2-5} \cmidrule(l){6-8}
                       & \sagaplus                  & Syngen                       & InitNO+         & $\emptyset$     & \saga                    & \sdthree        & $\emptyset$     \\
            \midrule
            Semantic   & \textbf{\numprint{52.3}}   & \numprint{42.3}              & \numprint{39.6} & \numprint{17.6} & \textbf{\numprint{73.4}} & \numprint{50.5} & \numprint{9}    \\
            Preference & \textbf{\numprint{29.7}}   & \numprint{24.8}              & \numprint{24.3} & \numprint{21.2} & \textbf{\numprint{50.5}} & \numprint{29.7} & \numprint{19.8} \\
            \bottomrule
        \end{tabular}
    }
    \caption{User study results, reported as selection percentages. The $\emptyset$ symbol denotes trials where no method was chosen.} 
    \label{tab:user_study}
\end{table}

We conducted a user study with 37 participants comparing \sagaplus with top methods on \sdone (InitNO+, Syngen) and \saga with \sdthree (protocol in \citeappendix{appendix:user_study}). The study involved two distinct tasks: \textit{semantic} matching, where participants could select multiple images matching a caption or none, and \textit{preference} selection, where they could choose only their single preferred image. Users significantly prefer our method (\citetable{tab:user_study}).

\subsection{Qualitative Results}
\label{sec:quali}

\begin{figure}
    \centering
    \includegraphics[width=0.9\linewidth]{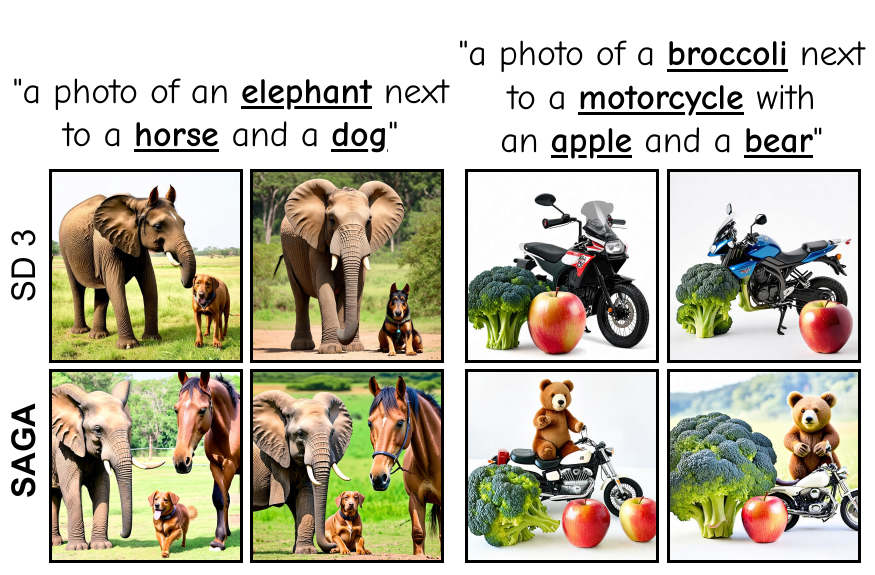}
    \caption{Images generated with \saga and \sdthree with the same seeds/prompts.}
    \label{fig:qualisd3}
\end{figure}
We used the same seeds and prompts for different methods with \gsn guidance to generate images. In \citefigure{fig:qualigsn}, the first column, only \sagaplus generates two semantically accurate images. In the last column, which features a complex prompt with three entities, only \sagaplus successfully generates the images with all the requested entities. \citefigure{fig:qualisd3} presents the results of \saga and \sdthree. Our method improves text-to-image alignment, as suggested by the more consistent depiction of all entities in response to complex prompts. Further samples are provided in \citeappendix{appendix:qualitative}.

\subsection{Ablation Study}
\label{sec:ablation_step}

\begin{figure}
    \centering
    \includegraphics[width=0.95\linewidth]{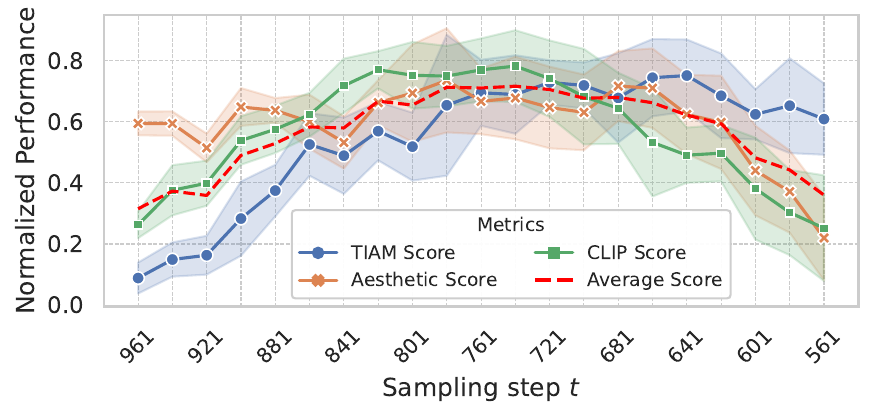}
    \caption{
        Performance of \saga on \sdone as a function of the sampling step. Each score is normalized. Points denote the performance averaged over various momentum values, with error bars indicating the standard deviation. The red curve highlights the Average Score.}
    \label{fig:study_t_sd14_main}
\end{figure}

We analyze the impact of the sampling step $t$ at which \saga{} is applied. As shown for \sdone{} in~\citefigure{fig:study_t_sd14_main}, performance peaks and then declines. This suggests a trade-off: learning the distribution too late in the denoising process is too difficult, perhaps due to the limited optimization budget or the difficulty of learning a more complex $\boldsymbol{\mu}_\mathbf{y}$. This confirms an optimal ``sweet spot'' exists. A similar analysis is conducted on \sdthree{} (see \citeappendix{appendix:sampling_step}).

\section{Limitations}

Our approach, like GSN methods, requires backpropagation through the model, which remains computationally expensive. Moreover, as it relies solely on the model's internal knowledge, the extent of correction is inherently limited. For a fair comparison with other approaches, we used only 50 optimization steps. However, we believe that additional optimization could further improve results, albeit at the cost of increased generation time.

\section{Conclusion}

We introduced a novel modeling approach to enhance image generation at inference time, offering new perspectives on controllable generation. By generalizing the GSN framework with a Gaussian-based formulation, we explicitly model the learned signal $\boldsymbol{\tilde{\mu}}_\mathbf{y}$, providing a new mechanism for guiding the generation process.
Experiments showed that the approach improves text-image alignment for both flow matching and diffusion models. Future work could explore alternative forms of control over the learning of $\boldsymbol{\tilde{\mu}}_\mathbf{y}$.

\section*{Acknowledgments}

This work was granted access to the HPC resources of IDRIS under the allocation 2022-AD011014009 made by GENCI. It relied on the use of the FactoryIA supercomputer, financially supported by the Ile-de-France Regional Council and was partly supported by the  ANR-21-CE23-0024 IDeGeN project and the ANR-23-PEIA-0008 SHARP project in the context of the France 2030 program.

{
    \small
    \bibliographystyle{ieee_fullname}
    \bibliography{ref}

@String(CVPR= {IEEE Conf. Comput. Vis. Pattern Recog.})

@String(ICCV= {Int. Conf. Comput. Vis.})

@String(ECCV= {Eur. Conf. Comput. Vis.})

@String(BMVC= {Brit. Mach. Vis. Conf.})

@String(ICLR = {Int. Conf. Learn. Represent.})

@String(AAAI = {AAAI})

@String(CVPR  = {CVPR})

@String(ICCV  = {ICCV})

@String(ECCV  = {ECCV})

@String(BMVC  =	{BMVC})

@String(ICLR  = {ICLR})

@InProceedings{Everaert_2024_WACV,
    author   = {Everaert, Martin Nicolas and Fitsios, Athanasios and Bocchio, Marco and Arpa, Sami and Süsstrunk, Sabine and Achanta, Radhakrishna},
    title    = {{E}xploiting the {S}ignal-{L}eak {B}ias in {D}iffusion {M}odels}, 
    booktitle = {Proceedings of the IEEE/CVF Winter Conference on Applications of Computer Vision (WACV)},
    month     = {January},
    year      = {2024},
    pages     = {4025-4034}
}

@InProceedings{Lin_2024_WACV,
    author    = {Lin, Shanchuan and Liu, Bingchen and Li, Jiashi and Yang, Xiao},
    title     = {Common Diffusion Noise Schedules and Sample Steps Are Flawed},
    booktitle = {Proceedings of the IEEE/CVF Winter Conference on Applications of Computer Vision (WACV)},
    month     = {January},
    year      = {2024},
    pages     = {5404-5411}
}

@article{holmquist1996multivariate_hermite_poly,
title = {The d-variate vector hermite polynomial of order k},
journal = {Linear Algebra and its Applications},
volume = {237-238},
pages = {155-190},
year = {1996},
note = {Linear Algebra and Statistics: In Celebration of C. R. Rao's 75th Birthday (September 10, 1995)},
issn = {0024-3795},
doi = {https://doi.org/10.1016/0024-3795(95)00595-1},
url = {https://www.sciencedirect.com/science/article/pii/0024379595005951},
author = {Björn Holmquist}
}

@article{chung2024flanT5,
  author  = {Hyung Won Chung and Le Hou and Shayne Longpre and Barret Zoph and Yi Tay and William Fedus and Yunxuan Li and Xuezhi Wang and Mostafa Dehghani and Siddhartha Brahma and Albert Webson and Shixiang Shane Gu and Zhuyun Dai and Mirac Suzgun and Xinyun Chen and Aakanksha Chowdhery and Alex Castro-Ros and Marie Pellat and Kevin Robinson and Dasha Valter and Sharan Narang and Gaurav Mishra and Adams Yu and Vincent Zhao and Yanping Huang and Andrew Dai and Hongkun Yu and Slav Petrov and Ed H. Chi and Jeff Dean and Jacob Devlin and Adam Roberts and Denny Zhou and Quoc V. Le and Jason Wei},
  title   = {Scaling Instruction-Finetuned Language Models},
  journal = {Journal of Machine Learning Research},
  year    = {2024},
  volume  = {25},
  number  = {70},
  pages   = {1--53},
  url     = {http://jmlr.org/papers/v25/23-0870.html}
}

@InProceedings{Grimal_2024_WACV,
    author    = {Grimal, Paul and Le Borgne, Herv\'e and Ferret, Olivier and Tourille, Julien},
    title     = {TIAM - A Metric for Evaluating Alignment in Text-to-Image Generation},
    booktitle = {Proceedings of the IEEE/CVF Winter Conference on Applications of Computer Vision (WACV)},
    month     = {January},
    year      = {2024},
    pages     = {2890-2899}
}

@InProceedings{radford2021learning,
  title = 	 {Learning Transferable Visual Models From Natural Language Supervision},
  author =       {Radford, Alec and Kim, Jong Wook and Hallacy, Chris and Ramesh, Aditya and Goh, Gabriel and Agarwal, Sandhini and Sastry, Girish and Askell, Amanda and Mishkin, Pamela and Clark, Jack and Krueger, Gretchen and Sutskever, Ilya},
  booktitle = 	 {Proceedings of the 38th International Conference on Machine Learning},
  pages = 	 {8748--8763},
  year = 	 {2021},
  editor = 	 {Meila, Marina and Zhang, Tong},
  volume = 	 {139},
  series = 	 {Proceedings of Machine Learning Research},
  month = 	 {18--24 Jul},
  publisher =    {PMLR},
  url = 	 {https://proceedings.mlr.press/v139/radford21a.html}
}

@inproceedings{ho2020denoising,
 author = {Ho, Jonathan and Jain, Ajay and Abbeel, Pieter},
 booktitle = {Advances in Neural Information Processing Systems},
 editor = {H. Larochelle and M. Ranzato and R. Hadsell and M.F. Balcan and H. Lin},
 pages = {6840--6851},
 publisher = {Curran Associates, Inc.},
 title = {Denoising {D}iffusion {P}robabilistic {M}odels},
 url = {https://proceedings.neurips.cc/paper_files/paper/2020/file/4c5bcfec8584af0d967f1ab10179ca4b-Paper.pdf},
 volume = {33},
 year = {2020}
}

@inproceedings{
lipman2023flow,
title={Flow Matching for Generative Modeling},
author={Yaron Lipman and Ricky T. Q. Chen and Heli Ben-Hamu and Maximilian Nickel and Matthew Le},
booktitle={The Eleventh International Conference on Learning Representations },
year={2023},
url={https://openreview.net/forum?id=PqvMRDCJT9t}
}

@inproceedings{rissanen2023generative,
  title={Generative modelling with inverse heat dissipation},
  author={Severi Rissanen and Markus Heinonen and Arno Solin},
  booktitle={International Conference on Learning Representations (ICLR)},
  year={2023}
}

@inproceedings{
park2023understanding,
title={Understanding the Latent Space of Diffusion Models through the Lens of Riemannian Geometry},
author={Yong-Hyun Park and Mingi Kwon and Jaewoong Choi and Junghyo Jo and Youngjung Uh},
booktitle={Thirty-seventh Conference on Neural Information Processing Systems},
year={2023},
url={https://openreview.net/forum?id=VUlYp3jiEI}
}

@inproceedings{rassin2023linguistic,
title={Linguistic Binding in Diffusion Models: Enhancing Attribute Correspondence through Attention Map Alignment},
author={Royi Rassin and Eran Hirsch and Daniel Glickman and Shauli Ravfogel and Yoav Goldberg and Gal Chechik},
booktitle={Thirty-seventh Conference on Neural Information Processing Systems},
year={2023},
url={https://openreview.net/forum?id=AOKU4nRw1W}
}

@article{chefer2023attendandexcite,
author = {Chefer, Hila and Alaluf, Yuval and Vinker, Yael and Wolf, Lior and Cohen-Or, Daniel},
title = {Attend-and-Excite: Attention-Based Semantic Guidance for Text-to-Image Diffusion Models},
year = {2023},
issue_date = {August 2023},
publisher = {Association for Computing Machinery},
address = {New York, NY, USA},
volume = {42},
number = {4},
issn = {0730-0301},
url = {https://doi.org/10.1145/3592116},
doi = {10.1145/3592116},
journal = {ACM Trans. Graph.},
month = {jul},
articleno = {148},
numpages = {10}
}

@inproceedings{li2023divide,
  title={Divide \& bind your attention for improved generative semantic nursing},
  author={Li, Yumeng and Keuper, Margret and Zhang, Dan and Khoreva, Anna},
  booktitle={34th British Machine Vision Conference 2023, {BMVC} 2023},
  year={2023}
}

@inproceedings{feng2023trainingfreestructureddiffusionguidance,
title={Training-Free Structured Diffusion Guidance for Compositional Text-to-Image Synthesis},
author={Weixi Feng and Xuehai He and Tsu-Jui Fu and Varun Jampani and Arjun Reddy Akula and Pradyumna Narayana and Sugato Basu and Xin Eric Wang and William Yang Wang},
booktitle={The Eleventh International Conference on Learning Representations },
year={2023},
url={https://openreview.net/forum?id=PUIqjT4rzq7}
}

@InProceedings{Agarwal_2023_ICCV,
    author    = {Agarwal, Aishwarya and Karanam, Srikrishna and Joseph, K J and Saxena, Apoorv and Goswami, Koustava and Srinivasan, Balaji Vasan},
    title     = {A-STAR: Test-time Attention Segregation and Retention for Text-to-image Synthesis},
    booktitle = {Proceedings of the IEEE/CVF International Conference on Computer Vision (ICCV)},
    month     = {October},
    year      = {2023},
    pages     = {2283-2293}
}

@inproceedings{guo2024initno,
    title     = {Init{NO}: {B}oosting {T}ext-to-{I}mage {D}iffusion {M}odels via {I}nitial {N}oise {O}ptimization},
    author    = {Guo, Xiefan and Liu, Jinlin and Cui, Miaomiao and Li, Jiankai and Yang, Hongyu and Huang, Di},
    booktitle = {CVPR},
    year      = {2024}
}

@article{dharmani2018multivariate,
  title={Multivariate generalized Gram--Charlier series in vector notations},
  author={Dharmani, Bhaveshkumar C},
  journal={Journal of Mathematical Chemistry},
  volume={56},
  pages={1631--1655},
  year={2018},
  publisher={Springer}
}

@inproceedings{hertz2022prompttopromptimageeditingcross,
title={Prompt-to-{P}rompt {I}mage {E}diting with {C}ross-{A}ttention {C}ontrol},
author={Amir Hertz and Ron Mokady and Jay Tenenbaum and Kfir Aberman and Yael Pritch and Daniel Cohen-or},
booktitle={The Eleventh International Conference on Learning Representations },
year={2023},
url={https://openreview.net/forum?id=_CDixzkzeyb}
}

@inproceedings{tang2023daam,
    title = "What the {DAAM}: Interpreting Stable Diffusion Using Cross Attention",
    author = "Tang, Raphael  and
      Liu, Linqing  and
      Pandey, Akshat  and
      Jiang, Zhiying  and
      Yang, Gefei  and
      Kumar, Karun  and
      Stenetorp, Pontus  and
      Lin, Jimmy  and
      Ture, Ferhan",
    booktitle = "Proceedings of the 61st Annual Meeting of the Association for Computational Linguistics (Volume 1: Long Papers)",
    year = "2023",
    url = "https://aclanthology.org/2023.acl-long.310",
}

@InProceedings{zhang2023adding,
  title={Adding {C}onditional {C}ontrol to {T}ext-to-{I}mage {D}iffusion {M}odels}, 
  author={Lvmin Zhang and Anyi Rao and Maneesh Agrawala},
  booktitle={IEEE International Conference on Computer Vision (ICCV)},
  pages     = {3836--3847},
  year={2023},
}

@article{mou2023t2i,
  title =	 {T2{I}-{A}dapter: {L}earning {A}dapters to {D}ig {O}ut {M}ore {C}ontrollable {A}bility for {T}ext-to-{I}mage {D}iffusion {M}odels},
  volume =	 38,
  url = {https://ojs.aaai.org/index.php/AAAI/article/view/28226},
  DOI =		 {10.1609/aaai.v38i5.28226},
  number =	 5,
  journal =	 {Proceedings of the AAAI Conference on Artificial
                  Intelligence},
  author =	 {Mou, Chong and Wang, Xintao and Xie, Liangbin and Wu, Yanze and Zhang, Jian and Qi, Zhongang and Shan, Ying},
  year =	 2024,
  month =	 {Mar},
  pages =	 {4296--4304}
}

@article{li2023gligen,
  title={GLIGEN: Open-Set Grounded Text-to-Image Generation},
  author={Li, Yuheng and Liu, Haotian and Wu, Qingyang and Mu, Fangzhou and Yang, Jianwei and Gao, Jianfeng and Li, Chunyuan and Lee, Yong Jae},
  journal={CVPR},
  year={2023}
}

@article{raffel2020exploring,
  author  = {Colin Raffel and Noam Shazeer and Adam Roberts and Katherine Lee and Sharan Narang and Michael Matena and Yanqi Zhou and Wei Li and Peter J. Liu},
  title   = {Exploring the {L}imits of {T}ransfer {L}earning with a {U}nified {T}ext-to-{T}ext {T}ransformer},
  journal = {Journal of Machine Learning Research},
  year    = {2020},
  volume  = {21},
  number  = {140},
  pages   = {1--67},
  url     = {http://jmlr.org/papers/v21/20-074.html}
}

@InProceedings{Xie_2023_ICCV,
    author    = {Xie, Jinheng and Li, Yuexiang and Huang, Yawen and Liu, Haozhe and Zhang, Wentian and Zheng, Yefeng and Shou, Mike Zheng},
    title     = {BoxDiff: Text-to-Image Synthesis with Training-Free Box-Constrained Diffusion},
    booktitle = {Proceedings of the IEEE/CVF International Conference on Computer Vision (ICCV)},
    year      = {2023},
    pages     = {7452-7461}
}

@InProceedings{rombach2021highresolution,
    author    = {Rombach, Robin and Blattmann, Andreas and Lorenz, Dominik and Esser, Patrick and Ommer, Bj\"orn},
    title     = {High-Resolution Image Synthesis With Latent Diffusion Models},
    booktitle = {Proceedings of the IEEE/CVF Conference on Computer Vision and Pattern Recognition (CVPR)},
    month     = {June},
    year      = {2022},
    pages     = {10684-10695}
}

@inproceedings{podell2023sdxl,
title={{SDXL}: Improving {L}atent {D}iffusion {M}odels for {H}igh-{R}esolution {I}mage {S}ynthesis},
author={Dustin Podell and Zion English and Kyle Lacey and Andreas Blattmann and Tim Dockhorn and Jonas M{\"u}ller and Joe Penna and Robin Rombach},
booktitle={The Twelfth International Conference on Learning Representations},
year={2024},
url={https://openreview.net/forum?id=di52zR8xgf}
}

@article{ramesh2022hierarchical,
      title={Hierarchical {T}ext-{C}onditional {I}mage {G}eneration with {CLIP} {L}atents}, 
      author={Aditya Ramesh and Prafulla Dhariwal and Alex Nichol and Casey Chu and Mark Chen},
      journal={arXiv 2204.06125},
      year={2022},
      eprint={2204.06125},
      archivePrefix={arXiv},
      primaryClass={cs.CV}
}

@article{balaji2023ediffi,
      title={{eDiff-I: Text-to-Image Diffusion Models with an Ensemble of Expert Denoisers}}, 
      author={Yogesh Balaji and Seungjun Nah and Xun Huang and Arash Vahdat and Jiaming Song and Qinsheng Zhang and Karsten Kreis and Miika Aittala and Timo Aila and Samuli Laine and Bryan Catanzaro and Tero Karras and Ming-Yu Liu},
      journal={arXiv 2211.01324},
      year={2023},
}

@inproceedings{saharia2022photorealistic,
    title={Photorealistic Text-to-Image Diffusion Models with Deep Language Understanding},
    author={Chitwan Saharia and William Chan and Saurabh Saxena and Lala Li and Jay Whang and Emily Denton and Seyed Kamyar Seyed Ghasemipour and Raphael Gontijo-Lopes and Burcu Karagol Ayan and Tim Salimans and Jonathan Ho and David J. Fleet and Mohammad Norouzi},
    booktitle={Advances in Neural Information Processing Systems},
    editor={Alice H. Oh and Alekh Agarwal and Danielle Belgrave and Kyunghyun Cho},
    year={2022},
    url={https://openreview.net/forum?id=08Yk-n5l2Al}
    }

@misc{chen2023pixartalpha,
      title={PixArt-$\alpha$: Fast {T}raining of {D}iffusion {T}ransformer for {P}hotorealistic {T}ext-to-{I}mage {S}ynthesis}, 
      author={Junsong Chen and Jincheng Yu and Chongjian Ge and Lewei Yao and Enze Xie and Yue Wu and Zhongdao Wang and James Kwok and Ping Luo and Huchuan Lu and Zhenguo Li},
      year={2023},
      eprint={2310.00426},
      archivePrefix={arXiv},
      primaryClass={cs.CV}
}

@inproceedings{chen2024pixartsigma,
title={PixArt-$\alpha$: Fast Training of Diffusion Transformer for Photorealistic Text-to-Image Synthesis},
author={Junsong Chen and Jincheng YU and Chongjian GE and Lewei Yao and Enze Xie and Zhongdao Wang and James Kwok and Ping Luo and Huchuan Lu and Zhenguo Li},
booktitle={The Twelfth International Conference on Learning Representations},
year={2024},
url={https://openreview.net/forum?id=eAKmQPe3m1}
}

@article{segalis2023picture,
      title={A {P}icture is {W}orth a {T}housand {W}ords: {P}rincipled {R}ecaptioning {I}mproves {I}mage {G}eneration}, 
      author={Eyal Segalis and Dani Valevski and Danny Lumen and Yossi Matias and Yaniv Leviathan},
      year={2023},
      journal={arXiv preprint arXiv:2310.16656},
      eprint={2310.16656},
      archivePrefix={arXiv},
      primaryClass={cs.CV}
}

@INPROCEEDINGS {Peebles2022DiT,
author = { Peebles, William and Xie, Saining },
booktitle = { 2023 IEEE/CVF International Conference on Computer Vision (ICCV) },
title = {{ Scalable Diffusion Models with Transformers }},
year = {2023},
pages = {4172--4182},

doi = {10.1109/ICCV51070.2023.00387},
url = {https://doi.ieeecomputersociety.org/10.1109/ICCV51070.2023.00387},
publisher = {IEEE Computer Society},
address = {Los Alamitos, CA, USA},
month ={Oct}
}

@inproceedings{yuksekgonul2023when,
title={When and Why Vision-Language Models Behave like Bags-Of-Words, and What to Do About It?},
author={Mert Yuksekgonul and Federico Bianchi and Pratyusha Kalluri and Dan Jurafsky and James Zou},
booktitle={The Eleventh International Conference on Learning Representations },
year={2023},
url={https://openreview.net/forum?id=KRLUvxh8uaX}
}

@misc{schuhmann2022laion5bopenlargescaledataset,
      title={LAION-5B: An open large-scale dataset for training next generation image-text models}, 
      author={Christoph Schuhmann and Romain Beaumont and Richard Vencu and Cade Gordon and Ross Wightman and Mehdi Cherti and Theo Coombes and Aarush Katta and Clayton Mullis and Mitchell Wortsman and Patrick Schramowski and Srivatsa Kundurthy and Katherine Crowson and Ludwig Schmidt and Robert Kaczmarczyk and Jenia Jitsev},
      year={2022},
      eprint={2210.08402},
      archivePrefix={arXiv},
      primaryClass={cs.CV},
      url={https://arxiv.org/abs/2210.08402}, 
}

@inproceedings{li2022blip,
      title={BLIP: Bootstrapping Language-Image Pre-training for Unified Vision-Language Understanding and Generation}, 
      author={Junnan Li and Dongxu Li and Caiming Xiong and Steven Hoi},
      year={2022},
      booktitle={ICML},
}

@InProceedings{esser2024scalingrectifiedflowtransformers,
  title = 	 {Scaling {R}ectified {F}low {T}ransformers for {H}igh-{R}esolution {I}mage {S}ynthesis},
  author =       {Esser, Patrick and Kulal, Sumith and Blattmann, Andreas and Entezari, Rahim and M\"{u}ller, Jonas and Saini, Harry and Levi, Yam and Lorenz, Dominik and Sauer, Axel and Boesel, Frederic and Podell, Dustin and Dockhorn, Tim and English, Zion and Rombach, Robin},
  booktitle = 	 {Proceedings of the 41st International Conference on Machine Learning},
  pages = 	 {12606--12633},
  year = 	 {2024},
  editor = 	 {Salakhutdinov, Ruslan and Kolter, Zico and Heller, Katherine and Weller, Adrian and Oliver, Nuria and Scarlett, Jonathan and Berkenkamp, Felix},
  volume = 	 {235},
  series = 	 {Proceedings of Machine Learning Research},
  month = 	 {21--27 Jul},
  publisher =    {PMLR},
  url = 	 {https://proceedings.mlr.press/v235/esser24a.html},
}

@article{mao2024theLottery,
  author    = {Jiafeng Mao, Xueting Wang and Kiyoharu Aizawa},
  title     = {The Lottery Ticket Hypothesis in Denoising: Towards Semantic-Driven Initialization},
  journal   = {ECCV},
  year      = {2024},
}

@inproceedings{linvqascore2025,
author = {Lin, Zhiqiu and Pathak, Deepak and Li, Baiqi and Li, Jiayao and Xia, Xide and Neubig, Graham and Zhang, Pengchuan and Ramanan, Deva},
title = {Evaluating Text-to-Visual Generation with Image-to-Text Generation},
year = {2024},
isbn = {978-3-031-72672-9},
publisher = {Springer-Verlag},
address = {Berlin, Heidelberg},
url = {https://doi.org/10.1007/978-3-031-72673-6_20},
doi = {10.1007/978-3-031-72673-6_20},
booktitle = {Computer Vision – ECCV 2024: 18th European Conference, Milan, Italy, September 29–October 4, 2024, Proceedings, Part IX},
pages = {366–384},
numpages = {19},
location = {Milan, Italy}
}

@article{yu2023freedom,
    title={FreeDoM: Training-Free Energy-Guided Conditional Diffusion Model},
    author={Yu, Jiwen and Wang, Yinhuai and Zhao, Chen and Ghanem, Bernard and Zhang, Jian},
    journal={Proceedings of the IEEE/CVF International Conference on Computer Vision (ICCV)},
    year={2023}
}

@inproceedings{bansal2024universal,
title={Universal Guidance for Diffusion Models},
author={Arpit Bansal and Hong-Min Chu and Avi Schwarzschild and Roni Sengupta and Micah Goldblum and Jonas Geiping and Tom Goldstein},
booktitle={The Twelfth International Conference on Learning Representations (ICLR)},
year={2024},
url={https://openreview.net/forum?id=pzpWBbnwiJ}
}

@inproceedings{dhariwal2021diffusion,
title={Diffusion Models Beat {GAN}s on Image Synthesis},
author={Prafulla Dhariwal and Alexander Quinn Nichol},
booktitle={Advances in Neural Information Processing Systems},
editor={A. Beygelzimer and Y. Dauphin and P. Liang and J. Wortman Vaughan},
year={2021},
url={https://openreview.net/forum?id=AAWuCvzaVt}
}

@inproceedings{ho2022classifierfree,
title={Classifier-{F}ree {D}iffusion {G}uidance},
author={Jonathan Ho and Tim Salimans},
booktitle={NeurIPS 2021 Workshop on Deep Generative Models and Downstream Applications},
year={2021},
url={https://openreview.net/forum?id=qw8AKxfYbI}
}

@inproceedings{
ghosh2023geneval,
title={GenEval: An object-focused framework for evaluating text-to-image alignment},
author={Dhruba Ghosh and Hannaneh Hajishirzi and Ludwig Schmidt},
booktitle={Thirty-seventh Conference on Neural Information Processing Systems Datasets and Benchmarks Track},
year={2023},
url={https://openreview.net/forum?id=Wbr51vK331}
}

@misc{ma2025janusflowharmonizingautoregressionrectified,
      title={JanusFlow: Harmonizing Autoregression and Rectified Flow for Unified Multimodal Understanding and Generation}, 
      author={Yiyang Ma and Xingchao Liu and Xiaokang Chen and Wen Liu and Chengyue Wu and Zhiyu Wu and Zizheng Pan and Zhenda Xie and Haowei Zhang and Xingkai yu and Liang Zhao and Yisong Wang and Jiaying Liu and Chong Ruan},
      year={2025},
      eprint={2411.07975},
      archivePrefix={arXiv},
      primaryClass={cs.CV},
      url={https://arxiv.org/abs/2411.07975}, 
}

@article{2024llamagen,
  title={Autoregressive Model Beats Diffusion: {LLaMA} for Scalable Image Generation},
  author={Sun, Peize and Jiang, Yi and Chen, Shoufa and Zhang, Shilong and Peng, Bingyue and Luo, Ping and Yuan, Zehuan},
  journal={arXiv preprint arXiv:2406.06525},
  year={2024}
}

@misc{2023IF,
    title={{DeepFloyd IF}},
    author={DeepFloyd},
    url={https://huggingface.co/DeepFloyd/IF-I-XL-v1.0},
    year={2023}
}

@article{2024emu3,
  title={Emu3: Next-token prediction is all you need},
  author={Wang, Xinlong and Zhang, Xiaosong and Luo, Zhengxiong and Sun, Quan and Cui, Yufeng and Wang, Jinsheng and Zhang, Fan and Wang, Yueze and Li, Zhen and Yu, Qiying and others},
  journal={arXiv preprint arXiv:2409.18869},
  year={2024}
}

@article{2024SeedX,
  title={{SEED-X}: Multimodal models with unified multi-granularity comprehension and generation},
  author={Ge, Yuying and Zhao, Sijie and Zhu, Jinguo and Ge, Yixiao and Yi, Kun and Song, Lin and Li, Chen and Ding, Xiaohan and Shan, Ying},
  journal={arXiv preprint arXiv:2404.14396},
  year={2024}
}

@article{2024Chameleon,
  title={Chameleon: Mixed-modal early-fusion foundation models},
  author={Team, Chameleon},
  journal={arXiv preprint arXiv:2405.09818},
  year={2024}
}

@article{2024Showo,
  title={Show-o: One single transformer to unify multimodal understanding and generation},
  author={Xie, Jinheng and Mao, Weijia and Bai, Zechen and Zhang, David Junhao and Wang, Weihao and Lin, Kevin Qinghong and Gu, Yuchao and Chen, Zhijie and Yang, Zhenheng and Shou, Mike Zheng},
  journal={arXiv preprint arXiv:2408.12528},
  year={2024}
}

@article{2024Transfusion,
  title={Transfusion: Predict the next token and diffuse images with one multi-modal model},
  author={Zhou, Chunting and Yu, Lili and Babu, Arun and Tirumala, Kushal and Yasunaga, Michihiro and Shamis, Leonid and Kahn, Jacob and Ma, Xuezhe and Zettlemoyer, Luke and Levy, Omer},
  journal={arXiv preprint arXiv:2408.11039},
  year={2024}
}

@article{2024LWM,
  title={World model on million-length video and language with ringattention},
  author={Liu, Hao and Yan, Wilson and Zaharia, Matei and Abbeel, Pieter},
  journal={arXiv preprint arXiv:2402.08268},
  year={2024}
}

@article{2023dalle3,
  title={Improving image generation with better captions},
  author={Betker, James and Goh, Gabriel and Jing, Li and Brooks, Tim and Wang, Jianfeng and Li, Linjie and Ouyang, Long and Zhuang, Juntang and Lee, Joyce and Guo, Yufei and others},
  journal={Computer Science},
  year={2023}
}

@article{2024Janus,
  title={Janus: Decoupling Visual Encoding for Unified Multimodal Understanding and Generation},
  author={Wu, Chengyue and Chen, Xiaokang and Wu, Zhiyu and Ma, Yiyang and Liu, Xingchao and Pan, Zizheng and Liu, Wen and Xie, Zhenda and Yu, Xingkai and Ruan, Chong and others},
  journal={arXiv preprint arXiv:2410.13848},
  year={2024}
}

@inproceedings{2023SDXL,
  title={{SDXL}: Improving latent diffusion models for high-resolution image synthesis},
  author={Podell, Dustin and English, Zion and Lacey, Kyle and Blattmann, Andreas and Dockhorn, Tim and M{\"u}ller, Jonas and Penna, Joe and Rombach, Robin},
  booktitle=iclr,
  year={2024}
}

@article{Ruderman1994statisticnaturalimage,
  title     = {Statistics of natural images: Scaling in the woods},
  author    = {Ruderman, Daniel L. and Bialek, William},
  journal   = {Phys. Rev. Lett.},
  volume    = {73},
  issue     = {6},
  pages     = {814--817},
  numpages  = {0},
  year      = {1994},
  month     = {8},
  publisher = {American Physical Society},
  doi       = {10.1103/PhysRevLett.73.814},
  url       = {https://link.aps.org/doi/10.1103/PhysRevLett.73.814}
}

@article{Torralba2003statistics,
  doi       = {10.1088/0954-898X/14/3/302},
  url       = {https://dx.doi.org/10.1088/0954-898X/14/3/302},
  year      = {2003},
  month     = {5},
  publisher = {},
  volume    = {14},
  number    = {3},
  pages     = {391},
  author    = {Antonio Torralba and Aude Oliva},
  title     = {Statistics of natural image categories},
  journal   = {Network: Computation in Neural Systems}
}
}
\clearpage
\appendix

\part*{Appendix}
\addcontentsline{toc}{part}{Appendix}
\localtableofcontents
\pagebreak

\section{Proof of \ref{theorem:expansion}}
\label{sec:proof_gaussian_approach}
We use the following lemma:
\begin{lemma}
\label{lemma:cumulants}
Let $a \in \mathbb{R}$; $X$ and $Y$ be random vectors such that $X|Y$ is Gaussian with mean $aY$ and covariance matrix $\mathbf{\Sigma}$ independent of $Y$. Let $\boldsymbol{\kappa}_{k}(X)$ and $\boldsymbol{\kappa}_{k}(Y)$ be the $k$-the cumulant tensors of $X$ and $Y$. Then:
\begin{equation}
    \text{For all } k \geq 3,\boldsymbol{\kappa}_{k}(X) = a^k \boldsymbol{\kappa}_{k}(Y)
\end{equation}
\end{lemma}

\begin{proof}
Since $X|Y$ is Gaussian, we can write:
\begin{align*}
    X = aY+Z,Z\sim \mathcal{N}(0,\mathbf{\Sigma})
\end{align*}
By independence and homogeneity:
\begin{align*}
  \boldsymbol{\kappa}_{k}({X}) &= \boldsymbol{\kappa}_{k}(a{Y}+ {Z})\\
   &=\boldsymbol{\kappa}_{k}(a{Y})+\boldsymbol{\kappa}_{k}({Z})\\
   &= a^k\boldsymbol{\kappa}_{k}({Y})+\boldsymbol{\kappa}_{k}({Z})
\end{align*}
and since $Z$ is Gaussian for $k \geq 3$, $\boldsymbol{\kappa}_{k}({Z})=0$, which yields the desired result.
\end{proof}
\noindent Let us now consider the main \citeprop{theorem:expansion}.

\expansion*
\let\originalqedsymbol\qedsymbol
\renewcommand{\qedsymbol}{}
\begin{proof}
    First, using the law of total expectation:
     \begin{align*}
        \mathbb{E}[\mathbf{z}_t|\mathbf{y}] = \mathbb{E}_{\mathbf{z}_0|\mathbf{y}}[\mathbb{E}[\mathbf{z}_t|\mathbf{z}_0]| \mathbf{y}]
    \end{align*}
But conditioned on $\mathbf{z}_0$, we have $\mathbf{z}_t = a_t \mathbf{z}_0 + b_t \boldsymbol{\epsilon}$, $\boldsymbol{\epsilon} \sim \mathcal{N}(\mathbf{0},\mathbf{I})$, so $\mathbb{E}[\mathbf{z}_t|\mathbf{z}_0] = a_t\mathbf{z}_0$, and therefore:
$$\mathbb{E}[\mathbf{z}_t|\mathbf{y}] = a_t \mathbb{E}[\mathbf{z}_0|\mathbf{y}] = a_t \boldsymbol{\mu}_\mathbf{y}$$
Second, using the law of total variance:
\begin{align*}
        \Var(\mathbf{z}_t|\mathbf{y}) &= \mathbb{E}_{\mathbf{z}_0|\mathbf{y}}[\Var(\mathbf{z}_t|\mathbf{z}_0)|\mathbf{y}] + \Var(\mathbb{E}_{\mathbf{z}_0|\mathbf{y}}[\mathbf{z}_t|\mathbf{z}_0]|\mathbf{y})\\
        &=\mathbb{E}_{\mathbf{z}_0|\mathbf{y}}[b^2_t \mathbf{I}] + \Var(a_t \mathbf{z}_0|\mathbf{y})\\
        &=b^2_t \mathbf{I} + a_t^2\Var(\mathbf{z}_0|\mathbf{y}) \\
        &=b^2_t \mathbf{I} + a_t^2 \boldsymbol{\Sigma}_\mathbf{y}
    \end{align*}
Finally, we perform a Gram-Charlier series A expansion \cite{dharmani2018multivariate} of $p(\mathbf{z}_t|\mathbf{y})$, given by:
\begin{align*}
    p(\mathbf{z}_t|\mathbf{y}) =& \varphi(\mathbf{z}_t; \boldsymbol{\mu}, \boldsymbol{\Sigma})\Bigl(1+\sum_{k=3}^\infty\frac{\boldsymbol{\kappa}_{k}}{k!}\mathbf{H}_{k}(\mathbf{z}-\boldsymbol{\mu}; \boldsymbol{\Sigma})\Bigr)
\end{align*}
where:
\begin{itemize}
    \item $\varphi(\mathbf{z}_t; \boldsymbol{\mu}, \boldsymbol{\Sigma})$ is the p.d.f of a normal distribution of parameters $(\boldsymbol{\mu}, \boldsymbol{\Sigma})$;
    \item $\boldsymbol{\mu} = a_t\boldsymbol{\mu}_\mathbf{y}$ and $\boldsymbol{\Sigma} = a^2_t \boldsymbol{\Sigma}_\mathbf{y}+b_t^2 \mathbf{I}$;
    \item $\boldsymbol{\kappa}_{k}$ is the $k$-th cumulant of $\mathbf{z}_t|\mathbf{y}$;
    \item $\mathbf{H}_k(\mathbf{z}-\boldsymbol{\mu}; \boldsymbol{\Sigma})$ is the multivariate Hermite polynomial of order $k$, given by \cite{holmquist1996multivariate_hermite_poly}:
    \begin{equation}
        \mathbf{H}_k(\mathbf{x}; \boldsymbol{\Sigma}) = \varphi(\mathbf{x}; \boldsymbol{\Sigma})^{-1}(-1)^k(\boldsymbol{\Sigma}D_{\mathbf{x}})^{\otimes k}\varphi(\mathbf{x}; \boldsymbol{\Sigma})
    \end{equation}
    where $D_{\mathbf{x}}$ is the derivative operator and $\otimes$ is the Kronecker Product Operator.
\end{itemize}

Since $\mathbf{z}_t|(\mathbf{z}_0, \mathbf{y}) = \mathbf{z}_t|\mathbf{z}_0$ is Gaussian with covariance matrix $b_t\mathbf{I}$ (by definition of the forward process), we can use Lemma~\ref{lemma:cumulants} on $\mathbf{z}_t|\mathbf{y}$ and $\mathbf{z}_t|(\mathbf{z}_0, \mathbf{y})$. The cumulants $\boldsymbol{\kappa}_k$ simplify to $a_t^k \boldsymbol{\kappa}_{k}(\mathbf{z}_0|\mathbf{y})$, that is $O(a_t^k)$ since the cumulants are assumed finite. Additionally, since $\underset{t\rightarrow T}{\lim}(b_t) \neq 0$, the Hermite polynomials remain bounded as $t$ approaches $T$. Therefore, the Gram-Charlier expansion can be written as:
\begin{align*}
   p(\mathbf{z}_t|\mathbf{y})=& \varphi(\mathbf{z}_t; \boldsymbol{\mu}, \boldsymbol{\Sigma})\Bigl(1+\sum_{k=3}^\infty\frac{\boldsymbol{\kappa}_{k}}{k!}\mathbf{H}_{k}(\mathbf{z}-\boldsymbol{\mu}; \boldsymbol{\Sigma})\Bigr)\\
    =& \varphi(\mathbf{z}_t; \boldsymbol{\mu}, \boldsymbol{\Sigma})\Bigl(1+\sum_{k=3}^\infty\frac{a_t^k\boldsymbol{\kappa}_{k}(\mathbf{z}_0|\mathbf{y})}{k!}\mathbf{H}_{k}(\mathbf{z}-\boldsymbol{\mu}; \boldsymbol{\Sigma})\Bigr)\\
    =& \varphi(\mathbf{z}_t; \boldsymbol{\mu}, \boldsymbol{\Sigma})+O(a_t^3) \quad \tag*{\originalqedsymbol}
\end{align*}
\end{proof}
\renewcommand{\qedsymbol}{\originalqedsymbol}

\section{A Deeper Interpretation of the \saga{} Method}
\label{sec:appendix_interpretation}

In this section, we provide a comprehensive interpretation of our \saga{} method. We begin by positioning it within the landscape of generative guidance and then unfold its mechanics through a signal-processing lens, progressing from our base model to its more advanced theoretical implications.

\subsection{From Point-Wise Guidance to Distributional Modeling}

Methods for improving prompt alignment in diffusion models have historically focused on modifying individual generation trajectories. For instance, approaches like GSN~\cite{chefer2023attendandexcite} guide a specific latent $\mathbf{z}_t$ during denoising. Others, like InitNO~\cite{guo2024initno}, address the issue of out-of-distribution latents by optimizing the initial noise $\mathbf{z}_T$ to find a more favorable starting point.

Our \saga{} method introduces a fundamental shift in perspective from \textit{point-wise} manipulation to \textit{distributional} modeling. The core insight is that for a given prompt $\mathbf{y}$, the true posterior distribution of latents $p(\mathbf{z}_t|\mathbf{y})$ is complex and possibly multi-modal, with each mode representing a distinct semantic interpretation of the prompt (e.g., different dog breeds for the prompt "a photo of a dog").

\saga{} is designed to efficiently locate and model the \textit{single most dominant mode} relevant to the prompt, as guided by an alignment criterion $\mathcal{L}$. Our theoretical foundation, Proposition 1, establishes that for $t \to T$, any such mode is well-approximated by a Gaussian distribution. Our approach, therefore, consists in optimizing the parameters of an approximating Gaussian, $q(\mathbf{z}_t|\mathbf{y})$, to match this target mode.

Rather than modeling this entire complex distribution, \saga{} is designed to efficiently locate and approximate the \textit{single most dominant mode} relevant to the prompt, guided by our alignment criterion $\mathcal{L}$. As established in \citeprop{theorem:expansion}, any single mode of this distribution is well-approximated by a Gaussian for $t \rightarrow T$. Our method, therefore, optimizes the parameters of a single Gaussian $q(\mathbf{z}_t|\mathbf{y}, \boldsymbol{\tilde{\mu}}_\mathbf{y})$ to match this target mode. The final learned parameter $\boldsymbol{\tilde{\mu}}_\mathbf{y}^*$ represents the mean of this mode---the central tendency of a specific, coherent interpretation of the prompt.

\subsection{A Signal-Processing Perspective: Decomposing Generation by Frequency}

We analyze this approach from a signal-processing viewpoint to understand why it is effective. The generation of a sample in our framework can be written as:
\begin{equation}
    \mathbf{z}_t = a_t \boldsymbol{\tilde{\mu}}_\mathbf{y} + b_t\boldsymbol{\epsilon}, \quad \boldsymbol{\epsilon} \sim \mathcal{N}(\mathbf{0}, \mathbf{I})
\end{equation}

This equation represents a decomposition of the latent $\mathbf{z}_t$ into two distinct components:
\begin{itemize}
    \item \textit{A deterministic, low-frequency signal component}: The term $a_t \boldsymbol{\tilde{\mu}}_\mathbf{y}$ represents the structural foundation of the image. The vector $\boldsymbol{\tilde{\mu}}_\mathbf{y}$ is constant for all samples drawn for a given prompt. As it approximates $\mathbb{E}[\mathbf{z}_0|\mathbf{y}]$, it captures the shared, essential characteristics (\textit{the coarse layout, the primary subjects, the overall composition}) which are primarily encoded in the \textit{low-frequency bands} of the image spectrum.
    \item \textit{A stochastic, high-frequency detail component}: The term $b_t\boldsymbol{\epsilon}$ introduces random variations. Since $\boldsymbol{\epsilon}$ is white noise, its power is distributed across all frequencies. This component is responsible for the high-frequency details that make each generated image unique: textures, fine lines, and other stochastic elements.
\end{itemize}

This decomposition is particularly meaningful in the context of natural image statistics. The Power Spectral Density (PSD) of natural images is known to follow a power law, approximately $P(f) \propto 1/f^2$~\cite{Ruderman1994statisticnaturalimage,Torralba2003statistics}, meaning most of the signal energy is concentrated in low frequencies. Diffusion models implicitly leverage this structure, learning to reconstruct the high-power, low-frequency components at early denoising stages (high $t$) before adding low-power, high-frequency details at later stages (low $t$)~\cite{rissanen2023generative,park2023understanding}.

Our optimization of $\mathbb{E}_{q}[\mathcal{L}(\mathbf{z}_t, \mathbf{y}, t)]$ can be interpreted as a search for the optimal \textit{low-frequency foundation} $\boldsymbol{\tilde{\mu}}_\mathbf{y}$ that, when combined with random high-frequency noise, consistently produces samples that minimize the alignment loss. By isolating the structural component in $\boldsymbol{\tilde{\mu}}_\mathbf{y}$, we force the optimization to focus on what is essential for prompt alignment. This explains a key behavior of \saga{}: samples generated from a single learned distribution $q(\mathbf{z}_t|\mathbf{y}, \boldsymbol{\tilde{\mu}}_\mathbf{y}^*)$ exhibit strong structural consistency (e.g., same object placement) but rich diversity in appearance and texture. The structure is ``locked in'' by the shared $\boldsymbol{\tilde{\mu}}_\mathbf{y}^*$, while the diversity comes from the random sampling of $\boldsymbol{\epsilon}$.

\subsection{Implications for the Initialization Strategy}
\label{appendix:initialization_strategy}
This frequency-based interpretation also clarifies why our initialization strategy is particularly effective. We initialize the optimization with $\boldsymbol{\tilde{\mu}}_\mathbf{y} \leftarrow  \mathbf{\hat{z}}_0(\mathbf{z}_t, \mathbf{y}, t)_{HW}$ (per-channel spatial mean). Since $\mathbf{\hat{z}}_0$ is the pre-trained model's prediction of the average final clean image, it already contains a strong, plausible estimate of the low-frequency content corresponding to the prompt $\mathbf{y}$. By using this as a starting point, we begin our optimization from a point already in a promising region of the parameter space, significantly accelerating convergence to a high-quality mode.

\begin{figure}[]
    \centering
   \includegraphics[width=1\linewidth]{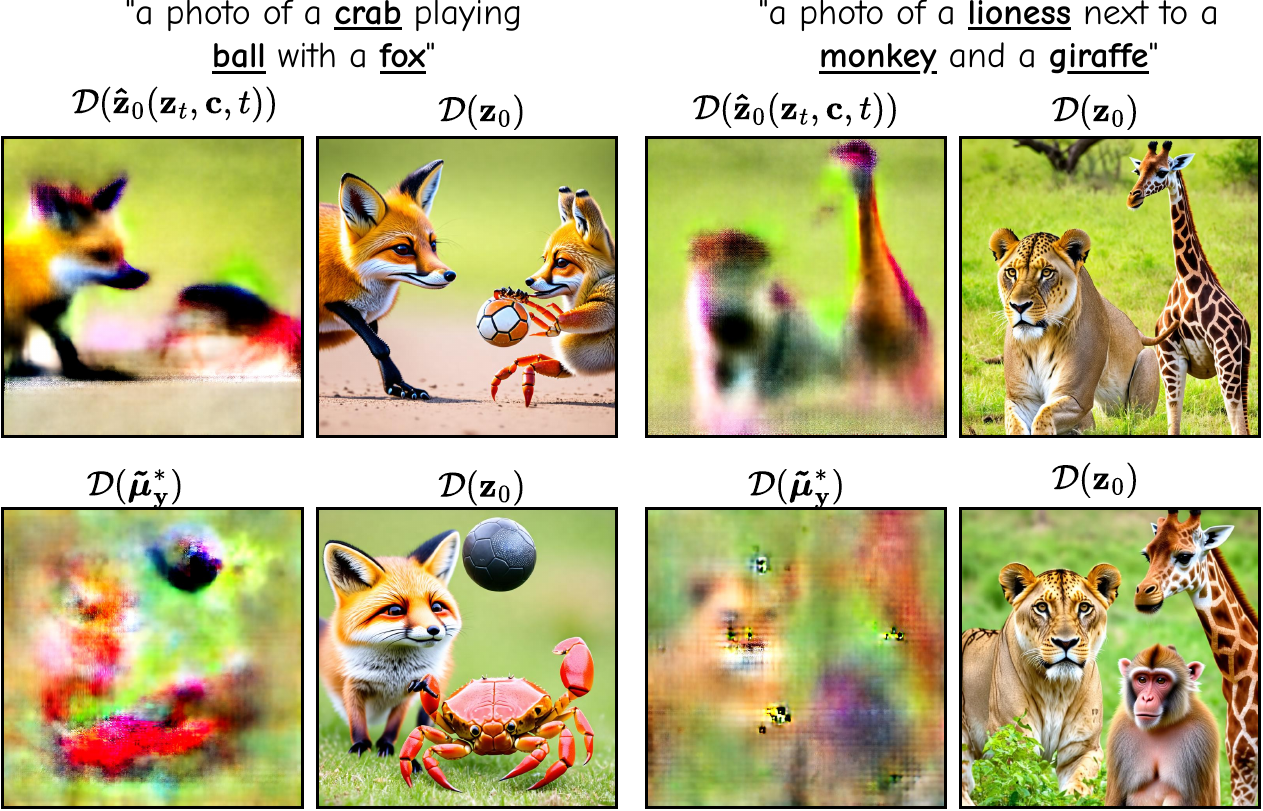}
    \caption{Visualization comparing a baseline generation (top row) with our \saga-optimized method (bottom row) on \sdthree. The baseline shows the estimated $\mathbf{\hat{z}}_0(\mathbf{z}_t, \mathbf{y}, t)$ and its corresponding final image, while \saga displays the optimized $\boldsymbol{\tilde{\mu}}_\mathbf{y}$ and its resulting image. By constructing low-frequency details in $\boldsymbol{\tilde{\mu}}_\mathbf{y}$, \saga produces images with improved alignment to the prompt.}
    \label{fig:samples_low_fq}
\end{figure}
Finally, this perspective underscores that the mode found by \saga{} depends on the initialization. Different starting points for $\mathbf{z}_t$ (used to compute the initial $\mathbf{\hat{z}}_0$) can lead the optimization to converge to different local minima of the loss landscape. These different minima correspond to different valid low-frequency interpretations (i.e., different modes) of the prompt $\mathbf{y}$. Our method thus provides a tool not for finding a single, universal representation, but for efficiently exploring and sampling from distinct, plausible interpretations of a textual concept. We illustrate potential low-frequency bases in \citefigure{fig:samples_low_fq}.
\begin{figure}[]
    \centering
    \includegraphics[width=1\linewidth]{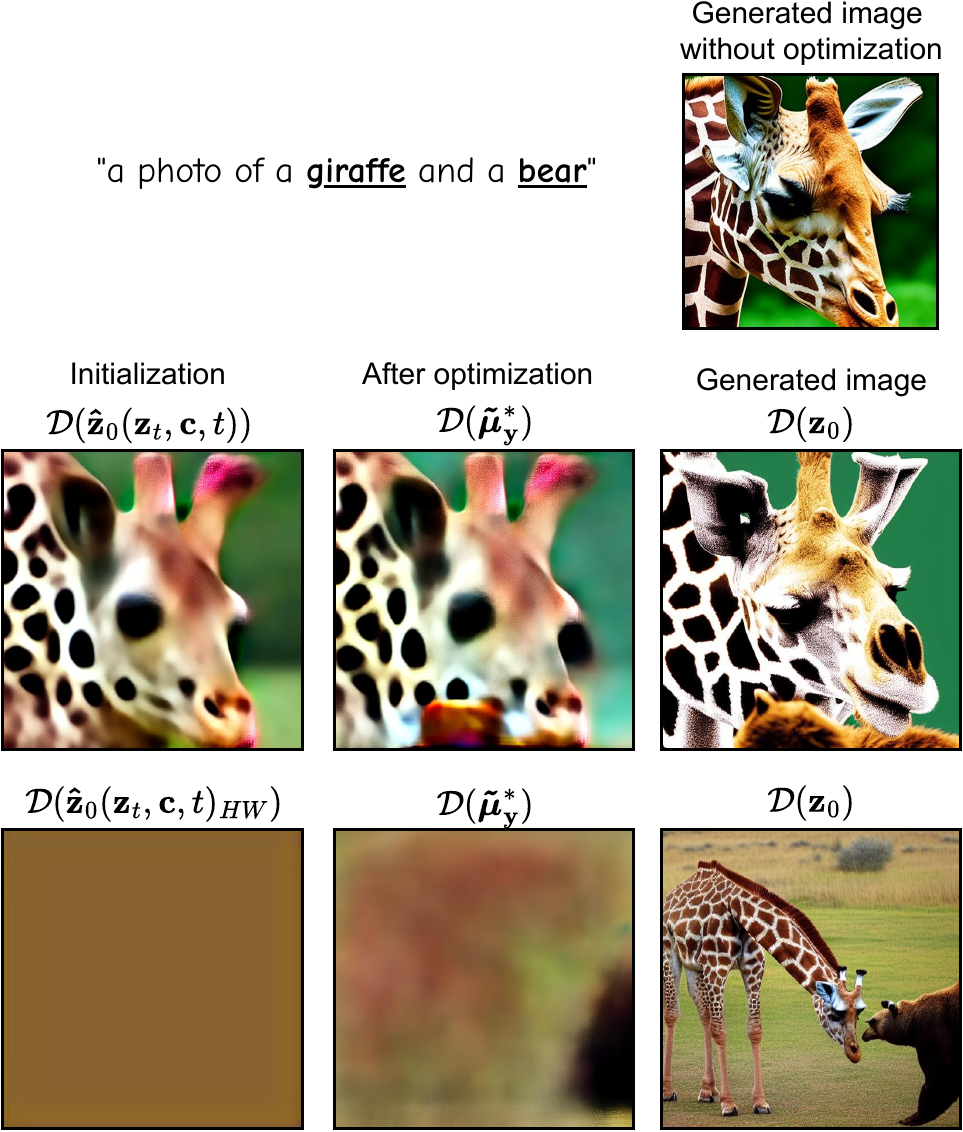}
    \caption{Comparison of optimization outcomes with different initialization strategies on \sdone with \saga. Top row: A poor initialization using the full $\mathbf{\hat{z}}_0(\mathbf{z}_t, \mathbf{y}, t)$ leads to optimization failure. Bottom row: Initializing with $\boldsymbol{\tilde{\mu}}_\mathbf{y} \leftarrow  \mathbf{\hat{z}}_0(\mathbf{z}_t, \mathbf{y}, t)_{HW}$ allows the optimization process to construct the required frequency components successfully.}
    \label{fig:bad_init}
\end{figure}

We avoid initializing $\boldsymbol{\tilde{\mu}}_\mathbf{y}$ with the denoised estimate $\mathbf{\hat{z}}_0(\mathbf{z}_t, \mathbf{y}, t)$ as it can be detrimental. At intermediate timesteps $t$, $\mathbf{\hat{z}}_0(\mathbf{z}_t, \mathbf{y}, t)$ already contains significant low-frequency information, defining a coarse scene structure. If this structure is flawed, for example, as in~\citefigure{fig:bad_init}, the optimization starts from a poor initial point. This increases the risk of converging to a poor local minimum, especially with a limited optimization budget. To avoid committing to a potentially flawed structure and allow for greater flexibility during optimization, we initialize the process using only the zeroth frequency coefficient. We leave a deeper study of the best initialization methods for future work.

This also opens up avenues for explainability, as inspecting the learned $\boldsymbol{\tilde{\mu}}_\mathbf{y}$ can reveal the core structural features the model associates with a given prompt.
Other initializations could have been used, such as $\mathbf{0}$ or random noise. We leave these for future work.

\subsection{Advanced Control via Learned Covariance}
\label{sec:variance_details}

\begin{algorithm}[t]

\caption{Find optimal $\boldsymbol{\tilde{\mu}}_\mathbf{y}$, $\boldsymbol{\tilde{\Sigma}}_\mathbf{y}$}\label{alg:method_var}
\KwInput{$t$, a prompt $\mathbf{y}$, $N$ steps of optimization, an optimization step size $\alpha$, noise scheduler parameters $a_t, b_t$}
\KwOutput{ $\boldsymbol{\tilde{\mu}}_\mathbf{y}$,$\boldsymbol{\tilde{\Sigma}}_\mathbf{y}$ such that $q(\mathbf{z}_t|\mathbf{y}, \boldsymbol{\tilde{\mu}}_\mathbf{y})$ approximates $p(\mathbf{z}_t|\mathbf{y})$.}
Initialize $\boldsymbol{\tilde{\mu}}_\mathbf{y}$and $\boldsymbol{\tilde{L}}$ the lower triangular as $\boldsymbol{\tilde{L}}\boldsymbol{\tilde{L}}^T=\boldsymbol{\tilde{\Sigma}}_\mathbf{y}$

\For{$i = 1$ to $N$} {
$\boldsymbol{\epsilon} \sim \mathcal{N}(\mathbf{0}, \mathbf{I})$

$\mathbf{z}_t = a_t \boldsymbol{\tilde{\mu}}_\mathbf{y} + \boldsymbol{\tilde{L}} \boldsymbol{\epsilon}$

$\boldsymbol{\tilde{\mu}}_\mathbf{y} \leftarrow \boldsymbol{\tilde{\mu}}_\mathbf{y} - \alpha \nabla_{\boldsymbol{\tilde{\mu}}_\mathbf{y}}\mathcal{L}(\mathbf{z}_t, \mathbf{y},t)$

$\mathbf{\tilde{L}}_\mathbf{y} \leftarrow \mathbf{\tilde{L}}_\mathbf{y} - \alpha \nabla_{\mathbf{\tilde{L}}_\mathbf{y}}\mathcal{L}(\mathbf{z}_t, \mathbf{y},t)$
}
\textbf{return} $\boldsymbol{\tilde{\mu}}_\mathbf{y}$,$\boldsymbol{\tilde{\Sigma}}_\mathbf{y}=\boldsymbol{\tilde{L}}\boldsymbol{\tilde{L}}^T$

\end{algorithm}

\paragraph{Learning a Spatially-Variant Covariance.}
While powerful, the base model assumes a spatially-uniform noise covariance, effectively modeling the total covariance as $b_t^2\mathbf{I}$ by neglecting  $a_t^2\boldsymbol{\tilde{\Sigma}}_\mathbf{y}$ and thereby treating all image regions equally. To achieve finer-grained control, we extend our method to learn a data-dependent, spatially-variant covariance, an extension motivated by the formulation in \citeprop{theorem:expansion}.

To this end, we rewrite $\boldsymbol{\tilde{\Sigma}}_\mathbf{y}$ to be the total conditional covariance via a re-parameterization that separates the fixed part from a learnable component $\boldsymbol{\Sigma}_{\theta}(\mathbf{y}, t)$:
$$
\boldsymbol{\tilde{\Sigma}}_\mathbf{y} \triangleq a_t^2 \boldsymbol{\Sigma}_{\theta}(\mathbf{y}, t) + b_t^2\mathbf{I}
$$
This principled construction allows the sampling distribution to be expressed cleanly as:
$$
q(\mathbf{z}_t|\mathbf{y}) = \mathcal{N}(\mathbf{z}_t; a_t\boldsymbol{\tilde{\mu}}_\mathbf{y}, \boldsymbol{\tilde{\Sigma}}_\mathbf{y})
$$
For practical optimization, we learn the covariance $\boldsymbol{\tilde{\Sigma}}_\mathbf{y}$ via its Cholesky decomposition $\boldsymbol{\tilde{\Sigma}}_\mathbf{y} = \boldsymbol{\tilde{L}}\boldsymbol{\tilde{L}}^T$.
The complete algorithm for jointly optimizing both the mean $\boldsymbol{\tilde{\mu}}_\mathbf{y}$ and the Cholesky factor $\boldsymbol{\tilde{L}}$ is presented in ~\citealgorithm{alg:method_var}.

\paragraph{Spatially-Variant Noise with Diagonal Covariance.}

For computational tractability, we model $\boldsymbol{\tilde{\Sigma}}_\mathbf{y}$ as a diagonal matrix. This extension allows the model to learn a \textit{spatially-variant noise schedule}. The optimization gains a new degree of freedom: it can learn not only \textit{what} the foundational signal is, but also \textit{where} that signal is most critical. For latent features essential to prompt alignment (\eg, facial features), the model learns a smaller variance, thereby "protecting" the signal in $\boldsymbol{\tilde{\mu}}_\mathbf{y}$. Conversely, for background or textural regions, it can learn a larger variance to encourage diversity. This elevates the model from applying uniform noise to employing an intelligent, spatially-aware noise distribution optimized for alignment.

\paragraph{Theoretical Ideal: Structured Noise with Full Covariance.}
Hypothetically, learning a full, non-diagonal covariance matrix represents the ultimate form of generative control. This would allow the model to capture \textit{correlations} between latent features. The injected noise would no longer be simply static but would become \textit{structured noise}, possessing coherent patterns like wood grain, brushstrokes, or organic textures. Our signal-processing interpretation is thereby enriched: $\boldsymbol{\tilde{\mu}}_\mathbf{y}$ remains the low-frequency foundation, but the noise component evolves into a \textit{structured stochastic process} with its own complex spectral properties. In this ideal scenario, the model would learn not just the optimal base image, but the entire statistical family of valid "detail maps" to superimpose upon it, offering the most sophisticated and powerful balance between structural fidelity and generative diversity. 

An empirical study of the performance of our approach with regard to various (co)variance modeling is reported in Section~\ref{sec:estim_variance}.

\section{Signal Rescaling Details}\label{appendix:rescaling}

\begin{algorithm}[t]
\caption{Rescaling $\boldsymbol{\tilde{\mu}}_\mathbf{y}$}
\label{alg:rescaling}
\KwInput{A reference initialization $\mathbf{\hat{z}}_0(\mathbf{z}_t, \mathbf{y}, t)$. The vector $\boldsymbol{\tilde{\mu}}_\mathbf{y}$ to be rescaled.}
\KwOutput{The rescaled $\boldsymbol{\tilde{\mu}}_\mathbf{y}$.}
\BlankLine

$\sigma_{\text{ref}} \leftarrow \text{StandardDeviation}(\mathbf{\hat{z}}_0(\mathbf{z}_t, \mathbf{y}, t))$\;
$\sigma_{\text{opt}} \leftarrow \text{StandardDeviation}(\boldsymbol{\tilde{\mu}}_\mathbf{y})$\;
\BlankLine

\tcp{Shrink $\boldsymbol{\tilde{\mu}}_\mathbf{y}$ if its deviation is larger than the reference's}
\If{$\sigma_{\text{opt}} > \sigma_{\text{ref}}$}{
    $\boldsymbol{\tilde{\mu}}_\mathbf{y} \leftarrow \frac{\sigma_{\text{ref}}}{\sigma_{\text{opt}}} \cdot \boldsymbol{\tilde{\mu}}_\mathbf{y}$\;
}
\BlankLine
\textbf{return} $\boldsymbol{\tilde{\mu}}_\mathbf{y}$\;
\end{algorithm}

As detailed in the main paper, our optimization of the signal component $\boldsymbol{\tilde{\mu}}_\mathbf{y}$ can lead to latent vectors with excessively high variance, causing over-saturation in the final image. Our rescaling procedure, shown in \citealgorithm{alg:rescaling}, is designed to prevent this. The reference latent $\mathbf{\hat{z}}_0$ is computed once at the start of the process and is used to initialize $\boldsymbol{\tilde{\mu}}_\mathbf{y}$. This same $\mathbf{\hat{z}}_0$ then serves as a \textit{fixed reference} for its standard deviation, $\sigma_{\text{ref}}$, throughout the optimization. After each gradient update on $\boldsymbol{\tilde{\mu}}_\mathbf{y}$, we check if its standard deviation has grown larger than $\sigma_{\text{ref}}$. If it has, we multiplicatively shrink the vector to match the reference deviation. This per-step check provides stability, particularly when using aggressive hyperparameters (\eg, a high learning rate) for faster convergence, as it prevents the signal's dynamic range from diverging.

While this hard-clipping approach is effective, several alternative strategies could be explored in future work. For instance, a soft constraint could be implemented by adding a regularization term to the loss function, such as a penalty on the squared difference $||\text{std}(\boldsymbol{\tilde{\mu}}_\mathbf{y}) - \sigma_{\text{ref}}||^2$. 
Another direction is to operate directly in pixel space by decoding $\boldsymbol{\tilde{\mu}}_\mathbf{y}$, applying traditional image saturation adjustments, and then re-encoding the result. Finally, the hard clipping in \citealgorithm{alg:rescaling} could be relaxed by introducing a temperature parameter, allowing for more fine-grained control over the signal dynamic range.

\section{Analysis and Comparison of Alternative Method Configurations}
\label{sec:definition_approaches}

\paragraph{All the Methods}

Our primary method, denoted \saga, learns only the optimal mean $\boldsymbol{\tilde{\mu}}_\mathbf{y}$ while keeping the covariance fixed. The variant that also learns the covariance $\boldsymbol{\tilde{\Sigma}}_\mathbf{y}$ is denoted \sagavar. Finally, both approaches can be combined with Syngen-GSN guidance after the distribution is learned; we denote these enhanced variants \sagaplus and \sagaplusvar, respectively.

\begin{figure*}[htbp]
  \centering
  \subcaptionbox{One distribution is learned and $N = 3$ latent images are sampled from it.\label{fig:saga_one_distrib}}
  {\includegraphics[width=0.49\linewidth, valign=b]{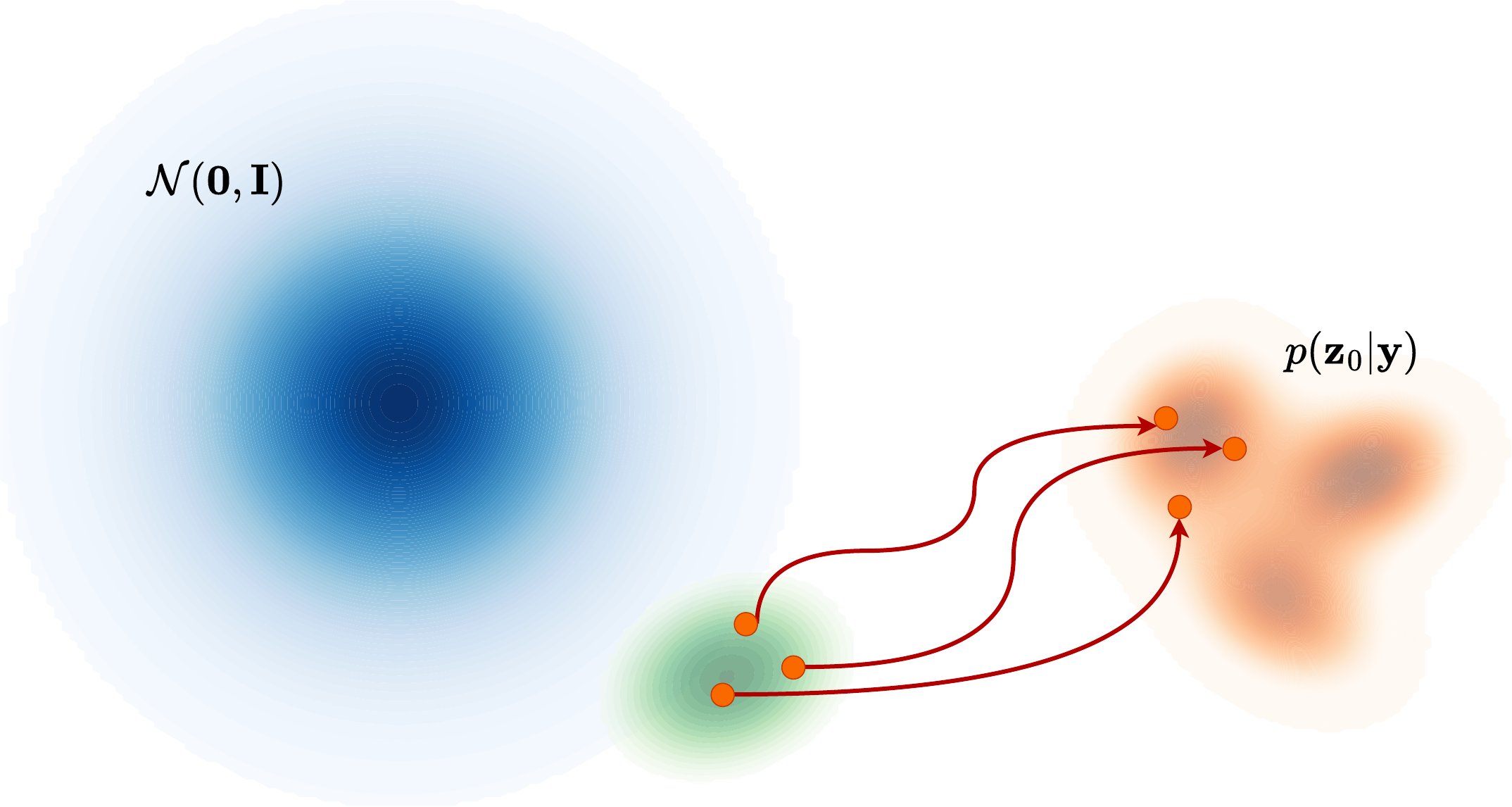}}%
  \hfill
  \subcaptionbox{$N = 3$ distributions are learned and one latent image is sampled from each.\label{fig:saga_three_distrib}}
  {\includegraphics[width=0.49\linewidth, valign=b]{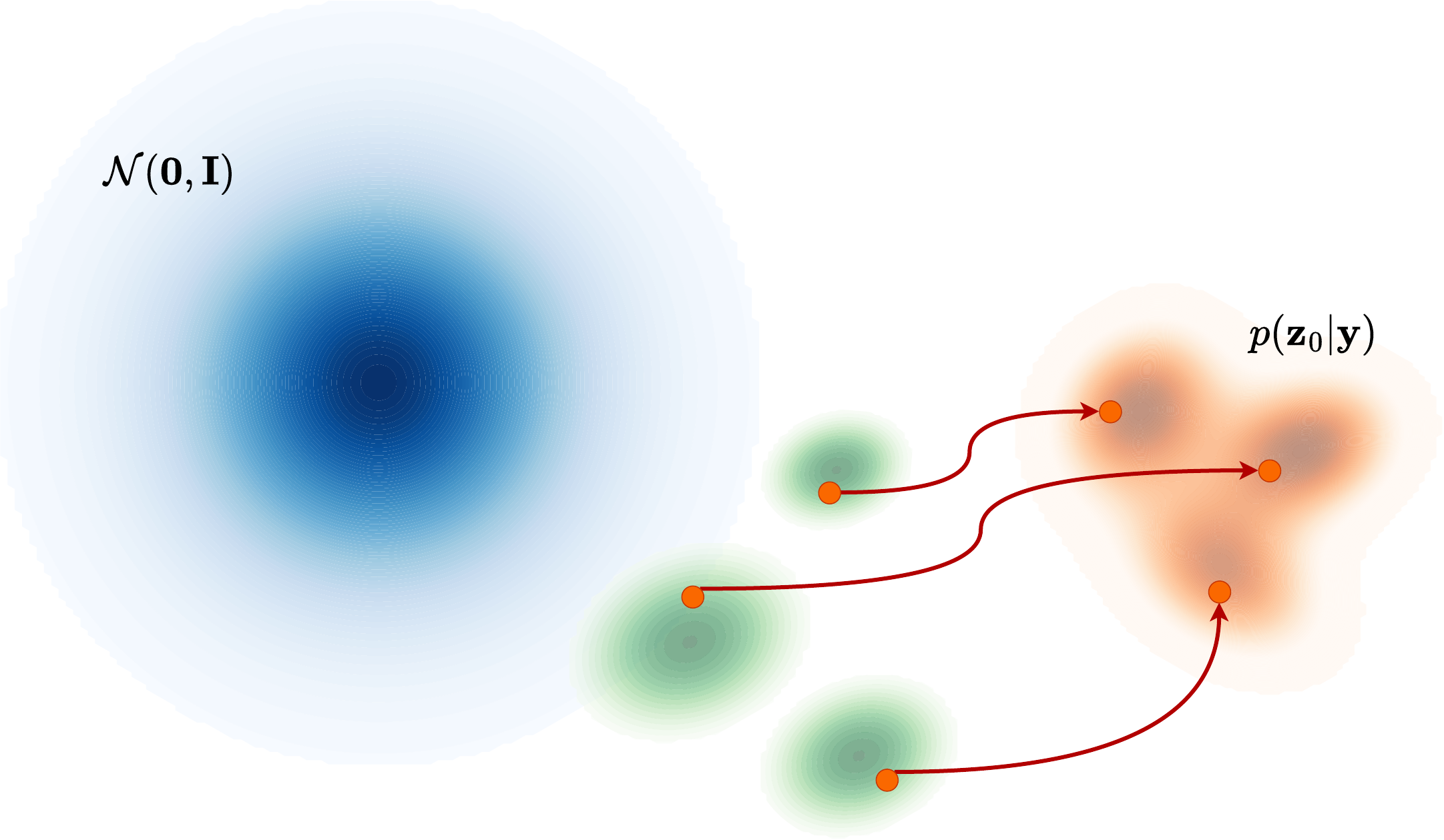}}%
  \caption{The two sampling strategies of \saga.}
  \label{fig:main_saga}
\end{figure*}

Additionally, our method provides two sampling strategies to generate $N$ images for a prompt $\mathbf{y}$. The first involves denoising a single latent $\mathbf{z}_T$ to an intermediate step $t$. This single $\mathbf{z}_t$ then initializes a distribution $q(\mathbf{z}_t|\mathbf{y})$ from which we draw $N$ samples that are fully denoised to obtain the final images. This strategy is illustrated in Figure~\ref{fig:saga_one_distrib}. The second strategy we adopt in the main paper to maximize sample diversity starts with $N$ independent latents $\mathbf{z}_T$. Each is denoised to step $t$, and each of the $N$ resulting latents initializes its own distinct distribution. We then draw one latent from each of these $N$ distributions and complete the denoising process. The second strategy is illustrated in Figure~\ref{fig:saga_three_distrib}. We use the latter approach in the main paper. Both strategies yield similar quantitative scores, which are detailed in~\citeappendix{sec:evaluation_appendix}.
Approaches that learn a single, unique distribution per prompt are denoted with the subscript \unique. This gives rise to the variants \sagaone and \sagaplusone, along with their respective covariance-learning counterparts, \sagaonevar and \sagaplusonevar.

\paragraph{Diversity}

\begin{figure*}[]
    \centering
    \includegraphics[width=1\linewidth]{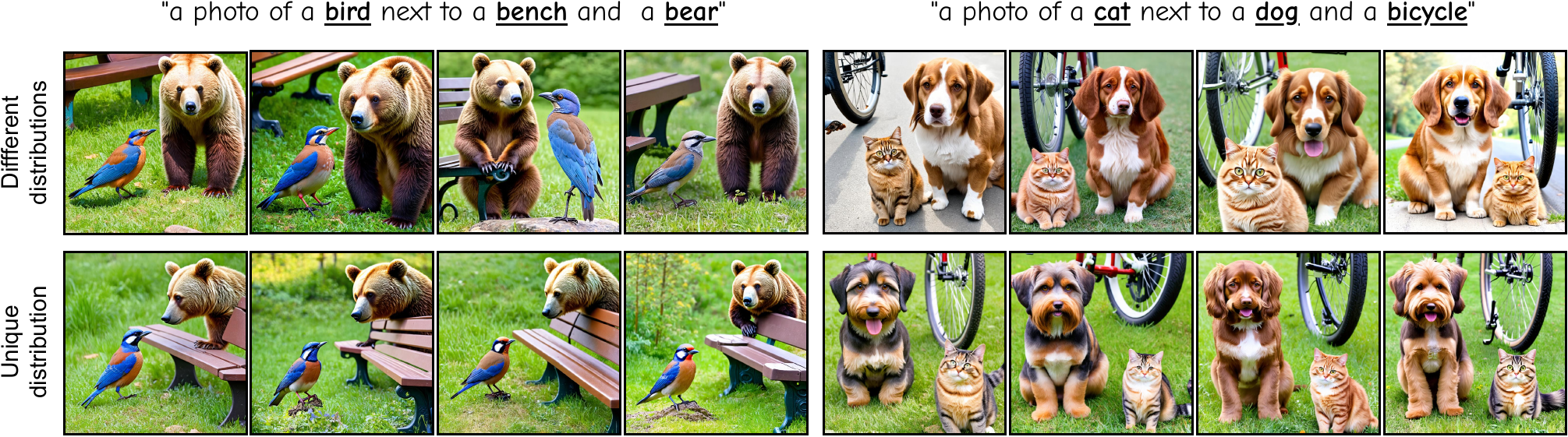}
   \caption{Qualitative comparison of generated images using the same \sdthree backbone. 
(Top) \saga (\textit{Different distributions}) yields significant compositional variability, altering the spatial arrangement of entities. 
(Bottom) The \sagaone variant (\textit{Unique distribution}) maintains a more consistent composition while still producing variations in entity appearance. Both methods successfully generate stylistic diversity.}
    \label{fig:variation}
\end{figure*}

\citefigure{fig:variation} illustrates the qualitative differences between \saga and \sagaone when generating samples from identical prompts. \saga demonstrates superior variety in the image composition. For instance, the relative positions of the bear, the bird, and the bench change across samples. In contrast, \sagaone fixes the global scene layout, consistently placing entities in similar locations. Despite its fixed composition, \sagaone still generates meaningful stylistic variations; for example, the cat and the dog are altered by sampling different latent vectors $\mathbf{z}_t$. This makes \sagaone a compelling and computationally efficient alternative. Its efficiency stems from learning a single distribution, allowing new latent vectors $\mathbf{z}_t$ to be sampled to generate images instantly. This is in direct contrast to \saga, which requires a per-sample optimization to produce each new instance. \saga offers a trade-off between compositional diversity and generative speed.

\section{Implementation Details \label{sec:implementation_details}}

\subsection{Stable Diffusion 1.4 (\sdone)}

We use the model available on \url{https://huggingface.co/CompVis/stable-diffusion-v1-4}. All generations, whether with standard diffusion or the methods described below, were performed using Classifier-Free Guidance \citep{ho2022classifierfree} set to 7.5. We performed sampling using the DDPM scheduler \url{https://huggingface.co/docs/diffusers/api/schedulers/ddpm})
with 50 steps and \textit{float32} precision. \citetable{tab:diffusion_steps_sd14} details the mapping from the discrete step index to its corresponding sampling step value $t$. All images were generated at a resolution of $512\times512$ using NVIDIA H100 GPUs, with an average time 
30 seconds per image for \saga.

\begin{table}[h]
  
    \centering
     \caption{The mapping between the discrete step index (idx) and its corresponding sampling step value $t$.}    
    {
    \small
    \begin{tabular}{rr|rr|rr|rr}
    \toprule
    \textbf{idx} & \textbf{$t$} & \textbf{idx} & \textbf{$t$} & \textbf{idx} & \textbf{$t$} & \textbf{idx} & \textbf{$t$} \\
    \midrule
    0 & 981 & 13 & 721 & 26 & 461 & 39 & 201 \\
    1 & 961 & 14 & 701 & 27 & 441 & 40 & 181 \\
    2 & 941 & 15 & 681 & 28 & 421 & 41 & 161 \\
    3 & 921 & 16 & 661 & 29 & 401 & 42 & 141 \\
    4 & 901 & 17 & 641 & 30 & 381 & 43 & 121 \\
    5 & 881 & 18 & 621 & 31 & 361 & 44 & 101 \\
    6 & 861 & 19 & 601 & 32 & 341 & 45 &  81 \\
    7 & 841 & 20 & 581 & 33 & 321 & 46 &  61 \\
    8 & 821 & 21 & 561 & 34 & 301 & 47 &  41 \\
    9 & 801 & 22 & 541 & 35 & 281 & 48 &  21 \\
    10 & 781 & 23 & 521 & 36 & 261 & 49 &   1 \\
    11 & 761 & 24 & 501 & 37 & 241 &    &     \\
    12 & 741 & 25 & 481 & 38 & 221 &    &     \\
    \bottomrule
    \end{tabular}
    
    }
     
    \label{tab:diffusion_steps_sd14}
\end{table}

For attention map preprocessing, InitNO, Attend\&Excite, and our approach follow the same procedure as described in \cite{chefer2023attendandexcite}. Specifically, attention maps are scaled by a factor of 100, followed by the application of a softmax function. A Gaussian smoothing operation is then performed using a kernel of size 3 and a standard deviation of 0.5. In contrast, Syngen applies only a rescaling step without softmax normalization or smoothing.

\paragraph{InitNO~\cite{guo2024initno}}

We generate images using the official implementation: \url{https://github.com/xiefan-guo/initno}). The original method optimizes a loss function composed of three terms: self-attention loss, cross-attention loss, and KL divergence loss. Notably, the KL divergence loss is applied only after backpropagation on the attention losses. However, its implementation differs from the description in the paper, as the optimization proceeds in two separate stages: first, minimizing the attention-based losses, followed by optimizing the KL divergence term.  
To define an initial latent, the optimization is performed multiple times (up to five) with different noise samples drawn from a Gaussian distribution. If the cross-attention and self-attention losses do not meet predefined thresholds, a new latent initialization is sampled, and the process is repeated. If convergence is not achieved within five attempts, the inference is conducted using the latent initialization that yielded the best objective score. An additional loss, not mentioned in their paper and named \textit{clean cross-attention loss}, is used with a special processing.
InitNO+ adds iterative refinement steps at sampling steps 10 and 20, where up to 20 iterative refinement steps are performed. The losses must meet specified thresholds of 0.2 for the cross-attention loss and 0.3 for the self-attention loss. Additionally, GSN guidance is applied during the first 25 sampling steps. The learning rate decreases progressively with each sampling step, starting from an initial value of 20. They also introduce a new loss, not mentioned in their paper, referred to as the \textit{cross-attention alignment loss}, which aligns cross-attention maps from previous sampling steps stored in cache.

\paragraph{Attend\&Excite~\cite{chefer2023attendandexcite}}

We use the implementation from the \url{https://huggingface.co/docs/diffusers/api/pipelines/attend_and_excite}{Diffusers library}. A gradient-based shift is applied to the latent image during the first 25 sampling steps. The learning rate starts at 20 and gradually decreases throughout the process. Furthermore, they apply an iterative refinement step  at sampling steps 0, 10, and 20 to ensure the loss reaches the thresholds of 0.05, 0.5, and 0.8, respectively. Up to 20 iterative refinement steps are performed.

\paragraph{Syngen~\cite{rassin2023linguistic}}

We employ \url{https://github.com/RoyiRa/Linguistic-Binding-in-Diffusion-Models}{the official implementation}, where GSN guidance is applied exclusively during the first 25 sampling steps. The optimization uses a fixed learning rate of 20 without any decay. Syngen is specifically designed to process prompts composed only of entities with associated attributes. 

To improve generation quality, we discard cross-attention maps for the initial tokens \textit{``a photo of''}, as they degrade performance. All Syngen scores are computed using the official implementation with our patch. For GenEval only, we reimplemented Syngen within our framework to handle subject words composed of multiple tokens (\eg, ``stop sign'') by averaging attention maps across tokens belonging to the same entity.

\paragraph{Boxdiff~\cite{Xie_2023_ICCV}}

For BoxDiff, we have reimplemented the method to integrate it into our code, using the same hyperparameters. However, the scores may differ from those reported in the original implementation: \url{https://github.com/showlab/BoxDiff/tree/master}.

\paragraph{\saga, \sagaplus}

We apply 50 optimization steps using the SGD optimizer with a learning rate of $20$ and a batch size of 10. For \sagaplus, after the initial distribution is learned, we apply Syngen-GSN guidance once per sampling step until the 25th sampling step is reached. Our chosen hyperparameters are described in~\citeappendix{sec:hyperparameters}, precisely in~\citetable{tab:hparams} and in~\citetable{tab:hparams_var}.

\paragraph{\saga, \sagaplus and Bounding Boxes Loss \label{sec:lossbbox}}

We also apply 50 optimization steps using the SGD optimizer with a learning rate of $20$ and a batch size of 10 (hyperparameters provided in~\citetable{tab:hparams}). For \sagaplus, after the initial distribution is learned, we apply GSN guidance using the $\mathcal{L}$ criterion once per sampling step until the 25th sampling step is reached. For the GSN guidance, the learning rate starts at 20 and gradually decreases throughout the process.
For the bounding boxes conditioning, we replace the $\mathcal{L}_2$ with
\begin{equation}
    \mathcal{L}_2 =  \frac{1}{\vert \mathcal{C}\vert} \underset{(m, n) \in \mathcal{C}}{\sum} 1 - \left( \frac{\underset{i,j}{\sum} \min(\mathbf{B}^m_{i,j}, \mathbf{M}^n_{i,j})}{\underset{i,j}{\sum} (\mathbf{B}^m_{i,j} +\mathbf{M}^n_{i,j} )} \right)
\end{equation}
where $\mathbf{B}^i$ is the mask corresponding to the conditional bounding boxes with $1$ inside the bounding boxes and $0$ outside the $i$ subject.

\paragraph{Universal Guided Diffusion (UGD)~\cite{bansal2024universal}}
We used the code released with the paper that is available on github (\url{https://github.com/arpitbansal297/Universal-Guided-Diffusion}). To compare to Lottery Ticket and our model conditioned by boxes, we used the model guided by an object detector, which is a Faster R-CNN model with a ResNet-50-FPN backbone. The generative diffusion backbone is Stable Diffusion 1.4. We used an unconditional guidance scale of 2, repeated the self-recurrence $k=3$ times and the forward guidance strength follows $s(t)=100\cdot\sqrt{1-\alpha_t}$ as proposed by the authors. The number of DDIM steps has been set to 75 instead of 50 (for our approach and other ones) to try to improve the performance.

The results reported in \citetable{tab:bbox} use the same predefined boxes as the other approaches (see~\citesection{sec:datasets}).
The generation of one image takes around 46 seconds with a H100 GPU, 133 seconds with a A100-SXM4-80GB, 139 seconds with a A100-SXM4-40GB, 235 seconds with a V100-SXM2-32GB, and 407 seconds with a P100-SXM2-16GB. 

\paragraph{Lottery Ticket~\cite{mao2024theLottery}}

We use the official implementation\footnote{\url{https://github.com/UT-Mao/Initial-Noise-Construction}} with the default parameters of the paper. They generate images using DDIM scheduler with 50 sampling steps.

\subsection{Stable Diffusion 3 (\sdthree)\label{sec:sd3_appendix}}

We use the Stable Diffusion 3 Medium model from Hugging Face (\url{https://huggingface.co/stabilityai/stable-diffusion-3-medium-diffusers}). Images are generated at a $768\times768$ resolution using the `FlowMatchEulerDiscreteScheduler` (\url{https://huggingface.co/docs/diffusers/api/schedulers/flow_match_euler_discrete}) with 28 sampling steps. 
\citetable{tab:flow_matching_steps} provides the mapping between the discrete step index and its corresponding sampling step value $t$. We use a CFG scale of 7.0 \cite{ho2022classifierfree} and \textit{bfloat16} precision. Generations were performed on NVIDIA H100 GPUs, with an average time of 39 seconds for \saga.

\begin{table}[h!]
   
    \centering
     \caption{The mapping between the discrete step index (idx) and its corresponding sampling step value $t$.}
    {
    \small
    \npdecimalsign{.}
    \nprounddigits{1}
    \begin{tabular}{cr|cr|cr|cr}
        \toprule
        \textbf{idx} & $t$ & \textbf{idx} & $t$ & \textbf{idx} & $t$ & \textbf{idx} & $t$ \\
        \midrule
        0 & \numprint{1000.0}    & 7 & \numprint{895.9003}    & 14 & \numprint{737.0558} & 21 & \numprint{464.8760} \\
        1 & \numprint{987.3806}    & 8 & \numprint{877.3818}    & 15 & \numprint{707.2785} & 22 & \numprint{409.2888} \\
        2 & \numprint{974.1077}    & 9 & \numprint{857.6923}    & 16 & \numprint{675.0823} & 23 & \numprint{347.3926} \\
        3 & \numprint{960.1293}    & 10 & \numprint{836.7166}    & 17 & \numprint{640.1602} & 24 & \numprint{278.0487} \\
        4 & \numprint{945.3875}    & 11 & \numprint{814.3247}    & 18 & \numprint{602.1505} & 25 & \numprint{199.8269} \\
        5 & \numprint{929.8179}    & 12 & \numprint{790.3682}    & 19 & \numprint{560.625}  & 26 & \numprint{110.9057} \\
        6 & \numprint{913.3489}    & 13 & \numprint{764.6771}    & 20 & \numprint{515.0720} & 27 & \numprint{8.9285} \\
        \bottomrule
    \end{tabular}}
     
    \label{tab:flow_matching_steps}
\end{table}

\paragraph{Our implementation of GSN for SD3}

The Stable Diffusion 3 architecture does not include a dedicated cross-attention layer or the cross-attention mechanism found in the MM-DiT block. Attention is computed between image patches and the textual embeddings from T5 and CLIP. When handling attention mechanisms, the generated attention maps $\mathbf{M}$ are of dimension ${(hw + n_{\text{CLIP}} + n_{\text{T5}})^2 \times n_{\text{head}}}$. Here, $n_{\text{head}}$ indicates the number of attention heads, and $hw$ represents the dimensions of the patches obtained from the patchified $\mathbf{z}_t$. We then extract these attention maps where the image patches act as queries and the text embeddings serve as keys. Next, we process $\mathbf{M}$ by averaging across heads and excluding the start of text and end of text tokens. Following Attend\&Excite ~\cite{chefer2023attendandexcite}, we apply softmax and Gaussian smoothing on the preprocessed attention maps, then average T5 and CLIP attention maps, to obtain $\mathbf{M} \in \mathbb{R}^{hw \times \mathcal{S}}$ with $\mathcal{S}$ the subject token.

We apply 50 optimization steps using the SGD optimizer with a learning rate of $20$ and a batch size of 4. Our chosen hyperparameters are described in~\citeappendix{sec:hyperparameters}, precisely in~\citetable{tab:hparams} and in~\citetable{tab:hparams_var}.

\section{Hyperparameter study}
\label{sec:hyperparameters}
For the ablation study and hyperparameter selection, we generate two validation datasets following the TIAM benchmark methodology. Prompts are randomly composed from COCO labels, with 20 prompts drawn per set. To avoid overlap with test prompts, the sampling is repeated if necessary. The first validation set, for \sdone, includes 10 prompts with 2 entities and 10 with 3 entities; the second, for \sdthree, includes 10 prompts with 3 entities and 10 with 4 entities. We use distinct sets because \sdthree already performs well on 2 entity prompts, making 3 and 4 entity prompts more challenging.

For evaluation, 16 images are generated per prompt using fixed seeds, following TIAM recommendations. Evaluation is performed only on \sagaone, where a single distribution is trained and 16 images are sampled to reduce computational cost.

We use three automatic metrics for validation: CLIP Score, Aesthetic Score, and TIAM Score, which are more computationally efficient than the VQA Score that relies on large vision-language models.

To enable comparison, scores are normalized to the $[0,1]$ range, referred to as \textit{Normalized Performance}. We also report the \textit{Average Score}, which is defined as the mean of the normalized CLIP, TIAM, and Aesthetic Score.

Our initial small-scale experiments showed that introducing momentum can accelerate convergence and that the sampling step $t$, where the distribution is learned, affects image quality. However, excessive momentum can degrade fidelity and create a saturated image with artifacts, such as in \citefigure{fig:rescaling}. To identify optimal hyperparameters and study their interactions, we conduct experiments under a fixed optimization budget of 50 steps, consistent with prior GSN methods.

\subsection{Rescaling}
We evaluate momentum values from $0.0$ to $0.9$ in increments of $0.1$ and examine various sampling steps $t$. For \sdone, we test steps with indices from 1 to 21 (out of 50 total, see \citetable{tab:diffusion_steps_sd14}). For \sdthree, we test steps with indices from 1 to 12 (out of 28 total, see \citetable{tab:flow_matching_steps}).

We start by evaluating the impact of rescaling during training, applying the standard deviation of $\mathbf{\hat{z}}_0(\mathbf{z}_t, \mathbf{y}, t)$ on $\boldsymbol{\tilde{\mu}}_\mathbf{y}$, and comparing performance with and without this adjustment.

\begin{figure}
    \centering
    \includegraphics[width=1\linewidth]{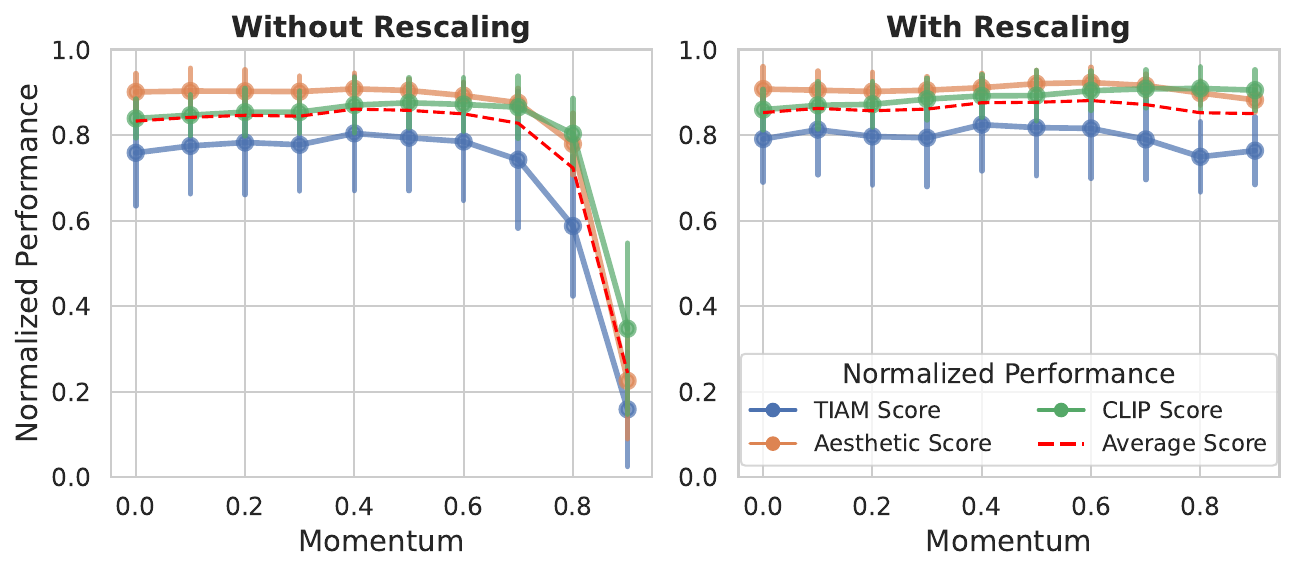}
   \caption{Normalized Performance of \sagaone as a function of the momentum value on \sdone. Each point represents the mean of each individual score (CLIP, Aesthetic, TIAM) across sampling steps, with error bars indicating the standard deviation. Left: results without rescaling; right: with rescaling. The red curve shows the trend of the average score.}  
    \label{fig:rescale_study_sd14}
\end{figure}

\begin{figure}
    \centering
    \includegraphics[width=1\linewidth]{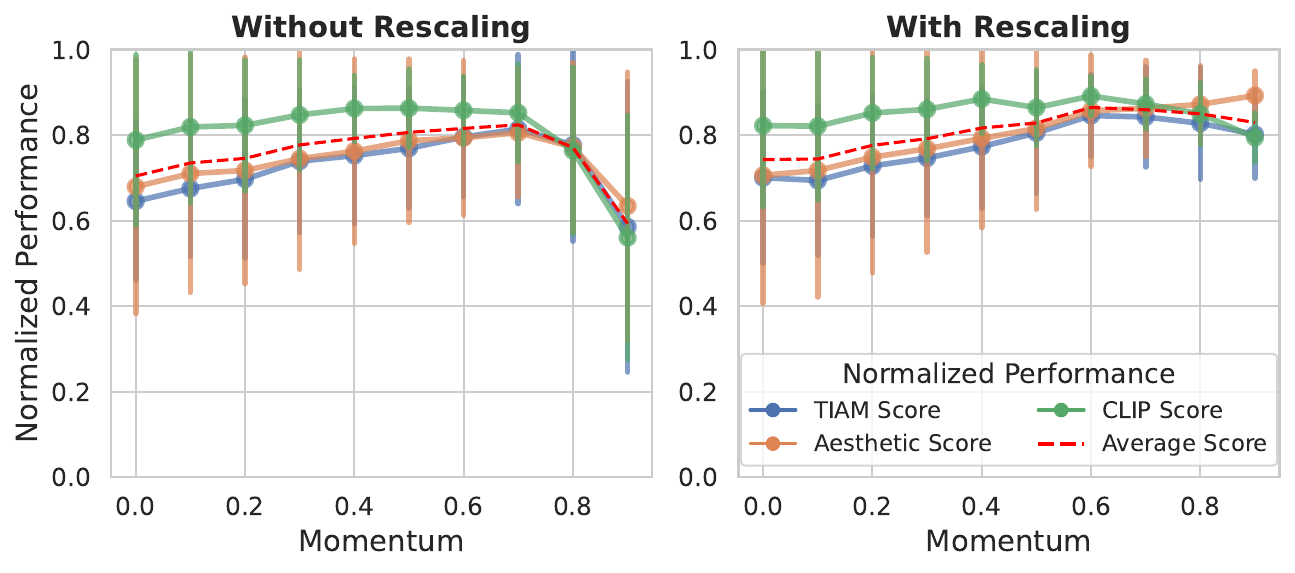}
    \caption{Normalized Performance of \sagaone as a function of the momentum value on \sdthree. Each point represents the mean of each individual score (CLIP, Aesthetic, TIAM) across sampling steps, with error bars indicating the standard deviation. Left: results without rescaling; right: with rescaling. The red curve shows the trend of the average score.}
    \label{fig:rescale_study_sd3}
\end{figure}

\citefigure{fig:rescale_study_sd14} and \citefigure{fig:rescale_study_sd3} show that our rescaling mechanism achieves performance comparable to the baseline for momentum values up to $0.6$ for \sdone and $0.7$ for \sdthree. Beyond these values, the baseline scores degrade, whereas our method maintains performance and exhibits a lower standard deviation. Since higher momentum accelerates convergence under a limited optimization budget, we adopt our rescaling mechanism for all subsequent experiments. 

\citefigure{fig:rescale_study_sd14} and \citefigure{fig:rescale_study_sd3} show that our rescaling mechanism achieves performance comparable to the baseline for momentum values up to 0.6 for \sdone and 0.7 for \sdthree. Beyond these thresholds, the baseline scores degrade, whereas our rescaling method maintains stable performance with a lower standard deviation. Although dispensing with rescaling is possible with careful hyperparameter tuning, our mechanism is crucial for ensuring stability when using aggressive parameters for rapid convergence. Since higher momentum accelerates convergence under a limited optimization budget, we adopt our rescaling mechanism for all subsequent experiments to prevent instabilities.

\subsection{Sampling Step}
\label{appendix:sampling_step}
\begin{figure}
    \centering
    \includegraphics[width=1\linewidth]{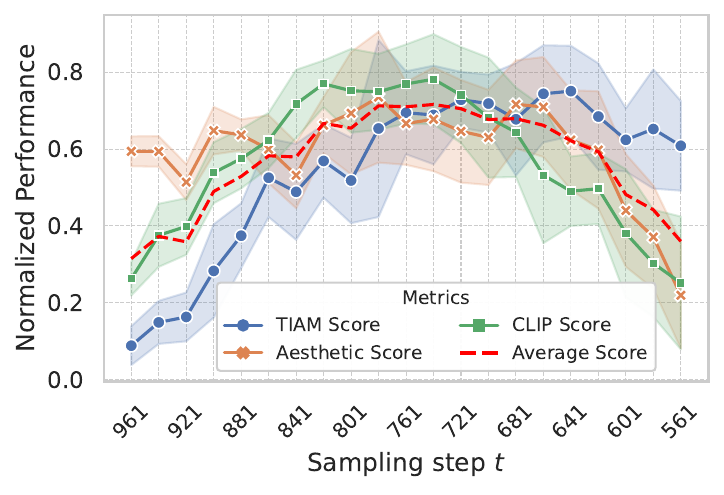}
    \caption{Normalized Performance of \sagaone on \sdone as a function of the sampling step. Points denote the mean performance averaged over various momentum values, with error bars indicating the standard deviation. The red curve highlights the Average Score.}
    \label{fig:study_t_sd14}
\end{figure}

\begin{figure}
    \centering
    \includegraphics[width=1\linewidth]{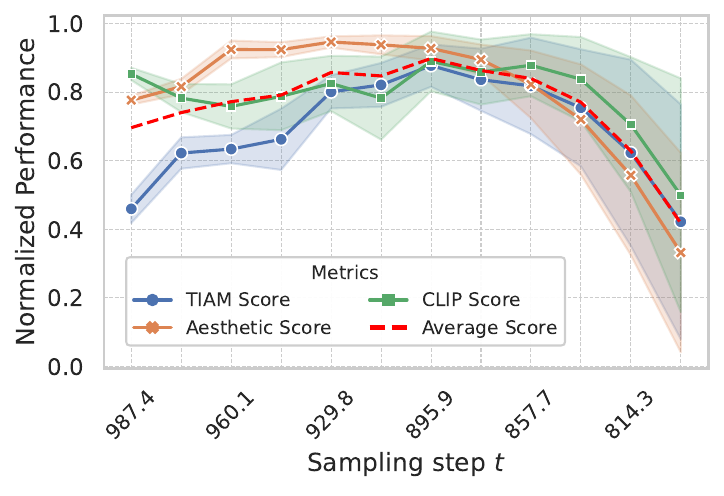}
    \caption{Normalized Performance of \sagaone on \sdthree as a function of the sampling step. Points denote the mean performance averaged over various momentum values, with error bars indicating the standard deviation. The red curve highlights the Average Score.}
    \label{fig:study_step_sd3}
\end{figure}

We analyze the impact of the sampling step $t$ used to learn the distribution. \citefigure{fig:study_t_sd14} and \citefigure{fig:study_step_sd3} plot the Normalized Performance as a function of $t$. For both models, performance initially improves with $t$, peaks at an optimal value, and then sharply degrades.
This decline likely occurs because learning the distribution from a large sampling step $t$ makes the regression task for $\boldsymbol{\mu}_\mathbf{y}$ too difficult. Given a limited optimization budget, the process must determine a wide range of structural frequencies, starting from only the DC component (average color).
This suggests the existence of an optimal sampling step that balances task difficulty and performance.

\begin{figure*}[]
    \centering

    \includegraphics[width=1\linewidth]{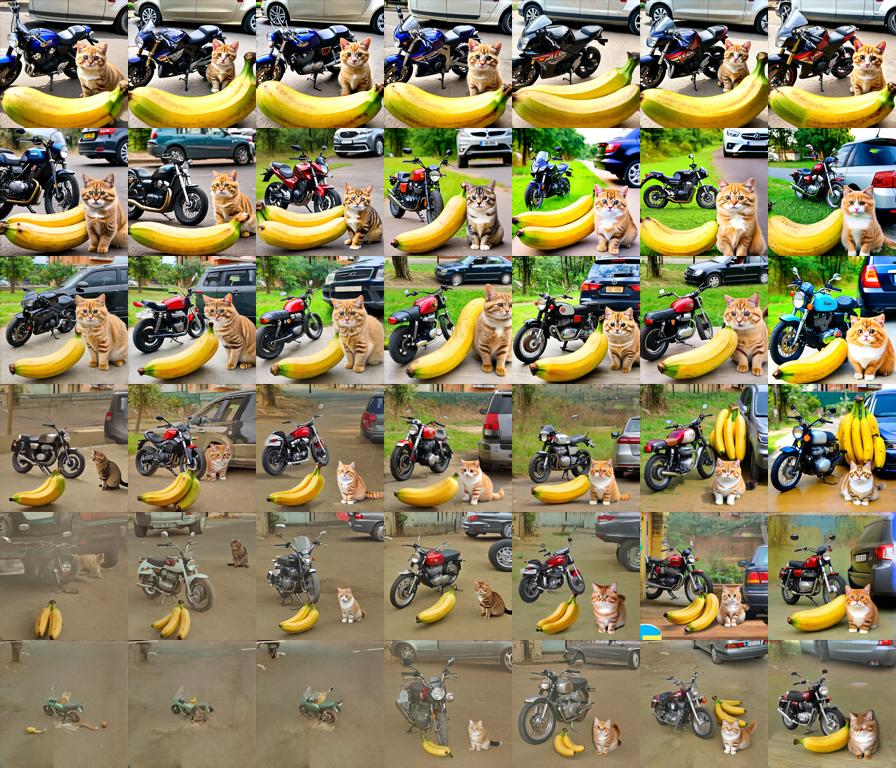}
    \caption{Samples for the prompt \textit{``A photo of a motorcycle next to a banana with a car and a cat,''} generated using \sagaone on \sdthree with a fixed seed. The grid illustrates the effect of varying the momentum (x-axis) and the sampling step $t$ (y-axis). From left to right, the momentum values are $0, 0.2, 0.4, 0.6, 0.7, 0.8,$ and $0.9$. From top to bottom, the sampling steps $t$ are $987.4, 960.1, 929.8, 895.9, 857.7,$ and $814.3$. We observe that with low momentum, the generation struggles to converge around sampling step $t \leq 860$. Conversely, images generated with high momentum appear more colorful.}
    \label{fig:momentum_step_sdthree_a}
\end{figure*}

Additionally, \citefigure{fig:momentum_step_sdthree_a} qualitatively illustrates how generated images for the same prompt and seed are affected by different momentum values and sampling steps $t$. Excessive momentum can lead to color saturation in the generated images, whereas insufficient momentum, particularly at larger steps $t$, may result in poor convergence and images with monotonous coloration.

\subsubsection{Best Hyperparameters}

\begin{table}[htbp]

  \centering
   \caption{Momentum values are explored in the range $[0, 0.9]$ with increments of 0.1 and sampling steps in intervals of 1. All hyperparameter experiments are conducted using \sagaone, learning one distribution per prompt and generating 16 images from it. Results on the test dataset show comparable scores when using one distribution per prompt versus one per seed. Idx Steps and Idx denote the sampling step's index. The corresponding step value can be found by cross-referencing this index with the appropriate mapping table: \citetable{tab:diffusion_steps_sd14} for \sdone and \citetable{tab:flow_matching_steps} for \sdthree.}
  \small
  \begin{tabular}{cclcccc}
    \toprule
     & \multirow{2}{*}{\textbf{Cond.}} & \multirow{2}{*}{\textbf{Method}} & \multirow{2}{*}{\textbf{Idx Steps}} & \multirow{2}{*}{\textbf{Moment.}} & \multicolumn{2}{c}{\textbf{ Params}} \\
    &&&& & \textbf{Idx} & $m$ \\
    \midrule
    \multirow[c]{4}{*}{\rotatebox{90}{\textbf{\sdone}}} & \multirow[c]{2}{*}{\rotatebox{90}{\textbf{Text}}} & \saga & $[1, 21]$ & $[0, 0.9]$ & $10$&$0.4$ \\
    
    & & \sagaplus & $[1, 14]$ & $[0, 0.9]$ & $6$&$0.1$ \\
    \cmidrule(rl){2-7}
    & \multirow[c]{2}{*}{\rotatebox{90}{\textbf{Bbox}}} & \saga & $[1, 12]$ & $[0, 0.9]$ & $11$&$0.5$ \\
    & & \sagaplus & $[1, 12]$ & $[0, 0.9]$ & $12$ &$0.1$ \\
    \midrule
    \multirow[c]{2}{*}{\rotatebox{90}{\textbf{\sdthree}}} & \multirow[c]{2}{*}{\rotatebox{90}{\textbf{Text}}} & \multirow{2}{*}{\saga} & \multirow{2}{*}{$[1, 12]$} & \multirow{2}{*}{$[0, 0.9]$} & $5$&$0.7$ \\
    & & & & & $7$&$0.4$ \\
    \bottomrule
  \end{tabular}
    
  \label{tab:hparams}
\end{table}

We select the optimal hyperparameters for each approach based on the Average Score. We conducted a dedicated study for each hyperparameter set, and a summary of the explored search space and the final selected configurations is provided in \citetable{tab:hparams}.

For \sdthree, we evaluate two distinct configurations. The first, featured in the main paper, uses a sampling step index of idx $=5$ and a momentum of $0.7$, which achieves a high aesthetic score. The second configuration, referred to as \variantB in \citesection{sec:evaluation_appendix}, uses a step index of idx $=7$ and a momentum of $0.4$. This latter setup yields the best overall Average Score at the cost of a slightly lower Aesthetic Score, as detailed in \citefigure{fig:study_step_sd3}. The mapping from indices to sampling step values is provided in \citetable{tab:flow_matching_steps}.

\subsection{Variance}
\label{sec:estim_variance}

The study is still conducted with the generation of 16 images per prompt on the validation dataset. We analyze the effect of the learning of $\mathbf{\Sigma_y}$ by examining the Average Score.

For a latent space of size $h\times w\times c$, the covariance matrix has a size $hwc\times hwc$ (set to

$64\times 64\times 4$ for \sdone and $96\times 96\times 16$ for \sdthree in our case), which is huge and hard to  estimate fully. In the experiments above, we estimate  $\boldsymbol{\tilde{\mu}}_\mathbf{y}$ only and thus assume that $a_t^2\boldsymbol{\Sigma}_\mathbf{y}$ is negligible w.r.t $b_t^2\mathbf{I}$. 
We tried slightly more complex setups, by learning $\boldsymbol{\Sigma}_\mathbf{y}$ with 1 to $whc$ values to estimate. All the models were tested with an estimation shared across the channels of the latent or per-channel. With $\mathbf{\Sigma_y}^{\text{diag}}$ we kept the actual variances of each value but set the off-diagonal values to $0$, leading to $wh$ values (shared) or $whc$ values (per-channel) to estimate. With $\boldsymbol{\Sigma_y}^{\text{uniq}}$ we estimated one value for the full matrix (shared) or $c$, one for each channel of the latent codes. Finally, we estimated $\boldsymbol{\Sigma}_\mathbf{y}^{b\times b}$ by considering blocks of size $b\times b$ along the diagonal, corresponding to some neighborhoods of the latent values (we report the results for $b=2, 4,8, 16, 32$ for \sdone and $b = 2, 3, 8, 16, 32, 48$ for \sdthree).

Hence $\boldsymbol{\Sigma}_\mathbf{y}^{\text{diag}}=\boldsymbol{\Sigma}_\mathbf{y}^{1\times 1}$, $\frac{wh}{b}$ (shared) or $\frac{whc}{b}$ (per-channel) different values.

\begin{figure}
    \centering
    \includegraphics[width=1\linewidth]{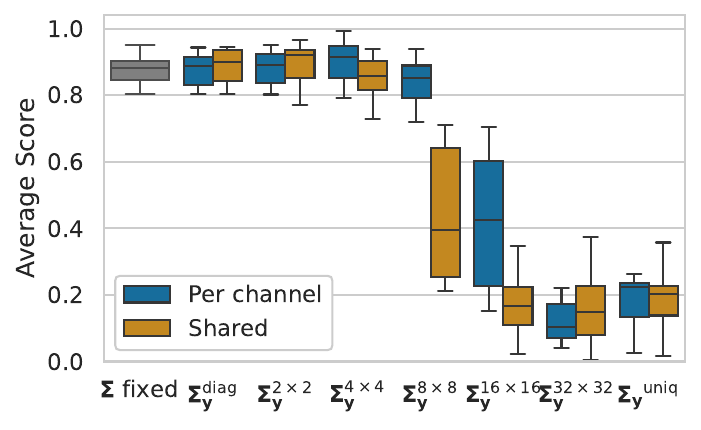}
    \caption{Ablation on the covariance matrix modeling for \saga on \sdone.}
    \label{fig:var_average_saga_sd14}
\end{figure}

For \sdone We study the variance setup using sampling steps with indices idx$\in \left[1, 12\right]$ (mapping from these indices to their corresponding step values is detailed in \citetable{tab:diffusion_steps_sd14}). We keep the momentum fixed at $0.4$ to simplify the optimization. As shown in \citefigure{fig:var_average_saga_sd14}, we observe a slight performance improvement for $b=4$, particularly when using \textit{per-channel} parameters.

\begin{figure}
    \centering
    \includegraphics[width=1\linewidth]{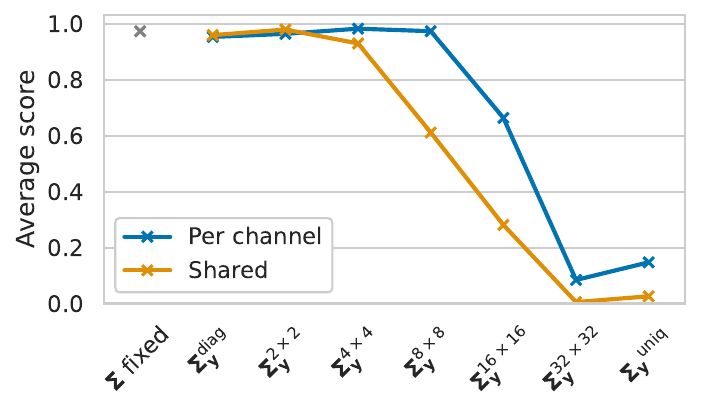}
    \caption{Ablation on the covariance matrix modeling for \sagaplus on \sdone.}
    \label{fig:var_average_sagaplus_sd14}
\end{figure}

\begin{figure}
    \centering
    \includegraphics[width=1\linewidth]{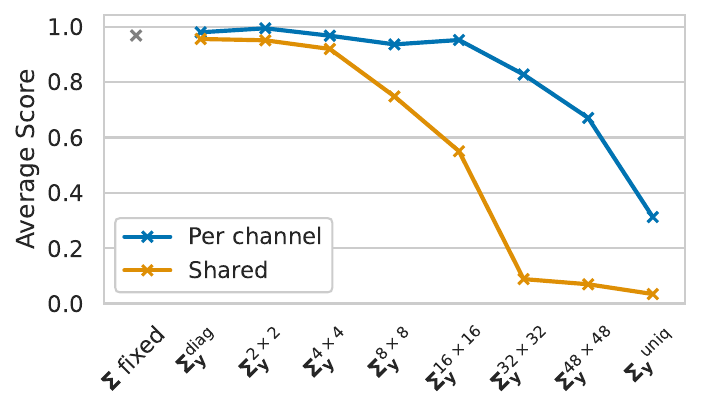}
    \caption{Ablation study on covariance matrix modeling for \saga on \sdthree, conducted with a fixed sampling step index of idx $=5$ and a momentum of $0.7$ (see \citetable{tab:flow_matching_steps} for the full index-to-value mapping).}
    \label{fig:var_average_saga_sd3_step_5}
\end{figure}

\begin{figure}
    \centering
    \includegraphics[width=1\linewidth]{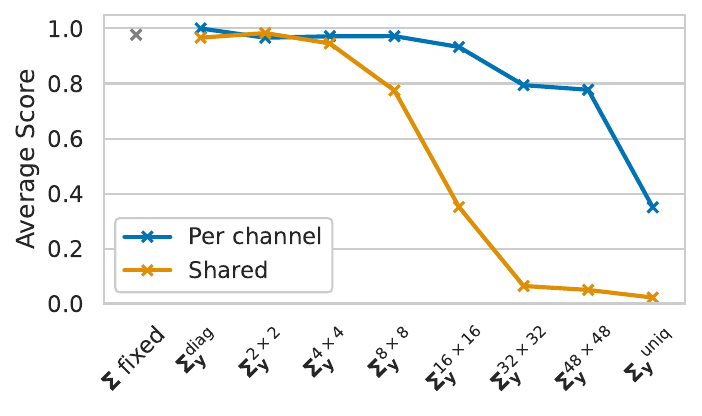}
    \caption{Ablation study on covariance matrix modeling for \saga on \sdthree, conducted with a fixed sampling step index of idx $=7$ and momentum of $0.4$ (see \citetable{tab:flow_matching_steps} for the full index-to-value mapping).}
    \label{fig:var_average_saga_sd3_step_7}
\end{figure}

\begin{table}[htbp]

  \centering
    \caption{Optimal hyperparameters for our methods. The learning is a \textit{per-channel} setup for all the configurations. Idx Steps denotes the sampling step's index. The corresponding step value can be found by cross-referencing this index with the appropriate mapping table: \citetable{tab:diffusion_steps_sd14} for \sdone and \citetable{tab:flow_matching_steps} for \sdthree.}
  {\small
  \begin{tabular}{clcccc}
    \toprule
     & \textbf{Method} & Idx Steps & Moment. &  $\boldsymbol{\Sigma_\mathbf{y}}$ \\
    \midrule
    \multirow[c]{2}{*}{\textbf{\sdone}} & \saga & $12$ & $0.4$ & $\boldsymbol{\Sigma}_\mathbf{y}^{4\times 4}$  \\
    & \sagaplus & $6$ & $0.1$ &  $\boldsymbol{\Sigma}_\mathbf{y}^{4\times 4}$\\
    \midrule
    \multirow[c]{2}{*}{\textbf{\sdthree}} &\multirow{2}{*}{\saga} & $5$&$0.7$ &  $\boldsymbol{\Sigma}_\mathbf{y}^{2\times 2}$ \\
    & & $7$&$0.4$ & $\mathbf{\Sigma_y}^{\text{diag}}$ \\
    \bottomrule
  \end{tabular}}
  
  \label{tab:hparams_var}
\end{table}

This trend is consistent for \sagaplus on \sdthree, where we evaluate various setups with fixed $t$ and momentum. Per-channel learning consistently outperforms parameter sharing, and performance degrades as the number of learned parameters decreases (\citefigure{fig:var_average_sagaplus_sd14}, \citefigure{fig:var_average_saga_sd3_step_5}, \citefigure{fig:var_average_saga_sd3_step_7}). We conclude that learning per-channel parameters is optimal, though it presents a trade-off in the degree of parameter sharing. The best-performing hyperparameters for our variance setups are presented in \citetable{tab:hparams_var}.

\begin{figure}[]
    \centering
    \includegraphics[width=1\linewidth]{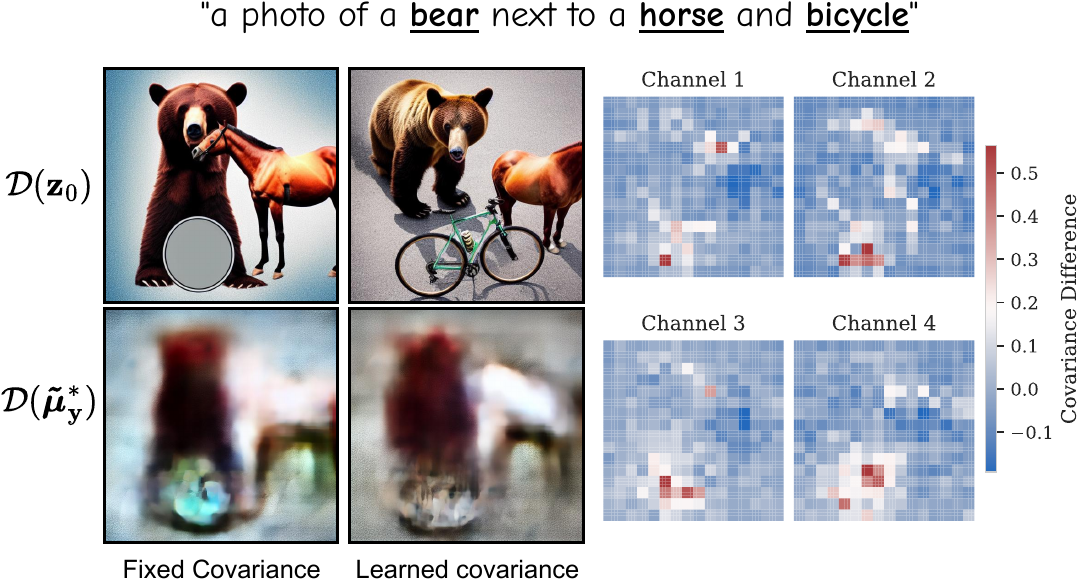}
    \caption{Samples generated by \saga and \sagavar with the \sdone backbone, comparing a fixed versus a learned covariance $\boldsymbol{\Sigma_\mathbf{y}}$. We visualize the difference between the learned covariance, $\boldsymbol{\tilde{\Sigma}}_\mathbf{y}$, and its fixed counterpart, $b_{t}^{2}\mathbf{I}$: $\boldsymbol{\tilde{\Sigma}}_\mathbf{y} - b_{t}^{2}\mathbf{I}$. Since the learned covariance is diagonal, its variance is plotted per channel, effectively controlling the per-point noise level. Learning the covariance increases stochasticity on the main subjects (the bicycle and the horse) while reducing it in other regions to better preserve the signal $\boldsymbol{\tilde{\mu}}_\mathbf{y}$.}
    \label{fig:covariance_vs_wo}
\end{figure}

In \citefigure{fig:covariance_vs_wo}, we show an example of the benefit of learning the covariance.

\begin{table*}[]
    \centering   
     \caption{Raw score difference between the methods with the learning of the covariance and without across various metrics. A positive value indicates that the method with the learning of $\boldsymbol{\Sigma_y}$ outperformed the one with only the learning of $\boldsymbol{\tilde{\mu}}_\mathbf{y}$.}

    {\small
    \setlength{\tabcolsep}{1mm}
\begin{tabular}{llccccccccccccccccccccc}
    \toprule
    & \multirow{2}{*}{Methods} & \multicolumn{3}{c}{TIAM} & \multicolumn{3}{c}{VQA Score} & \multicolumn{3}{c}{CLIP Score} & \multicolumn{3}{c}{Min. Obj. Sim.} & \multicolumn{3}{c}{ Text-Text Sim.} & \multicolumn{3}{c}{Full Prompt Sim.} & \multicolumn{3}{c}{Aesthetic Score}  \\
     &  & 2 & 3 & 4 & 2 & 3 & 4 & 2 & 3 & 4 & 2 & 3 & 4 & 2 & 3 & 4 & 2 & 3 & 4 & 2 & 3 & 4  \\
    \midrule
    \multirow{2}{*}{\textbf{\sdone}} & \saga & 1.2 & 4.8 & 2.2 & -1.1 & 2.8 & 3.0 & 0.1 & 0.3 & 0.2 & 0.3 & 0.5 & 0.4 & 0.4 & 0.5 & 0.2 & 0.1 & 0.3 & 0.3 & -0.0 & -0.0 & -0.0  \\
    & \sagaplus & 0.6 & 2.2 & 1.7 & -0.1 & 0.8 & 0.3 & 0.0 & 0.0 & 0.1 & 0.1 & 0.0 & 0.0 & 0.2 & 0.2 & 0.3 & 0.0 & 0.0 & 0.1 & 0.0 & 0.0 & 0.0 \\
    \midrule
    \textbf{\sdthree} & \saga & -0.4 & -0.2 & 0.8 & -0.0 & -0.1 & -0.2 & -0.0 & 0.1 & 0.0 & -0.0 & 0.1 & 0.0 & -0.1 & 0.0 & 0.1 & -0.0 & 0.1 & 0.0 & 0.0 & -0.0 & -0.0 \\
    \midrule
    \multicolumn{2}{c}{Mean} & 0.5 & 2.2 & 1.6 & -0.4 & 1.2 & 1.1 & 0.0 & 0.1 & 0.1 & 0.1 & 0.2 & 0.1 & 0.2 & 0.2 & 0.2 & 0.1 & 0.1 & 0.1 & 0.0 & -0.0 & -0.0 \\
    \bottomrule
\end{tabular}

    }
    
    \label{tab:difference_var}
\end{table*}

Finally, \citetable{tab:difference_var} quantifies the impact of learning the covariance by presenting the raw performance difference between paired methods on the test datasets. This component yields consistent performance improvements for \sdone across both configurations, demonstrating the approach's potential. For \sdthree, however, the results are more mixed, showing negligible performance differences. We also note that this extension does not affect the Aesthetic Score for either model. We hypothesize that the benefits could be further amplified through more extensive hyperparameter tuning; for instance, by using distinct learning rates for the covariance $\boldsymbol{\Sigma_y}$ and the mean $\boldsymbol{\tilde{\mu}}_\mathbf{y}$, or by increasing the number of optimization steps. A thorough exploration of these possibilities is left for future work. It is crucial to note that the metrics presented are heterogeneous and do not share a common scale. For a metric like TIAM, which represents the percentage of images where entities were correctly detected, the raw difference can be interpreted directly; for instance, a value of +5.0 indicates a 5 percentage point increase in the detection rate. However, for other metrics such as VQA, CLIP score, or aesthetics, the scores are derived from different models and may not be normalized. Therefore, while the direction (positive or negative) and relative magnitude of the change within a single metric are informative, a direct comparison of the raw differences between different metrics should be approached with caution, as a similar numerical change may correspond to a vastly different practical impact depending on the metric's specific scale and distribution.

\section{Evaluation Details and Complementary Results}
\label{sec:evaluation_appendix}

This section details the datasets and automatic metrics used. We also report additional results for all our configurations on the TIAM and VQA scores, as well as the CLIP score, Aesthetic score, Attend\&Excite metrics, and GenEval benchmark.

\subsection{Datasets}\label{sec:datasets}

\paragraph{Prompts}
The authors of TIAM (see Section~\ref{sec:tiam}) provide some benchmarks on (\url{https://github.com/CEA-LIST/TIAMv2})
of 300 prompts, with 2, 3 or 4 objects or animals. They also provide 300 other prompts where each entity (object or animal) is associated with a color attribute. The object or animals being in the set of the 80 COCO classes, the metric relies on yolo-v8 pretrained on this dataset to detect them in the synthetic image.

The validation datasets are available at:
\begin{itemize}
    \item 2-3 entities: \url{https://huggingface.co/datasets/Paulgrimal/validation_set_2_3_entities}
    \item 3-4 entities: \url{https://huggingface.co/datasets/Paulgrimal/validation_set_3_4_entities}
\end{itemize}

The test sets are available at the following locations:
\begin{itemize}
    \item 2 entities: \url{https://huggingface.co/datasets/Paulgrimal/2_entities_bbox}
    \item 3 entities: \url{https://huggingface.co/datasets/Paulgrimal/3_entities_bbox}
    \item 4 entities: \url{https://huggingface.co/datasets/Paulgrimal/4_entities}
\end{itemize}

\paragraph{Bounding-boxes}
\label{appendix:bboxes}
To extend the dataset with bounding box annotations, we use LLaMA-3.1-8B-Instruct (\url{https://huggingface.co/meta-llama/Llama-3.1-8B-Instruct}) to generate candidate bounding boxes for a given prompt, applied to datasets with 2 and 3 entities. We verify the coherence of the predicted bounding box coordinates. The system prompt used to generate the boxes is as follows:

\begin{lstlisting}[mathescape=true, frame=single, basicstyle=\ttfamily\small, breaklines=true]
You are a specialized system for generating non-overlapping bounding boxes (bbox) in a 16x16 coordinate space.

COORDINATE SYSTEM:
- Format: [x1, y1, x2, y2] following COCO dataset convention where:
  * (x1,y1): top-left corner coordinates
  * (x2,y2): bottom-right corner coordinates
  * All values MUST be integers between 0-15
  * Required: x1 < x2 and y1 < y2

BOX CONSTRAINTS:
1. Create logically sized boxes (width and height can vary)
2. Boxes can be next to each other but cannot share any coordinates
3. Use space efficiently while maintaining natural proportions
4. Example of valid adjacent boxes: box1[x2] + 1 = box2[x1]

VALIDATION RULES:
1. ALL coordinates must be integers: 0  $\le$ coord $\le$ 15
2. NO overlapping between boxes
3. NO shared coordinates between boxes
4. Each box must have distinct area (width > 0, height > 0)

EXAMPLES:

1. Valid adjacent boxes:
Input: ["mouse", "keyboard"]
Output:
{
    "mouse": [2, 0, 3, 2],
    "keyboard": [4, 1, 8, 3]
}

2. Multiple boxes:
Input: ["cup", "book", "phone"]
Output:
{
    "cup": [1, 1, 3, 3],
    "book": [4, 1, 6, 3],
    "phone": [7, 1, 9, 3]
}

INSTRUCTIONS:
- Generate bbox coordinates for the given entities
- Return ONLY valid JSON format
- No additional text or explanations}
\end{lstlisting}

The validation dataset (for 2 and 3 entities) is available at:
\url{https://huggingface.co/datasets/Paulgrimal/validation_set_2_3_entities_bbox}

The test sets are available at the following locations:
\begin{itemize}
    \item 2 entities: \url{https://huggingface.co/datasets/Paulgrimal/2_entities_bbox}
    \item 3 entities: \url{https://huggingface.co/datasets/Paulgrimal/3_entities_bbox}
\end{itemize}

\subsection{TIAM}\label{sec:tiam}
The TIAM metric~\cite{Grimal_2024_WACV} estimates the extent to which the objects specified in the prompt and their colors are actually visible in the synthetic images. It thus essentially focuses on penalizing \textit{catastrophic neglect} and \textit{attribute binding}, although it indirectly also penalizes errors due to \textit{object mixing}. Similarly to the VQA score, it relies on an external model to detect the objects, namely yolo-v8. The relative order of the approaches, or the score of a given approach on different benchmarks, is usually the same for both metrics. However, TIAM reports a larger range of scores than the VQA scores and is easier to interpret.

We used the official code released\footnote{\url{https://github.com/CEA-LIST/TIAMv2}} and the benchmarks described above. Each benchmark contains 300 prompts, and 16 images have been generated for each of them; thus the final score for each benchmark results from $4,800$ synthetic images. In all cases, we used the same threshold as in the original paper to report the score, namely 0.25.

All the performances of the models are reported in \citetable{tab:tiam_vqa_long} (prompt only) and \citetable{tab:bbox_appendix} (box conditioning).

\begin{table*}[]
    \centering
    \caption{Comparison of our method with vanilla sampling and state-of-the-art training-free methods on \sdone and \sdthree using TIAM and VQA scores, both multiplied by~$\times 100$. Methods with \variantB are another configuration we tested (see \citesection{sec:estim_variance}).}
    \begin{tabular}{llrrrrrr}
    \toprule
    & \multirow{2}{*}{Methods} & \multicolumn{3}{c}{TIAM} & \multicolumn{3}{c}{VQA Score} \\
    & & 2 & 3 & 4 & 2 & 3 & 4 \\
    \midrule
\multirow[c]{13}{*}{\rotatebox{90}{\textbf{\sdone}}} & Stable Diffusion & 45.4 & 8.4 & 1.0 & 61.3 & 31.9 & 23.5 \\
 & InitNO \textsubscript{(CVPR'24)}& 62.1 & 14.2 & 1.2 & 73.5 & 37.9 & 23.6 \\
 & \saga & 74.7 & 32.3 & 6.8 & 83.7 & 56.6 & 34.5 \\
 & \sagaone & 75.8 & 31.0 & 5.9 & 82.7 & 55.3 & 33.4 \\
 & \sagavar & 75.9 & 37.1 & 9.0 & 82.6 & 59.3 & 37.5 \\
 & \sagaonevar & 76.3 & 34.4 & 8.8 & 82.6 & 57.5 & 37.3 \\
\cmidrule(lr){2-8}
 & Attend\&Excite\textsubscript{(SIGGRAPH’23)} & 71.4 & 32.0 & 10.1 & 85.7 & 65.2 & 49.8 \\
 & InitNO+\textsubscript{(CVPR'24)} & 75.3 & 33.0 & 9.8 & 87.0 & 65.0 & 48.0 \\
 & Syngen\textsubscript{(NeurIPS'23)} & 78.5 & 39.2 & 13.1 & 85.4 & 63.4 & 47.3 \\
 & \sagaplus & \underline{85.5} & 50.7 & 17.9 & \textbf{88.3} & \underline{70.5} & 51.1 \\
 & \sagaplusone & 84.8 & 48.8 & 18.0 & 87.3 & 69.7 & 50.9 \\
 & \sagaplusvar & \textbf{86.1} & \textbf{52.9} & \textbf{19.6} & \underline{88.1} & \textbf{71.3} & \textbf{51.5} \\
 & \sagaplusonevar & 84.8 & \underline{51.7} & \underline{19.0} & 86.9 & 69.5 & \underline{51.5} \\
\midrule
\multirow[c]{9}{*}{\rotatebox{90}{\textbf{\sdthree}}} & Stable Diffusion\textsubscript{(ICML'24)} & 84.3 & 62.3 & 32.2 & 90.5 & 78.6 & 65.7 \\
 & \sagaStepsevenMomentumzerodotfour & 88.7 & 83.1 & 60.3 & 93.8 & 87.4 & 80.5 \\
 & \sagaStepfiveMomentumzerodotseven & 87.0 & 80.0 & 63.2 & 93.5 & 86.4 & 81.2 \\
 & \sagaoneStepsevenMomentumzerodotfour & \textbf{90.0} & \underline{83.7} & 61.0 & \textbf{94.7} & \underline{88.2} & 80.9 \\
 & \sagaoneStepfiveMomentumzerodotseven & 88.6 & 81.7 & \textbf{64.7} & 94.2 & 86.9 & 81.7 \\
 & \sagavarStepsevenMomentumzerodotfour & 88.1 & 83.4 & 61.4 & 93.9 & 87.4 & 80.7 \\
 & \sagavarStepfiveMomentumzerodotseven & 86.6 & 79.8 & \underline{64.0} & 93.4 & 86.3 & 80.9 \\
 & \sagaonevarStepsevenMomentumzerodotfour & \underline{90.0} & \textbf{84.7} & 61.3 & \underline{94.5} & \textbf{88.3} & \underline{81.7} \\
 & \sagaonevarStepfiveMomentumzerodotseven & 88.2 & 82.4 & 63.6 & 94.5 & 87.4 & \textbf{82.4} \\
\bottomrule
\end{tabular}

    \label{tab:tiam_vqa_long}
    
\end{table*}

\begin{table}[tb]
    \centering
    \caption{Results of the bounding box-conditioned methods on \sdone (scores by number of entities in the prompt: 2, 3)}
    {
    \setlength{\tabcolsep}{1mm}
    \begin{tabular}{lrrrr}
\toprule
\multirow{2}{*}{Methods} & \multicolumn{2}{c}{TIAM} & \multicolumn{2}{c}{VQA Score} \\
 & 2 & 3 & 2 & 3 \\
\midrule
Lottery Tickets & 42.5 & 8.4 & 58.6 & 31.3 \\
UGD & 38.3 & 6.2 & 66.0 & 42.0 \\
BoxDiff & 57.4 & 18.0 & 78.5 & 53.3 \\
\sagaonebbox & 73.5 & 32.7 & 81.9 & 56.0 \\
\sagabbox & 75.1 & 34.4 & 83.5 & 59.6 \\
\sagaonebboxgsn & \textbf{78.9} & \underline{38.3} & \textbf{85.5} & \underline{62.5} \\
\sagabboxgsn & \underline{78.0} & \textbf{40.5} & \underline{85.4} & \textbf{64.8} \\
\bottomrule
\end{tabular}

    }
    \label{tab:bbox_appendix}
    
\end{table}

\subsection{VQA score}
The VQA score \cite{linvqascore2025} is an automatic metric that ensures textual alignment by assessing consistency between the prompt and the generated image.
Each prompt is transformed into a question of the form ``Does this
figure show {prompt}?'', which the answer should be \texttt{yes} or \texttt{no}. Then, the approach consists in computing the probability of the \texttt{yes} answer, which is considered as the alignment score. As a VQA model, it combines a pre-trained CLIP vision-encoder with a pre-trained
FlanT5 model~\cite{chung2024flanT5}. All the performances of the models are reported in \citetable{tab:tiam_vqa_long} (prompt only) and \citetable{tab:bbox_appendix} (box conditioning).

\subsection{CLIP Score and Aesthetic Score}
\begin{table*}[tbh]
    \centering
     \caption{Aesthetic and CLIP scores for all variants of our approach and the baselines presented in the main manuscript. Methods with \variantB are another configuration we tested (see \citesection{sec:estim_variance}).} 
    \begin{tabular}{llrrrrrr}
    \toprule
    & \multirow{2}{*}{Methods} & \multicolumn{3}{c}{Aesthetic Score} & \multicolumn{3}{c}{CLIP Score} \\
    & & 2 & 3 & 4 & 2 & 3 & 4 \\
    \midrule
\multirow[c]{13}{*}{\rotatebox{90}{\textbf{\sdone}}} & Stable Diffusion & 5.48 & \textbf{5.47} & \textbf{5.46} & 32.19 & 33.54 & 34.00 \\
 & InitNO \textsubscript{(CVPR'24)}& 5.50 & 5.44 & 5.41 & 33.11 & 34.26 & 34.24 \\
 & \saga & \textbf{5.51} & \underline{5.47} & 5.45 & 33.99 & 35.66 & 35.33 \\
 & \sagaone & \underline{5.50} & 5.46 & \underline{5.46} & 33.98 & 35.50 & 35.20 \\
 & \sagavar & 5.50 & 5.46 & 5.42 & 34.04 & 35.95 & 35.56 \\
 & \sagaonevar & 5.50 & 5.45 & 5.43 & 34.04 & 35.70 & 35.45 \\
\cmidrule(lr){2-8}
 & Attend\&Excite\textsubscript{(SIGGRAPH’23)} & 5.48 & 5.38 & 5.30 & 34.02 & 35.94 & 36.46 \\
 & InitNO+\textsubscript{(CVPR'24)} & 5.50 & 5.36 & 5.23 & 34.14 & 35.95 & 36.36 \\
 & Syngen\textsubscript{(NeurIPS'23)} & 5.40 & 5.41 & 5.36 & 34.14 & 36.53 & 37.08 \\
 & \sagaplus & 5.29 & 5.31 & 5.27 & \underline{34.28} & \underline{36.83} & \underline{37.25} \\
 & \sagaplusone & 5.30 & 5.32 & 5.28 & 34.17 & 36.72 & 37.15 \\
 & \sagaplusvar & 5.30 & 5.32 & 5.28 & \textbf{34.29} & \textbf{36.86} & \textbf{37.34} \\
 & \sagaplusonevar & 5.31 & 5.32 & 5.28 & 34.16 & 36.79 & 37.19 \\
\midrule
\multirow[c]{9}{*}{\rotatebox{90}{\textbf{\sdthree}}} & Stable Diffusion\textsubscript{(ICML'24)} & 5.51 & 5.50 & 5.47 & \textbf{34.48} & 37.57 & 38.47 \\
 & \sagaStepsevenMomentumzerodotfour & 5.56 & 5.63 & 5.64 & 34.35 & 37.64 & 39.06 \\
 & \sagaStepfiveMomentumzerodotseven & 5.63 & 5.67 & 5.68 & 34.02 & 37.25 & 38.77 \\
 & \sagaoneStepsevenMomentumzerodotfour & 5.57 & 5.63 & 5.65 & 34.45 & \textbf{37.73} & \underline{39.12} \\
 & \sagaoneStepfiveMomentumzerodotseven & 5.63 & \textbf{5.68} & \textbf{5.69} & 34.11 & 37.32 & 38.85 \\
 & \sagavarStepsevenMomentumzerodotfour & 5.56 & 5.63 & 5.64 & 34.31 & 37.63 & 39.05 \\
 & \sagavarStepfiveMomentumzerodotseven & \underline{5.63} & 5.67 & 5.67 & 34.00 & 37.31 & 38.79 \\
 & \sagaonevarStepsevenMomentumzerodotfour & 5.57 & 5.63 & 5.65 & \underline{34.46} & \underline{37.70} & \textbf{39.13} \\
 & \sagaonevarStepfiveMomentumzerodotseven & \textbf{5.65} & \underline{5.67} & \underline{5.69} & 34.16 & 37.29 & 38.76 \\
\bottomrule
\end{tabular}

    \label{tab:clipscore}
\end{table*}

We compute the LAION Aesthetic score~\cite{schuhmann2022laion5bopenlargescaledataset}, which rates image aesthetics on a 1-10 scale. For the CLIP score, we use the OpenAI CLIP model\footnote{\label{foot:clip}\url{https://huggingface.co/openai/clip-vit-base-patch16}}~\cite{radford2021learning}, measuring cosine similarity between text and image embeddings. Results are shown in \citetable{tab:clipscore}, where \saga and its variants outperform other baselines.
 
We first note that the aesthetic scores remain close to the base model and other baselines, indicating that our method does not degrade image quality. For \sdthree, image quality is even improved across all configurations. We thus conclude that our approach preserves visual fidelity.

We now analyze the CLIP score. On \sdone, \saga outperforms both InitNO and the raw model across all datasets. For GSN-guided approaches, all \sagaplus configurations surpass competing methods on the three datasets. On \sdthree, scores are similar for the two entities dataset, where \sdthree already performs well. On the three-entity dataset, most configurations achieve higher scores, and for four entities, our method consistently outperforms all baselines.

Finally, we emphasize that CLIP scores should be interpreted with caution, as CLIP behaves like a bag-of-words model~\cite{yuksekgonul2023when}: it exhibits weak relational understanding, struggles to associate objects with attributes, and lacks sensitivity to word order.

\subsection{GenEval \label{sec:geneval}}

\begin{table*}[tb]
\centering
\caption{Model performance on the GenEval benchmark. Results are from~\cite{ma2025janusflowharmonizingautoregressionrectified}. \saga applied to \sdthree achieves the highest overall performance.}
\small
\npdecimalsign{.} 
\nprounddigits{1} 
\begin{tabular}{llllllllll}
        \toprule
        & \textbf{Method} & \textbf{Params} & \textbf{Single Obj.} & \textbf{Two Obj.} & \textbf{Count.} & \textbf{Colors} & \textbf{Pos.} & \textbf{Color Attri.} & \textbf{Overall$\uparrow$} \\
        \midrule
        \multirow[c]{17}{*}{\rotatebox{90}{\textbf{Reported}}} 
        & LlamaGen~\cite{2024llamagen} & $0.8$B & $71$ & $34$ & $21$ & $58$ & $07$ & $04$ & $32$ \\
        & LDM~\cite{rombach2021highresolution} & $1.4$B & $92$ & $29$ & $23$ & $70$ & $02$ & $05$ & $37$ \\
        & SDv$1.5$~\cite{rombach2021highresolution} & $0.9$B & $97$ & $38$ & $35$ & $76$ & $04$ & $06$ & $43$ \\
        & PixArt-$\alpha$~\cite{chen2023pixartalpha} & $0.6$B & $98$ & $50$ & $44$ & $80$ & $08$ & $07$ & $48$ \\
        & SDv$2.1$~\cite{rombach2021highresolution} & $0.9$B & $98$ & $51$ & $44$ & $85$ & $07$ & $17$ & $50$ \\
        & DALL-E $2$~\cite{ramesh2022hierarchical} & $6.5$B & $94$ & $66$ & $49$ & $77$ & $10$ & $19$ & $52$ \\
        & Emu$3$-Gen ~\cite{2024emu3} & $8$B & $98$ & $71$ & $34$ & $81$ & $17$ & $21$ & $54$ \\
        & SDXL~\cite{2023SDXL} & $2.6$B & $98$ & $74$ & $39$ & $85$ & $15$ & $23$ & $55$ \\
        & IF-XL~\cite{2023IF} & $4.3$B & $97$ & $74$ & $66$ & $81$ & $13$ & $35$ & $61$ \\
        & DALL-E $3$~\cite{2023dalle3} & - & $96$ & $87$ & $47$ & $83$ & $43$ & $45$ & $67$ \\
        & Chameleon~\cite{2024Chameleon} & $34$B & - & - & - & - & - & - & $39$ \\
        & LWM~\cite{2024LWM} & $7$B & $93$ & $41$ & $46$ & $79$ & $09$ & $15$ & $47$ \\
        & SEED-X$^\dagger$~\cite{2024SeedX} & $17$B & $97$ & $58$ & $26$ & $80$ & $19$ & $14$ & $49$ \\
        & Show-o~\cite{2024Showo} & $1.3$B & $95$ & $52$ & $49$ & $82$ & $11$ & $28$ & $53$ \\
        & Janus \cite{2024Janus} & $1.3$B & $97$ & $68$ & $30$ & $84$ & $46$ & $42$ & $61$ \\
        & Transfusion~\cite{2024Transfusion} & $7.3$B & - & - & - & - & - & - & 63 \\
        & JanusFlow~\cite{ma2025janusflowharmonizingautoregressionrectified} & 1.3B & 97 & 59 & 45 & 83 & 53 & 42 & 63 \\
        \midrule
        \multirow[c]{5}{*}{\rotatebox{90}{\textbf{Computed}}} & \sdone~\cite{rombach2021highresolution} & 0.9B & \numprint{98.75} & \numprint{38.64} & \numprint{37.50} & \numprint{77.66} & \numprint{3.75} & \numprint{8.00} & \numprint{44.049} \\
        & \saga \sdone & 0.9B  & \numprint{95.94} & \numprint{64.39} & \numprint{37.19} & \numprint{69.95} & \numprint{8.00} & \numprint{16.75} & \numprint{48.703} \\
        & Syngen \sdone & 0.9B & \numprint{98.12} & \numprint{70.96} & \numprint{40.62} & \numprint{80.32} & \numprint{9.50} & \numprint{34.75} & \numprint{55.713} \\
        & \sagaplus \sdone & 0.9B  & \numprint{99.38} & \numprint{74.49} & \numprint{35.62} & \numprint{78.19} & \numprint{12.50} & \numprint{36.75} & \numprint{56.156} \\
         & \sdthree~\cite{esser2024scalingrectifiedflowtransformers} & 2B & \numprint{99.06} & \numprint{85.61} & \numprint{55.00} & \numprint{87.23} & \numprint{25.50} & \numprint{60.00} & \numprint{68.734} \\ 
        & \saga \sdthree & 2B & \numprint{100.00} & \numprint{94.95} & \numprint{53.44} & \numprint{76.06} & \numprint{40.25} & \numprint{55.00} & \numprint{69.950} \\ 
        \bottomrule
\end{tabular}

\label{tab:geneval}
\end{table*}

GenEval~\cite{ghosh2023geneval} results are reported in \citetable{tab:geneval}, with prior results taken from~\cite{ma2025janusflowharmonizingautoregressionrectified}.
We evaluate only the Syngen GSN method, as it is the second-best performer. Other prior GSN approaches are excluded due to incompatibility with GenEval, where some entities in prompts consist of multiple tokens (\eg ``stop sign''), which these methods cannot handle. Our approach resolves this by averaging attention maps across sub-tokens of the same word, and we reimplemented Syngen accordingly.

\sagaplus outperforms methods with comparable parameter counts and improves over the raw \sdone and Syngen. On \sdthree, \saga achieves better results than other approaches. As our loss is specifically designed to address composition issues and catastrophic neglect, we observe a performance boost on the Two Obj. task. We acknowledge that alternative criterion loss may further enhance performance on specific subtasks.

\subsection{Attend\&Excite metrics}

\begin{table}[]
   
    \centering
      \caption{Full Prompt Similarity score $\times 100$ for our approach and the baseline presented in the main manuscript. Methods with \variantB are another configuration we tested (see \citesection{sec:estim_variance}).}
   
    {
    \small
    \begin{tabular}{llrrr}
    \toprule
     & \multirow{2}{*}{Methods} & \multicolumn{3}{c}{Full Prompt Similarity} \\
    & & 2 & 3 & 4 \\  
    \midrule
\multirow[c]{13}{*}{\rotatebox{90}{\textbf{\sdone}}} & Stable Diffusion & 33.13 & 34.50 & 34.74 \\
 & InitNO \textsubscript{(CVPR'24)}& 34.11 & 35.28 & 35.07 \\
 & \saga & 35.07 & 36.72 & 36.15 \\
 & \sagaone & 35.18 & 36.60 & 36.06 \\
 & \sagavar & 35.22 & 37.06 & 36.44 \\
 & \sagaonevar & 35.26 & 36.86 & 36.35 \\
\cmidrule(lr){2-5}
 & Attend\&Excite\textsubscript{(SIGGRAPH’23)} & 35.09 & 37.08 & 37.44 \\
 & InitNO+\textsubscript{(CVPR'24)} & 35.20 & 37.12 & 37.43 \\
 & Syngen\textsubscript{(NeurIPS'23)} & 35.18 & 37.50 & 37.72 \\
 & \sagaplus & \underline{35.31} & \underline{37.73} & \underline{37.81} \\
 & \sagaplusone & 35.27 & 37.62 & 37.74 \\
 & \sagaplusvar & \textbf{35.35} & \textbf{37.77} & \textbf{37.90} \\
 & \sagaplusonevar & 35.26 & 37.73 & 37.78 \\
\midrule
\multirow[c]{9}{*}{\rotatebox{90}{\textbf{\sdthree}}} & Stable Diffusion\textsubscript{(ICML'24)} & 35.14 & 38.17 & 38.74 \\
 & \sagaStepsevenMomentumzerodotfour & 35.09 & 38.25 & 39.28 \\
 & \sagaStepfiveMomentumzerodotseven & 34.72 & 37.90 & 39.07 \\
 & \sagaoneStepsevenMomentumzerodotfour & \underline{35.15} & \textbf{38.32} & \underline{39.31} \\
 & \sagaoneStepfiveMomentumzerodotseven & 34.76 & 37.93 & 39.13 \\
 & \sagavarStepsevenMomentumzerodotfour & 35.06 & 38.24 & 39.27 \\
 & \sagavarStepfiveMomentumzerodotseven & 34.71 & 37.95 & 39.09 \\
 & \sagaonevarStepsevenMomentumzerodotfour & \textbf{35.16} & \underline{38.29} & \textbf{39.33} \\
 & \sagaonevarStepfiveMomentumzerodotseven & 34.82 & 37.90 & 39.04 \\
\bottomrule
\end{tabular}

    }
   
    \label{tab:fulltext}
\end{table}

\begin{table}[]
    \centering
     \caption{Minimum Object Similarity $\times 100$ for our approach and the baselines presented in the main manuscript. Methods with \variantB are another configuration we tested (see \citesection{sec:estim_variance}).}
      {
    \small
    \begin{tabular}{llrrr}
    \toprule
     & \multirow{2}{*}{Methods} & \multicolumn{3}{c}{Minimum Object Similarity} \\
    & & 2 & 3 & 4 \\  
    \midrule
\multirow[c]{13}{*}{\rotatebox{90}{\textbf{\sdone}}} & Stable Diffusion & 24.00 & 20.63 & 18.96 \\
 & InitNO \textsubscript{(CVPR'24)}& 25.12 & 21.06 & 19.07 \\
 & \saga & 26.34 & 22.70 & 19.86 \\
 & \sagaone & 26.39 & 22.60 & 19.69 \\
 & \sagavar & 26.62 & 23.17 & 20.25 \\
 & \sagaonevar & 26.55 & 22.99 & 20.12 \\
\cmidrule(lr){2-5}
 & Attend\&Excite\textsubscript{(SIGGRAPH’23)} & 26.34 & 23.27 & 21.03 \\
 & InitNO+\textsubscript{(CVPR'24)} & 26.39 & 23.29 & 21.04 \\
 & Syngen\textsubscript{(NeurIPS'23)} & 26.40 & 23.20 & 20.87 \\
 & \sagaplus & 26.67 & 23.74 & 21.25 \\
 & \sagaplusone & 26.62 & 23.68 & 21.24 \\
 & \sagaplusvar & \textbf{26.72} & \textbf{23.78} & \textbf{21.27} \\
 & \sagaplusonevar & \underline{26.68} & \underline{23.76} & \underline{21.26} \\
\midrule
\multirow[c]{9}{*}{\rotatebox{90}{\textbf{\sdthree}}} & Stable Diffusion\textsubscript{(ICML'24)} & 26.09 & 23.34 & 21.25 \\
 & \sagaStepsevenMomentumzerodotfour & 26.19 & 23.55 & 21.97 \\
 & \sagaStepfiveMomentumzerodotseven & 25.72 & 23.20 & 21.70 \\
 & \sagaoneStepsevenMomentumzerodotfour & \textbf{26.29} & \textbf{23.60} & \underline{21.97} \\
 & \sagaoneStepfiveMomentumzerodotseven & 25.72 & 23.18 & 21.73 \\
 & \sagavarStepsevenMomentumzerodotfour & 26.15 & 23.55 & 21.96 \\
 & \sagavarStepfiveMomentumzerodotseven & 25.71 & 23.25 & 21.72 \\
 & \sagaonevarStepsevenMomentumzerodotfour & \underline{26.28} & \underline{23.58} & \textbf{21.98} \\
 & \sagaonevarStepfiveMomentumzerodotseven & 25.76 & 23.20 & 21.70 \\
\bottomrule
\end{tabular}

    }
    
    \label{tab:minobj}
\end{table}

\begin{table}[]
    \centering

     \caption{Text-Text Similarity $\times 100$ for our approach and the baselines presented in the main manuscript. Methods with \variantB are another configuration we tested (see \citesection{sec:estim_variance}).}
   
      {
    \small
    \begin{tabular}{llrrr}
    \toprule
     & \multirow{2}{*}{Methods} & \multicolumn{3}{c}{ Text-Text Similarity} \\
    & & 2 & 3 & 4 \\  
    \midrule
\multirow[c]{13}{*}{\rotatebox{90}{\textbf{\sdone}}} & Stable Diffusion & 76.81 & 73.22 & 68.73 \\
 & InitNO \textsubscript{(CVPR'24)}& 79.01 & 74.07 & 68.99 \\
 & \saga & 80.78 & 75.72 & 70.78 \\
 & \sagaone & 81.23 & 75.83 & 70.62 \\
 & \sagavar & 81.16 & 76.24 & 71.02 \\
 & \sagaonevar & 81.41 & 76.31 & 71.11 \\
\cmidrule(lr){2-5}
 & Attend\&Excite\textsubscript{(SIGGRAPH’23)} & 80.31 & 75.80 & 70.77 \\
 & InitNO+\textsubscript{(CVPR'24)} & 80.80 & 75.94 & 70.41 \\
 & Syngen\textsubscript{(NeurIPS'23)} & 81.22 & 76.60 & 71.65 \\
 & \sagaplus & 81.81 & 77.26 & 72.20 \\
 & \sagaplusone & \textbf{82.13} & 77.15 & 72.45 \\
 & \sagaplusvar & 82.05 & \textbf{77.42} & \textbf{72.53} \\
 & \sagaplusonevar & \underline{82.09} & \underline{77.39} & \underline{72.47} \\
\midrule
\multirow[c]{9}{*}{\rotatebox{90}{\textbf{\sdthree}}} & Stable Diffusion\textsubscript{(ICML'24)} & \textbf{81.71} & 77.98 & 72.86 \\
 & \sagaStepsevenMomentumzerodotfour & 81.28 & 77.94 & \underline{73.33} \\
 & \sagaStepfiveMomentumzerodotseven & 80.36 & 77.16 & 72.35 \\
 & \sagaoneStepsevenMomentumzerodotfour & \underline{81.32} & \textbf{78.14} & 73.07 \\
 & \sagaoneStepfiveMomentumzerodotseven & 80.21 & 77.03 & 72.45 \\
 & \sagavarStepsevenMomentumzerodotfour & 81.22 & \underline{78.13} & \textbf{73.35} \\
 & \sagavarStepfiveMomentumzerodotseven & 80.26 & 77.18 & 72.50 \\
 & \sagaonevarStepsevenMomentumzerodotfour & 81.20 & 77.99 & 73.19 \\
 & \sagaonevarStepfiveMomentumzerodotseven & 80.80 & 76.77 & 72.18 \\
\bottomrule
\end{tabular}

    }
  
    \label{tab:ttsim}
\end{table}

Authors in Attend\&Excite~\cite{chefer2023attendandexcite} propose three scores based on the CLIP Score, relying on a CLIP model\footnote{\url{https://huggingface.co/openai/clip-vit-base-patch16}}\citep{radford2021learning} and a BLIP model\footnote{\url{https://huggingface.co/Salesforce/blip-image-captioning-base}}\citep{li2022blip}. They derive 80 versions of the prompt using as many templates, available on their github\footnote{\url{https://github.com/yuval-alaluf/Attend-and-Excite/}}. Each template is filled with the entities from the original prompt. Afterward, they calculate the CLIP embedding for each prompt version and then compute the average of these embeddings. With these, they derive three distinct scores:
\begin{itemize} 
\item \textit{Full Prompt Similarity}: The cosine similarity between the CLIP embedding of the generated image and the averaged embedding of the 80 variations is computed. 
\item \textit{Minimum Object Similarity}: For each entity, the average text CLIP embedding is derived from the templates. The cosine similarity between the generated image and each entity's average embedding is computed, and the lowest similarity value is selected. 
\item \textit{Text-Text Similarity}: The caption generated for the image (using BLIP) is compared with the averaged embedding of the 80 variations from the original prompt, using cosine similarity. \end{itemize}

We present the Full Prompt Similarity in \citetable{tab:fulltext}, Minimum Object Similarity in \citetable{tab:minobj}, and the Text-Text Similarity in \citetable{tab:ttsim}.

Globally, whatever the metric, the scores are close from each other for all methods. It highlights the limits of the metrics that seem to be ‘crushing’ the scores around a small range of values. The text-text similarity has the largest range of scores among methods. According to all metrics, the methods we propose nevertheless exhibit the best performances, except for \sdthree with 2 entities on Text-Text Similarity.

\clearpage
\section{User Study}
\label{appendix:user_study}

\begin{figure}[h]
    \centering
    \includegraphics[width=0.75\linewidth]{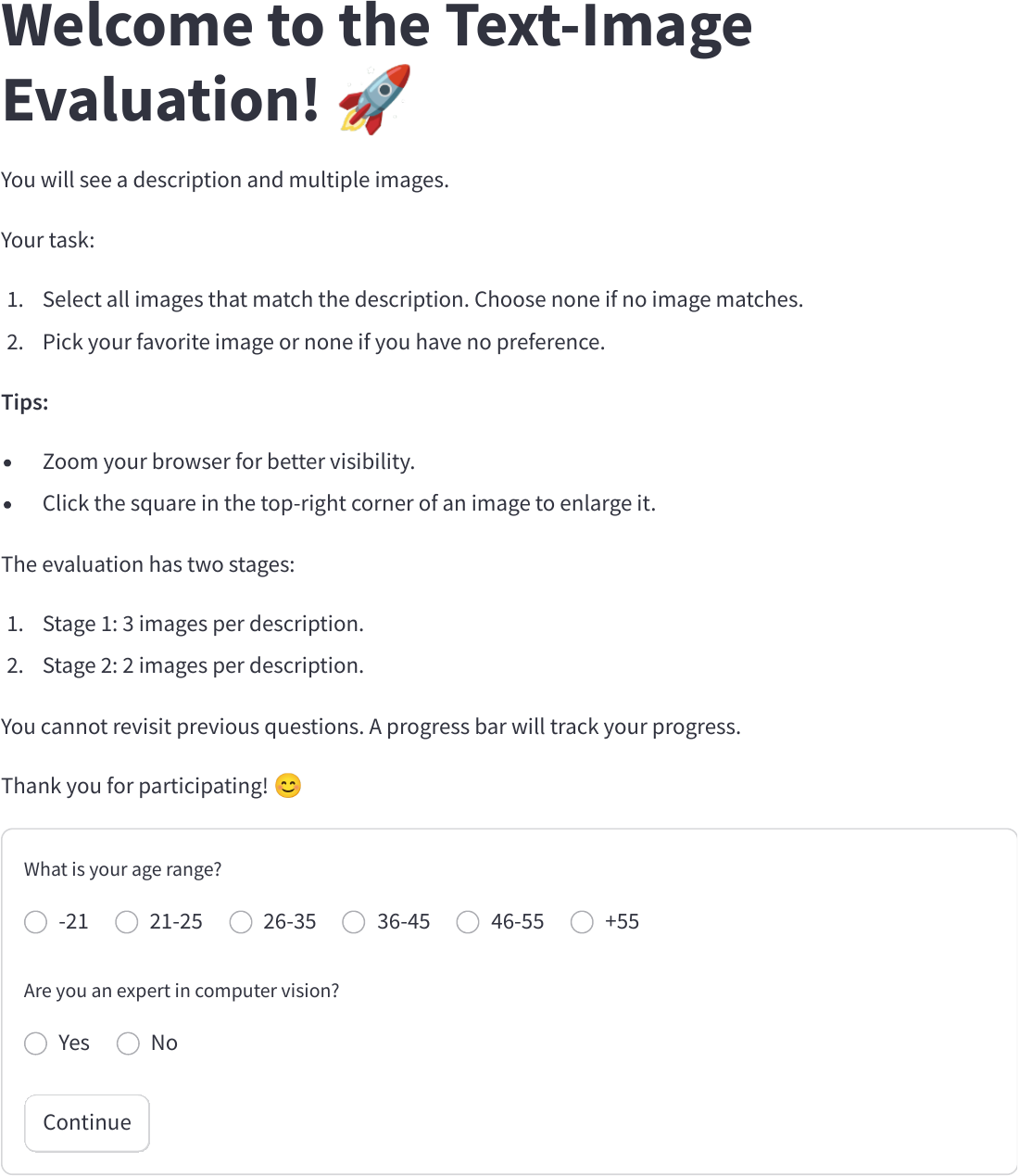}
    \caption{A screenshot of our user study interface, presenting the initial guidelines provided to participants before they begin the annotation task.}
    \label{fig:user_study_start}
\end{figure}

\begin{figure}[h]
    \centering
    \includegraphics[width=0.75\linewidth]{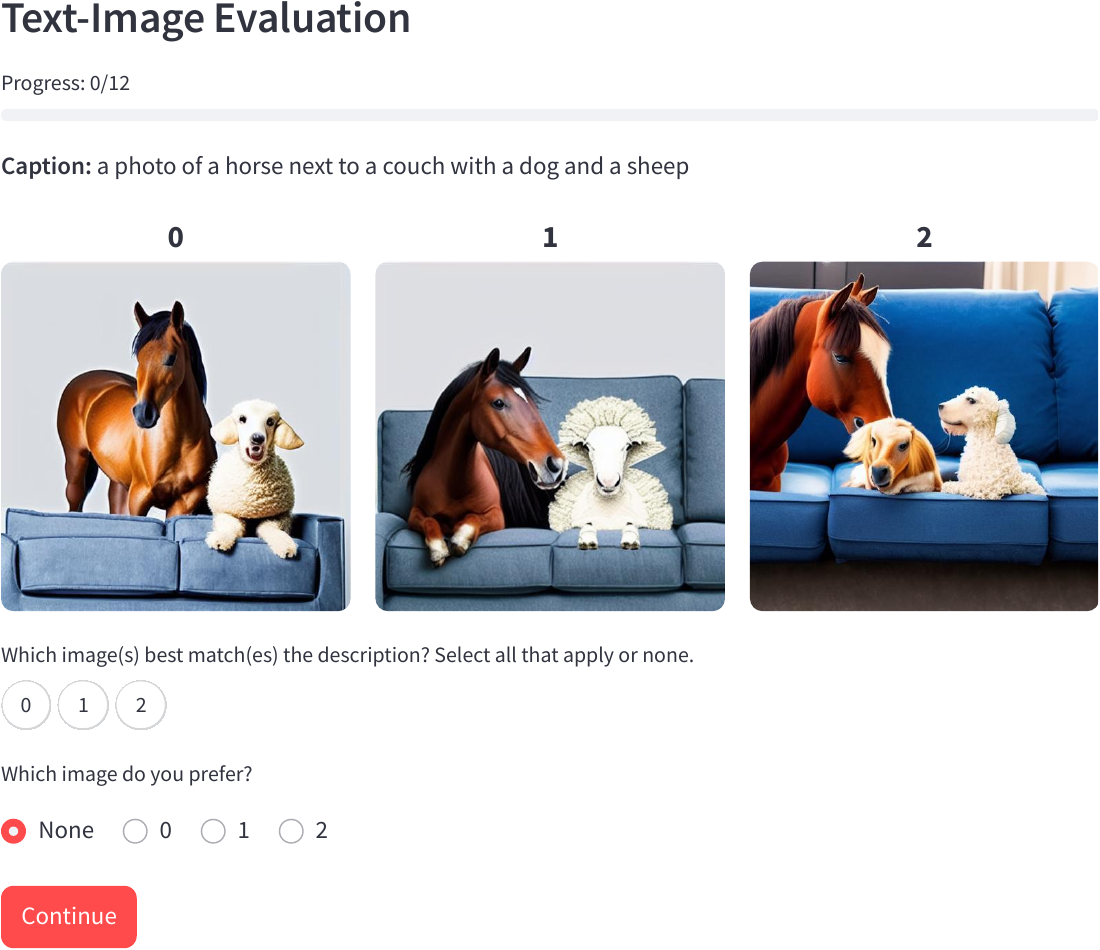}
    
     \caption{An example of the annotation page used to evaluate methods with \gsn guidance. Participants were asked to compare the generated images based on quality and prompt alignment.}
    \label{fig:user_study_sd14}
\end{figure}

\begin{figure}[h]
    \centering
    \includegraphics[width=0.75\linewidth]{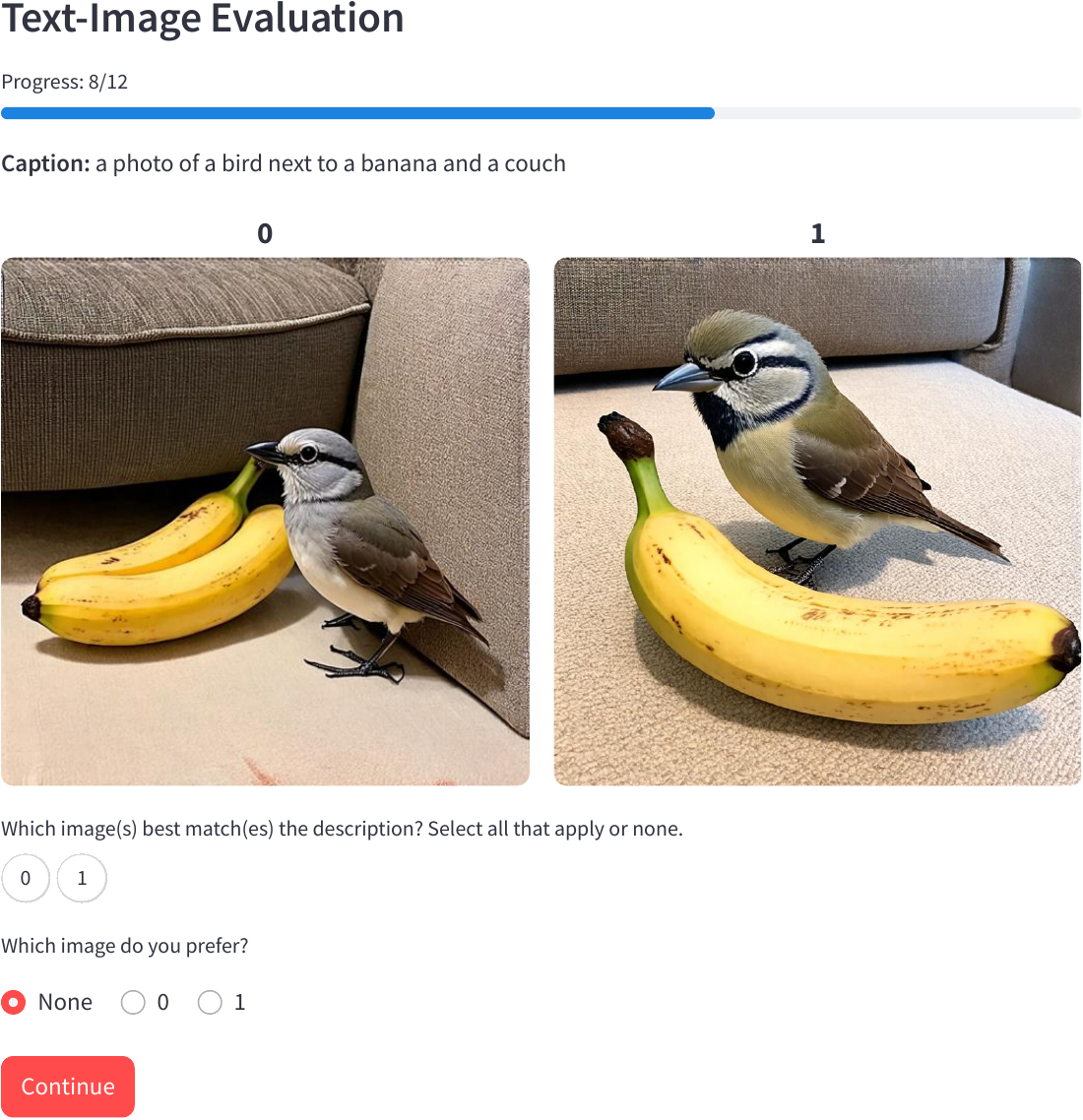}
   
    \caption{The annotation interface for the user-based evaluation on the \sdthree backbone.}
    \label{fig:user_study_sd3}
\end{figure}

\begin{figure}[h]
    \centering
    \includegraphics[width=0.9\linewidth]{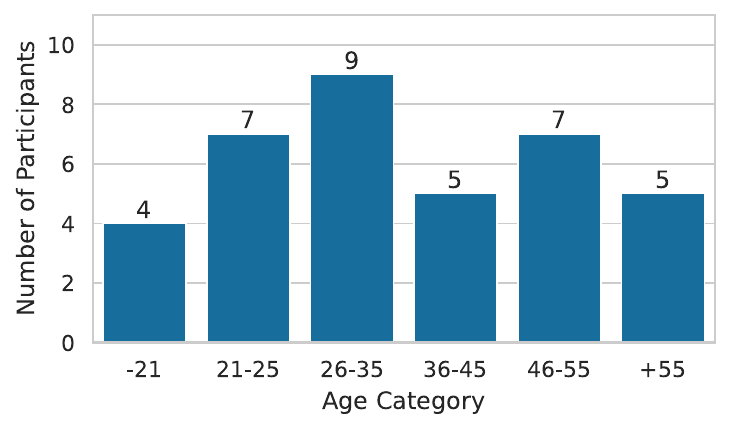}
    \caption{Distribution of participants by age category.}
    \label{fig:user_study_repartition}
\end{figure}

We created two distinct image datasets for the user study. For the comparison on \sdone, we evaluated the \gsn methods by randomly selecting 50 prompts from each of our datasets containing 2, 3, and 4 entities, for a total of 150 prompts. For each prompt, a single seed was chosen randomly from 16 available seeds, and we generated a corresponding image triplet using InitNO+, Syngen, and \sagaplus. For the comparison on \sdthree, which already performs well with two entities, we focused on prompts with 3 and 4 entities. We selected 75 prompts from each of these datasets, totaling 150 prompts. A random seed was again used for each prompt to generate image pairs from the baseline \sdthree method and our \saga.

Each participant was required to complete 12 annotation tasks, randomly drawn from our pre-compiled datasets: 6 for the \gsn methods and 6 for the \sdthree comparison. The protocol for each task was twofold: first, a semantic matching stage where participants could select multiple images that correctly matched the caption (or select none); second, a preference selection stage where they could choose only their single preferred image (or none). We present the user interface, including the guidelines, in \citefigure{fig:user_study_start}. An example of an annotation page for the \gsn methods is shown in \citefigure{fig:user_study_sd14}, with a similar layout used for the \sdthree evaluation.

A total of 37 participants participated in the study, including 6 experts in computer vision. We present the age distribution of the participants in \citefigure{fig:user_study_repartition}.

\clearpage
\section{Qualitative Examples \label{appendix:qualitative}}

\begin{figure*}[h]
    \centering
    \includegraphics[width=0.98\linewidth]{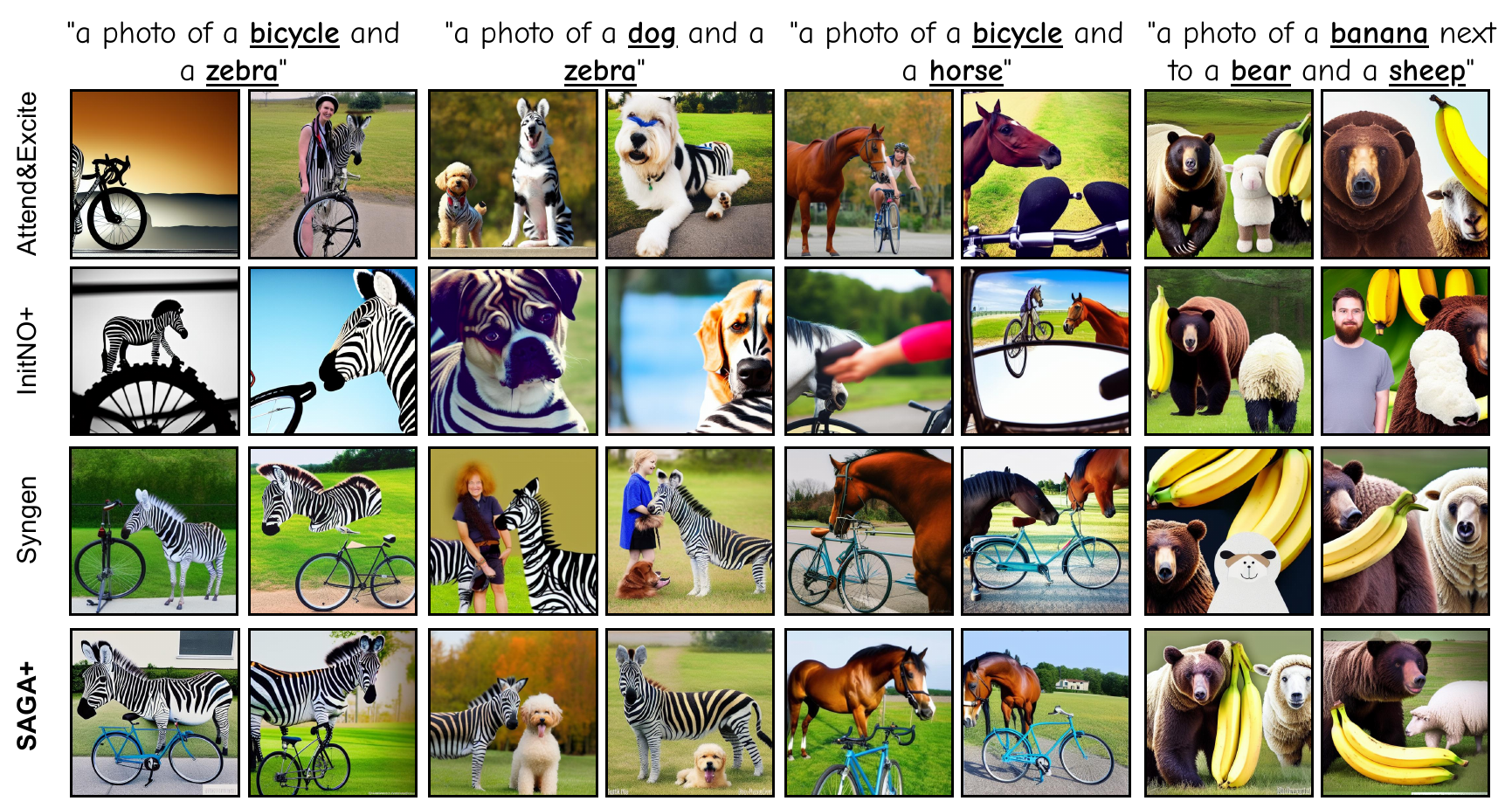}
    \caption{Generated images across different methods using \sdone. Images in the same column are generated with the same seed.}
    \label{fig:qualigsna}
\end{figure*}
\begin{figure*}[h]
    \centering    \includegraphics[width=0.98\linewidth]{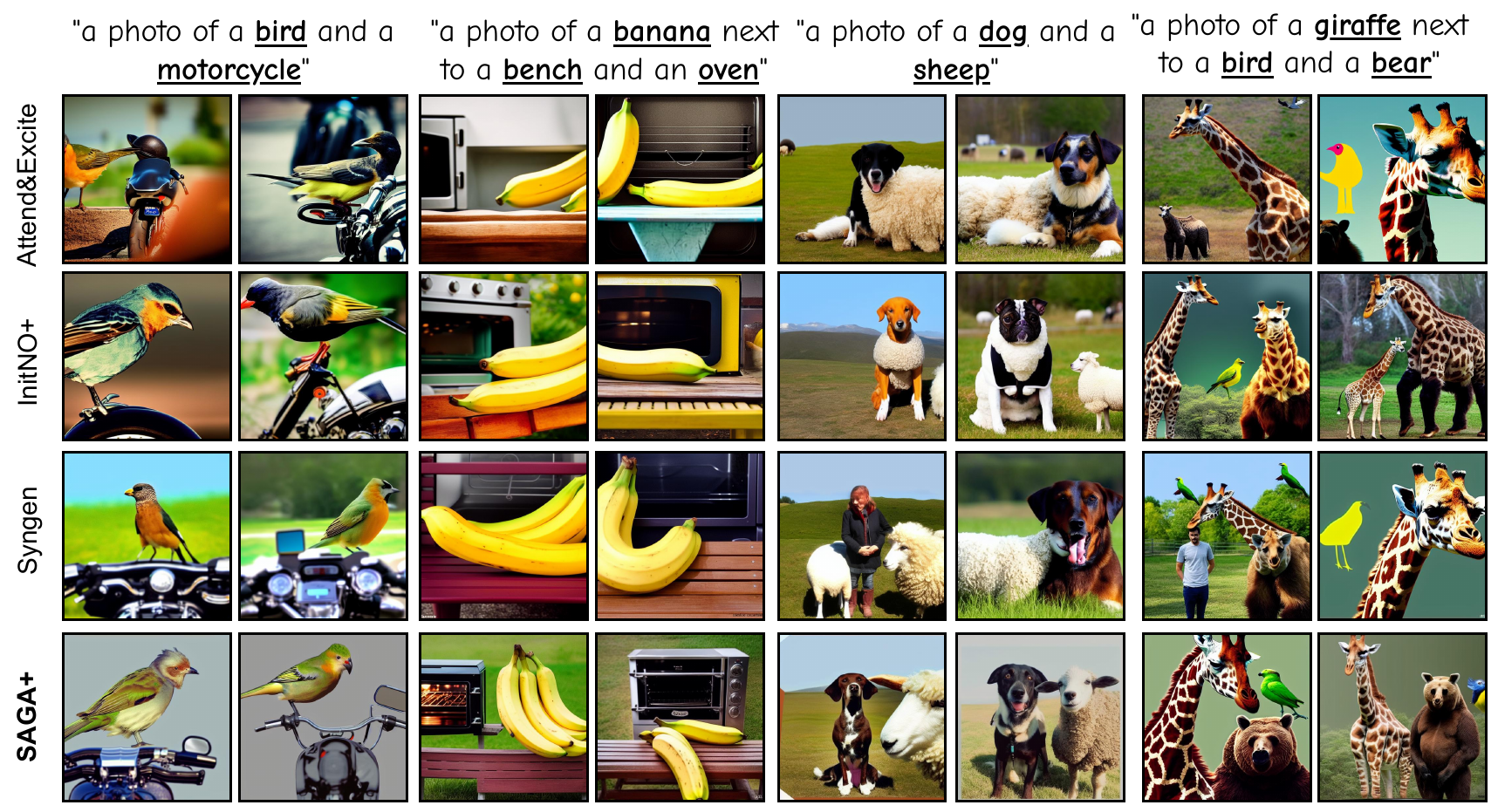}
   \caption{Generated images across different methods using \sdone. Images in the same column are generated with the same seed.}
    \label{fig:qualigsnb}
\end{figure*}

\begin{figure*}[h]
    \centering    \includegraphics[width=0.98\linewidth]{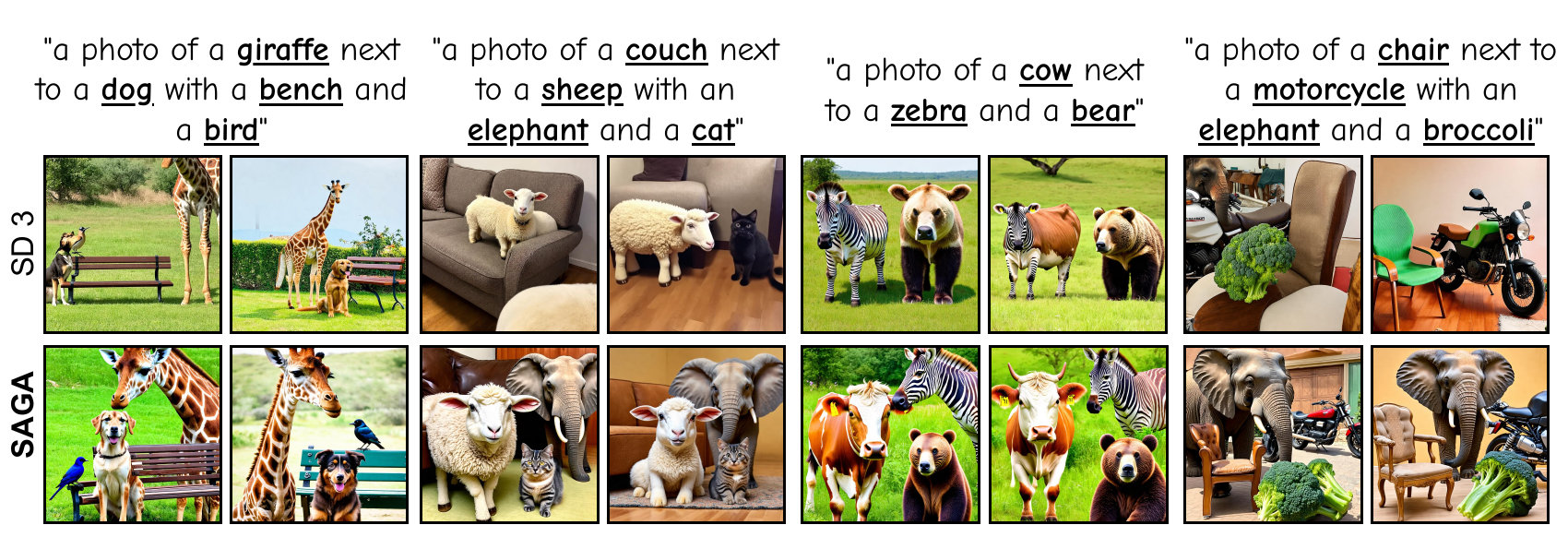}
    \caption{Images generated with \saga and \sdthree with the same seeds/prompts.}
    \label{fig:qualisd3a}
\end{figure*}

\begin{figure*}[h]
    \centering    \includegraphics[width=0.98\linewidth]{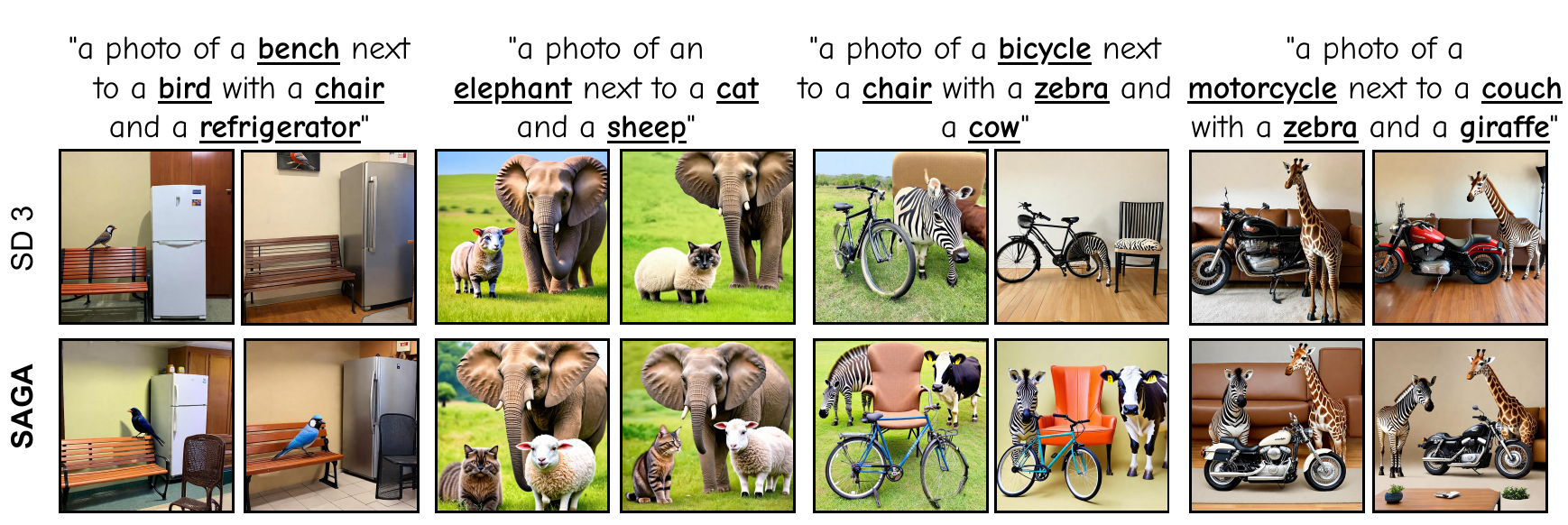}
    \caption{Images generated with \saga and \sdthree with the same seeds/prompts.}
    \label{fig:qualisd3b}
\end{figure*}

\begin{figure*}[h]
    \centering    \includegraphics[width=0.98\linewidth]{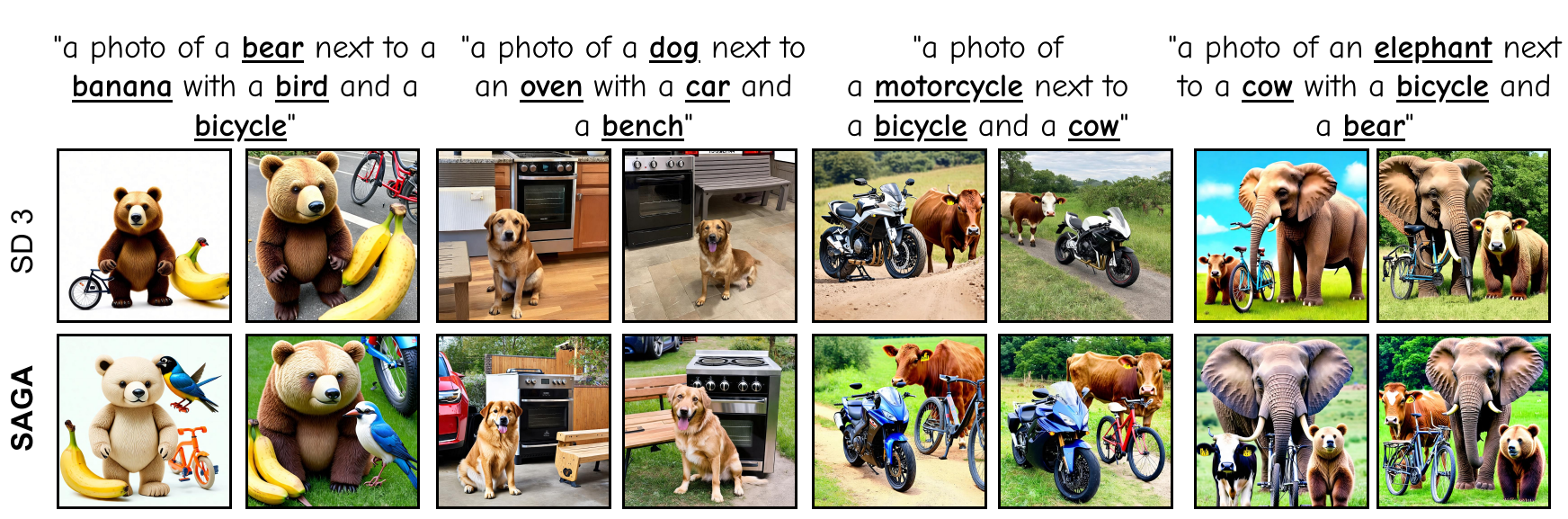}
    \caption{Images generated with \saga and \sdthree with the same seeds/prompts.}
    \label{fig:qualisd3c}
\end{figure*}

We provide additional qualitative samples. \citefigure{fig:qualigsna} and \citefigure{fig:qualigsnb} illustrate results for \gsn guidance with the \sdone backbone. Further images generated using the \sdthree backbone are presented in \citefigure{fig:qualisd3a}, \citefigure{fig:qualisd3b}, and \citefigure{fig:qualisd3c}.

\end{document}